\documentclass{article}
\usepackage{arxiv}

\usepackage[utf8]{inputenc} 
\usepackage[T1]{fontenc}    
\usepackage{hyperref}       
\usepackage{url}      
\usepackage{amssymb}

\usepackage{amsfonts}       
\usepackage{nicefrac}       
\usepackage{microtype}      
\usepackage{graphicx}%
\usepackage{multirow}%
\usepackage{caption}
\usepackage{amsthm}%
\usepackage{cleveref}
\usepackage{mathrsfs}%
\usepackage[title]{appendix}%
\usepackage{xcolor}%
\usepackage{textcomp}%
\usepackage{manyfoot}%
\usepackage{booktabs}%
\usepackage{algorithm}%
\usepackage{algorithmicx}%
\usepackage{algpseudocode}%
\usepackage{listings}%
\usepackage{dirtytalk}
\usepackage[english]{babel}
\usepackage{soul}
\usepackage{csquotes}
\usepackage[export]{adjustbox}
\usepackage{fontawesome5}
\usepackage{tikz}
\usepackage{blindtext}
\usepackage{comment}
\usepackage[numbers]{natbib} 

\usepackage{lscape} 
\usepackage{ulem}
\usepackage{graphicx}
\usepackage{natbib}
\usepackage{doi}
\usepackage{fancyhdr}
\title{SMART-Vision: Survey of Modern Action Recognition Techniques in Vision}

\newif\ifuniqueAffiliation
\uniqueAffiliationtrue

\ifuniqueAffiliation 
\author{ 
Ali K. AlShami$^{1\dagger}$, Ryan Rabinowitz$^{1}$, Khang Lam$^{2}$, Yousra Shleibik$^{1}$, Melkamu Mersha$^{1}$ \\ \textbf{Terrance Boult$^{1}$}, \textbf{Jugal Kalita$^{1}$} \\
$^{1}$Computer Science Department, University of Colorado, Colorado Springs \\
$^{2}$Information Technology Department, Can Tho University \\
\texttt{aalshami@uccs.edu}, \texttt{rrabinow@uccs.edu}, \texttt{lnkhang@ctu.edu.vn}, \texttt{yshleibi@uccs.edu}, \\
\texttt{mmersha@uccs.edu}, \texttt{tboult@uccs.edu}, \texttt{jkalita@uccs.edu}
}
\fi

\hypersetup{
pdftitle={A template for the arxiv style},
pdfsubject={q-bio.NC, q-bio.QM},
pdfauthor={David S.~Hippocampus, Elias D.~Striatum},
pdfkeywords={First keyword, Second keyword, More},
}

\begin{document}
\maketitle

\begin{abstract}
Human Action Recognition (HAR) is a challenging domain in computer vision, involving recognizing complex patterns by analyzing the spatiotemporal dynamics of individuals' movements in videos. These patterns arise in sequential data, such as video frames, which are often essential to accurately distinguish actions that would be ambiguous in a single image. HAR has garnered considerable interest due to its broad applicability, ranging from robotics and surveillance systems to sports motion analysis, healthcare, and the burgeoning field of autonomous vehicles. While several taxonomies have been proposed to categorize HAR approaches in surveys, they often overlook hybrid methodologies and fail to demonstrate how different models incorporate various architectures and modalities. In this comprehensive survey, we present the novel SMART-Vision taxonomy, which illustrates how innovations in deep learning for HAR complement one another, leading to hybrid approaches beyond traditional categories. Our survey provides a clear roadmap from foundational HAR works to current state-of-the-art systems, highlighting emerging research directions and addressing unresolved challenges in discussion sections for architectures within the HAR domain. We provide details of the research datasets that various approaches used to measure and compare goodness HAR approaches. We also explore the rapidly emerging field of Open-HAR systems, which challenges HAR systems by presenting samples from unknown, novel classes during test time.	
\end{abstract}


\section{Introduction}
Human Action Recognition (HAR) is a complex area focused on identifying and understanding human actions by analyzing the spatiotemporal dynamics of movements of individuals in videos. Effective HAR solutions are essential for many systems interacting with humans, from autonomous driving to intelligent self-checkout machines. With the advent of deep learning, numerous systems for HAR have emerged and rapidly evolved. However, multiple major paradigm shifts (such as adopting transformers) obscure the synergy between design choices in different paradigms. Through our Survey of Modern Action Recognition Techniques in Vision (SMART-Vision) taxonomy, we show that many of the proposed systems for HAR are inherently hybrid, bridging the traditional monolithic paradigms proposed in prior works. To this end, we comprehensively analyze five major shifts in HAR literature, revealing new insights about hybrid methodologies that could not be attained from prior surveys that analyzed a smaller scope of work or disregarded hybrid aspects. We also present the first comprehensive analysis of Open-HAR, an emerging evaluation methodology where systems encounter novel classes at test-time.

\begin{figure}[hbt!]
    \centering
    \includegraphics[width=17cm]{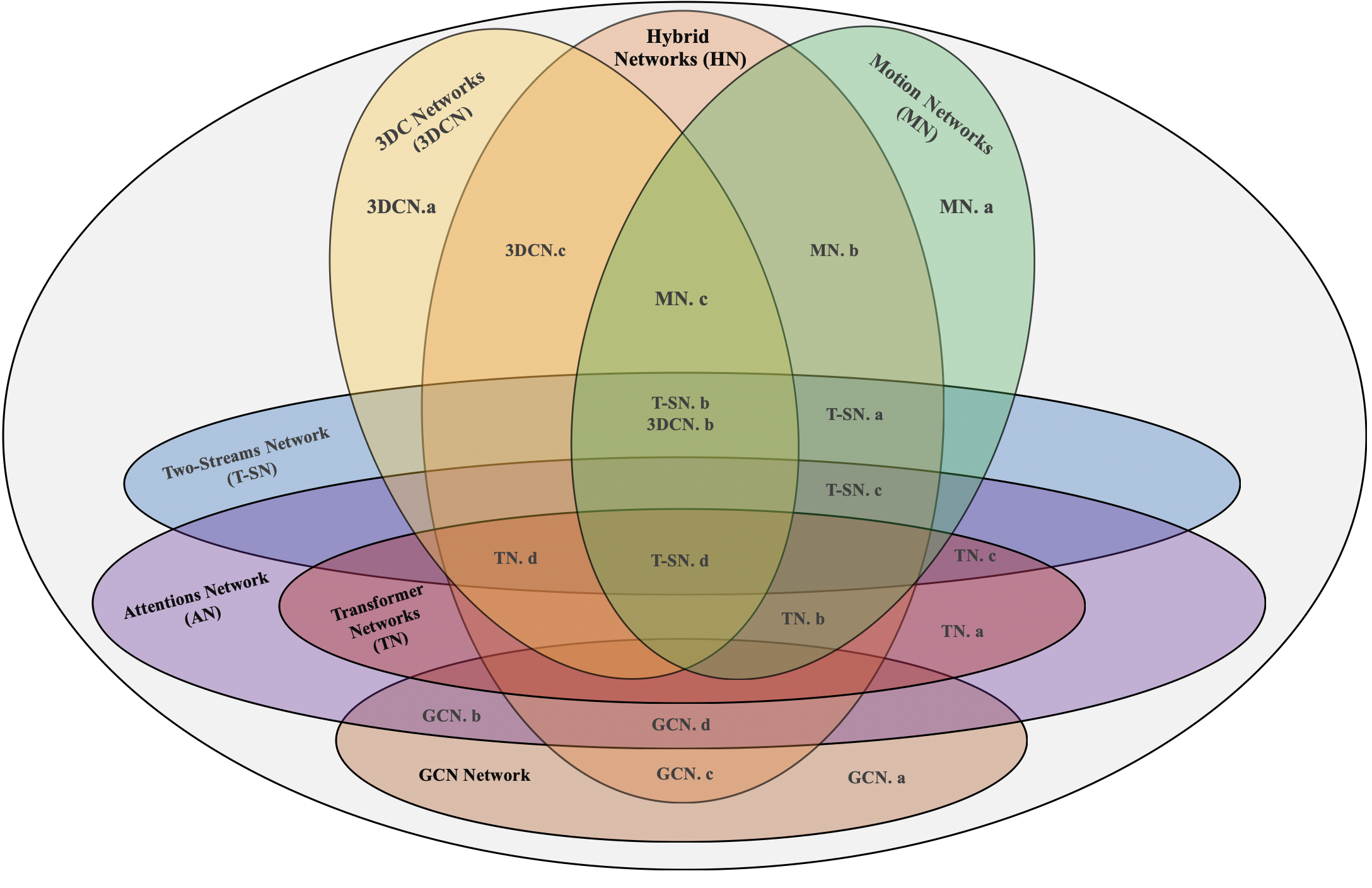}
    \caption{SMART-Vision Venn Diagram.
    \\ SMART-Vision diagram illustrates the formation of hybrid approaches that transcend the traditional categories in Section \ref{sec:Deep learning approaches}, including Two-Stream Networks (T-SNs) (Subsection~\ref{subsec:two-stream}, Table~\ref{Tab:TwoStream-based-models}), 3D Convolutional Networks (Subsection~\ref{subsec:3D CNNs}, Table~\ref{Tab:3DCN-models}), Graph Convolutional Networks (3DCN)(Subsection~\ref{subsec:GCNs}, Table~\ref{Tab:GCNs-based-models}), Motion Networks (MNs) (Subsection~\ref{subsec:Motion models}, Table~\ref{Tab:MNs-models}), Transformer Networks (TN)(Subsection~\ref{sec:Transformer Models}, Table ~\ref{Tab:Transformer-table}), and Hybrid Networks (Subsection~\ref{sec:Hybrid Models}).
    The SMART-Vision Taxonomy does not show some subsections, including additional novel work Subsection~\ref{sec:Additional Recent Novel Work}.\\
    \textbf{Note:} The relative sizes of the shapes in the diagram do not indicate the volume of research: the purpose is to depict the categorization and interrelation of these sub-areas.}
    \label{fig:HARVennDiagrame}
\end{figure}

Scholars have previously proposed taxonomies to classify Human Activity Recognition (HAR) methodologies. For instance, Sun et al.\cite{sun2022human} categorized HAR methods, techniques, and algorithms based on the type of input data modality. Morshed et al.\cite{morshed2023human} organized their taxonomy around feature extraction and activity type. Other surveys have taken more specific approaches, such as Ahmad et al.\cite{ahmad2021graph}, who discuss HAR methods utilizing Graph Convolutional Networks (GCNs); and Ulhaq et al.\cite{ulhaq2022vision}, who explore transformer-based architectures for HAR. Other surveys have explored HAR from various perspectives, such as action representation and analysis~\cite{pareek2021survey}, neural network techniques~\cite{jobanputra2019human}, and specific problem sub-types like localization/detection, classification, and prediction~\cite{kong2022human}. Two other surveys \cite{pareek2021survey, kumar2024survey} proposed taxonomies for human action classification spanning architectural advancements, but the discussion on multi-modal systems and hybrid architectures was short and cursory.

While numerous taxonomies have been proposed to categorize HAR approaches ~\cite{sun2022human,morshed2023human,ahmad2021graph,ulhaq2022vision,pareek2021survey,kong2022human,jobanputra2019human, kumar2024survey}, they either lack sufficient scope to analyze hybrid methodologies or include only a brief analysis.
In this comprehensive survey, we introduce the SMART-Vision taxonomy to demonstrate how innovations in Deep Learning-HAR are related, forming hybrid approaches that transcend traditional categories. The Venn Diagram in Figure \ref{fig:HARVennDiagrame} illustrates how various architectures  (shown in different colors) intersect, representing hybrid methodologies. We present the papers that form each intersection in accompanying tables, linked in the caption and listed in subsections of Section \ref{sec:Deep learning approaches}. For example, as illustrated in the Venn Diagram (Figure \ref{fig:HARVennDiagrame}), The Transformer Networks (TN) intersect with the attention in (TN.a), Attention and 3D Convolutional Networks in (TN.b), attention and two stream networks in (TN.c), and attention and two stream networks with 3D Convolutional Networks in (TN.b). For convenience, we also provide a table of all research papers from each intersection in Table~\ref{Tab:Transformer-table}.
Using the SMART-Vision taxonomy, our analysis traces the evolution of recent Deep Learning techniques from foundational shifts to advanced hybrid approaches, representing continual progress in the domain. Our taxonomy is designed to be an accessible tool for new researchers and experts, providing an overview of major shifts in HAR literature and their formation of hybrid approaches.

Further, we present a discussion for each network type with comprehensive evaluation and performance comparison in Subsection (\ref{sec:Holistic_Discussion}), as well as an in-depth analysis of datasets Section (\ref{sec:datasets}), research challenges and limitations Section (\ref{sec:research challenges}), and the emerging challenge of Open HAR Section (\ref{sec:Open World Recognition}), with a level of detail not previously seen in HAR surveys. By doing so, we aim to provide a valuable resource that will facilitate future research and development in the HAR domain.
\newline
\noindent
Our key contributions include:
\begin{itemize}

\item Our SMART-Vision taxonomy and analysis (illustrated in Figure~\ref{fig:HARVennDiagrame} and Tables provided in the caption), which reveals how fundamentally different HAR approaches have been extended into hybrid methodologies.
\item A bottom-up analysis of each major archetype in our taxonomy, providing foundations and forward-looking insights for the literature discussed.
\item A thorough analysis of Open-HAR, an emerging problem area dealing with novelty and unknown inputs in HAR.
\item Provide a comprehensive evaluation and performance comparison for Deep Learning Approaches in HAR.
\item A comprehensive discussion of modern datasets used for HAR.
\item A comprehensive discussion of research challenges and limitations

\end{itemize}

\section{Overview of Vision-Based Human Action Recognition and Its Applications} \label{sec:overview}

As Artificial Intelligence (AI) rapidly integrates with every facet of our daily lives, Human Action Recognition is empowering the digital systems that surround us. From enhancing automated threat detection for video surveillance to enabling autonomous cars to recognize pedestrian intentions, Human Action Recognition has the potential to revolutionize how technology interacts with humans.

\begin{figure}[hbt!]
    \centering
    \includegraphics[width=\linewidth]{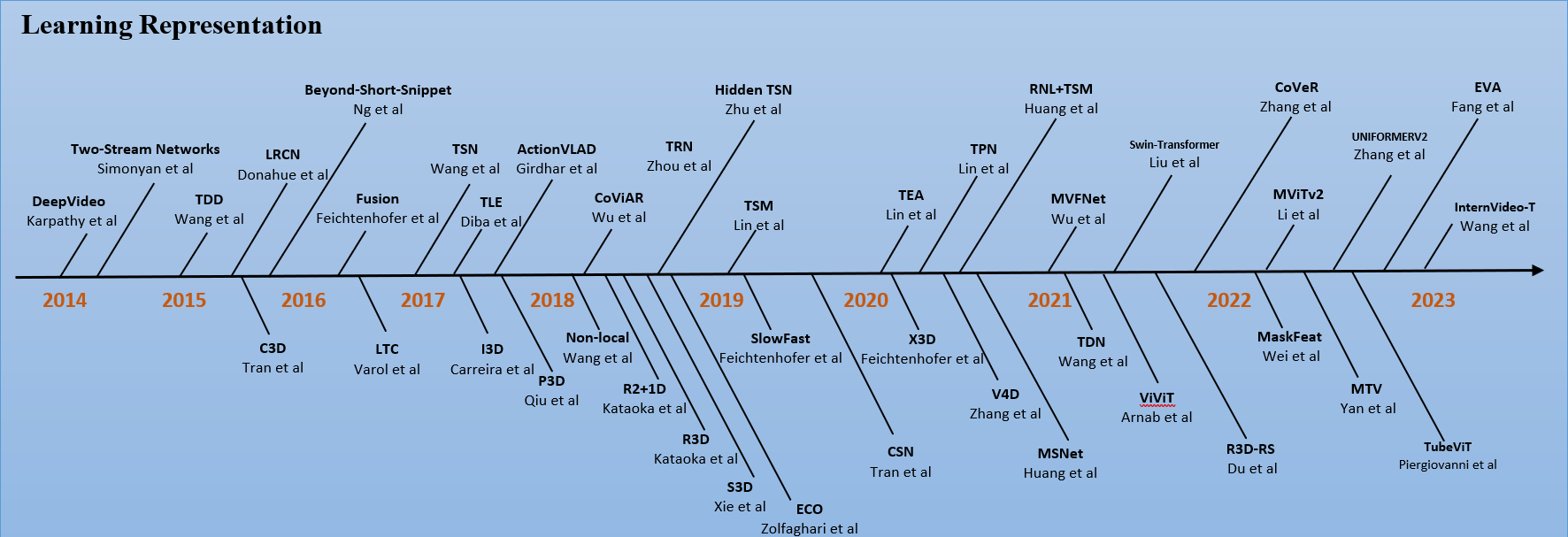}
    \caption{A chronological overview of recent representative work in HAR. The chronological overview extends the work of Zhu et al.~\cite{zhu2020comprehensive}. The papers listed here represent major advancements in HAR; we discuss many more in Section 3 that could not be shown here due to display limitations.}
    \label{figure:timeline}
\end{figure}

In the mid-2010s, after the start of the deep learning revolution~\cite{krizhevsky2012imagenet}, approaches such as improved Dense Trajectory (iDT) \cite{wang2014action} and DeepVideo \cite{karpathy2014large} incorporated deep learning in HAR for the first time. From there, as shown in Figure \ref{fig:HARVennDiagrame}, HAR exploded, rapidly adopting new deep learning paradigms from adjacent problems and creating purpose-built approaches for HAR. However, this incredible progress has created a web of entangled developments that --without robust analysis-- can obscure the synergy behind apparently divergent approaches.

Recognizing and differentiating the sub-problems that comprise HAR is essential for understanding proposed methodologies, distinguishing among various HAR approaches, explaining what works well, and discussing future challenges. We recognize three fundamental problems that HAR encompasses: Action classification - determining the class or label of action in videos \cite{zhu2016key}, Temporal action detection or localization - ``when'' an action occurs within a video \cite{shou2016temporal}, and Spatial-temporal action detection - ``when'' and ``where'' within a given space and time frame an action occurs.
Additionally, we recognize an \textbf{emerging} problem adjacent to Action classification: Open-HAR, where systems must detect novel actions (in the Open-Set setting) and learn to classify the detected novelty (in the Open-World setting). We provide the first detailed analysis of Open-HAR work in Section \ref{sec:Open World Recognition}.

Activities in HAR have been grouped into six categories based on complexity and duration~\cite{vrigkas2015review}: gestures, actions, interactions, multiple activities, human behaviors, and events. Gestures are basic movements of human body parts, like hand waving~\cite{yang2014analysis}. Actions are a single person's specific activities, such as jumping or playing tennis~\cite{ni2015motion}. Interactions encompass engagements between humans and objects or between people, like playing the piano or shaking hands~\cite{patron2012structured}. Multiple activities or group tasks, considered highly challenging in HAR, involve interactions among multiple persons~\cite{tran2012part}. Human behaviors relate to actions reflecting emotions, personality, and psychological states~\cite{martinez2014don}. Lastly, events represent high-level activities that denote social interactions and intentions~\cite{lan2011discriminative}. Systems that address specific categories by design are inherently tailored to different applications. \newline

HAR is a fundamental task within the broader realm of multimedia and computer vision applications. It plays a crucial role in a variety of applications, including interactive multimedia systems, smart robotics, autonomous vehicles, video surveillance, shopping experiences, and detailed sports motion analysis, as highlighted in Alshami et al. \cite{al2022generating}. We provide a brief discussion for each application as follows: \newline

\textbf{\textit{Smart Robotics:}}
Robots perform tasks without human intervention, including manufacturing, healthcare, earth and space exploration, packing and packaging, and transport. HAR can greatly enhance the capabilities of robots in these domains. By enabling robots to recognize and respond to human actions, such as hand gestures and body movements, HAR can enhance their ability to perform complex tasks and improve their accuracy, efficiency, and utility. For example, in manufacturing, robots can work collaboratively with humans, recognizing and responding to their actions. This may allow a robot to provide on-demand assistance with tasks or avoid duplication of effort if humans take over a task. In healthcare, robots can use HAR to recognize and respond to patient movements and gestures, enabling them to provide more personalized and effective care or alert healthcare personnel in the event of a crisis. Overall, HAR enhances robot interaction with humans, making them more useful and effective in various applications.

\textbf{\textit{Autonomous Vehicles:}}
Self-driving technology has ushered in a revolution in the automotive world. People are starting to prefer \say{self-driving} vehicles, not just for more comfortable driving, but also to keep drivers safer. HAR can significantly enhance the safety and performance of self-driving cars by enabling the car's AI system to recognize and respond accurately to human actions and behaviors on the road, inside and outside the car. HAR can help self-driving cars better understand the intentions of human drivers, pedestrians, and other road users, which can be difficult to predict without understanding their actions. By recognizing human actions, such as pedestrians crossing the road, cyclists moving on the road, and other vehicles changing lanes or making turns, the self-driving car can adjust its behavior and make appropriate decisions to avoid accidents. Thus, HAR can also help self-driving cars anticipate potential risks and take preventive measures to ensure the safety of road users.

\textbf{\textit{Video Surveillance:}}
The issue of security is fundamental to our daily lives. It keeps us physically safe, reducing the risk of falling victim to crime. Surveillance camera systems are not expensive these days. However, a traditional surveillance system needs human operators to watch it continuously. HAR plays a crucial role in improving the effectiveness and efficiency of video surveillance systems, ensuring public safety and security. It can significantly enhance video surveillance systems by enabling AI systems to detect, recognize, and respond to human actions in real-time. HAR can augment the traditional video surveillance processes by identifying potential threats or suspicious activities and aiding investigations. Law enforcement can analyze large volumes of video footage efficiently by automatically detecting and localizing these events. HAR can direct them to the precise time an event of interest takes place. HAR can detect and recognize abnormal human behaviors in safety-critical zones, such as train stations, airports, and concerts, alerting security personnel to take necessary action before catastrophe strikes promptly. 

\textbf{\textit{Shopping:}}
HAR is a powerful tool that can greatly enhance the shopping experience by giving retailers insights into their customers' behaviors and preferences. It can help retailers offer personalized shopping experiences by recognizing and interpreting customers' gestures and movements, such as reaching for a product or placing an item in their shopping cart, and using such data to recommend similar products or offer relevant discounts. It can also improve inventory management by identifying when a product is running low and alerting to restock the shelves quickly. Amazon Go was the first convenience store that used HAR technology for shopping, calling it the \say{Just walk-out}. Amazon established its first store in Washington, Seattle, in 2017, and as of late 2024, Amazon Go has 22 stores in Seattle, Chicago, San Francisco, and New York City. The Just Walk-out technology depends on human action recognition and prediction to analyze every move customers make using cameras installed in aisles and around the store \cite{al2022generating}. Customers can walk through the store, shop, and leave without waiting for checkout. The system recognizes the items a customer puts in pockets or bags and even the items taken from the shelf to read labels and return. Widespread adoption of this technology in stores will help save money by reducing theft and manpower costs. 

\textbf{\textit{Sports Motion Analysis:}}
Human action recognition is a game changer in sports analysis, providing valuable insights into athletes' performance and enhancing training programs. By analyzing athletes' movements and actions, HAR can help coaches identify strengths and weaknesses in the techniques, enabling the coaches to provide targeted feedback to improve player skills. For instance, HAR can recognize the proper form of a tennis movement \cite{alshami2023pose2trajectory} or a basketball shot and recognize deviations, allowing coaches to identify areas for improvement and develop customized training programs. HAR can also help sports analysts and commentators provide better insights during live broadcasts or post-game analysis. By recognizing and interpreting athletes' movements and actions, HAR can help explain the strategy behind specific plays or highlight exceptional performances. Moreover, HAR can enable the development of new metrics to evaluate athletes' performance, such as the speed and accuracy of their movements or the efficiency of their techniques. Today, many professional sports clubs use sports analysis to better understand players' potential successes. By incorporating HAR into their training and analysis programs, teams can gain a competitive edge by identifying critical areas for improvement and developing more effective training strategies. Today, Many professional sports clubs use sports analysis to better understand their players' potential better \cite{cui2020deep}.

\section{Deep Learning-Based Approaches for HAR} \label{sec:Deep learning approaches}
Human action recognition has experienced significant advances by leveraging the advent of deep learning in the mid-2010s \cite{krizhevsky2012imagenet}. Integrating deep learning concepts has become essential, from Single-stream and Two-stream networks, Motion Networks, and 3D Convolutional Networks to the recent Transformer-based approaches. This section delves into the notable and recent contributions, highlighting how recent network architectures have been utilized to enhance performance. We have organized the learning-based approaches into categories based on their architectural frameworks. Each category is thoroughly explored, providing a comprehensive background, detailed analysis, and (with the exception of single-stream networks) a discussion to facilitate a deeper understanding of the subject matter.

\subsection{Single Frame Convolutional Networks} \label{subsec:single stream networks}
Single-frame convolutional architecture is used to understand the contribution of static appearance to classification accuracy. The improved Dense Trajectory (iDT) \cite{wang2014action} method uses a deep CNN to recognize static appearance by combining motion and appearance features, which provide essential cues for understanding human actions from temporally untrimmed videos. The DeepVideo method uses a multi-resolution CNN architecture and a single 2D model on each video frame to capture connected temporal patterns to facilitate learning spatial-temporal features for video action recognition \cite{karpathy2014large}. The first is the context stream to model low-resolution images, and the second is the fovea stream to process the high-resolution center crops. Although it looks like a two-stream network, it is not really so. The streams in the DeepVideo method are for analyzing images at two levels of resolution, not for motion modeling to augment static modeling. Testing the DeepVideo method on the UCF101 dataset, one of the most popular datasets at the time, was 20.0\% less accurate than the iDT method.

\subsection{Two-stream Networks} \label{subsec:two-stream}
\subsubsection{Background}
The shortcomings of single-stream networks demonstrated that local convolutional operations are not enough to capture motion information in human action-oriented videos. Because videos have spatial and temporal dimensions, capturing only spatial information is ineffective for recognizing actions with multiple steps or long sequences, e.g., throwing a ball versus holding a ball, jumping versus squatting, and laying versus rolling. 

\begin{figure}[hbt!]
    \centering
    \includegraphics[width=15cm]{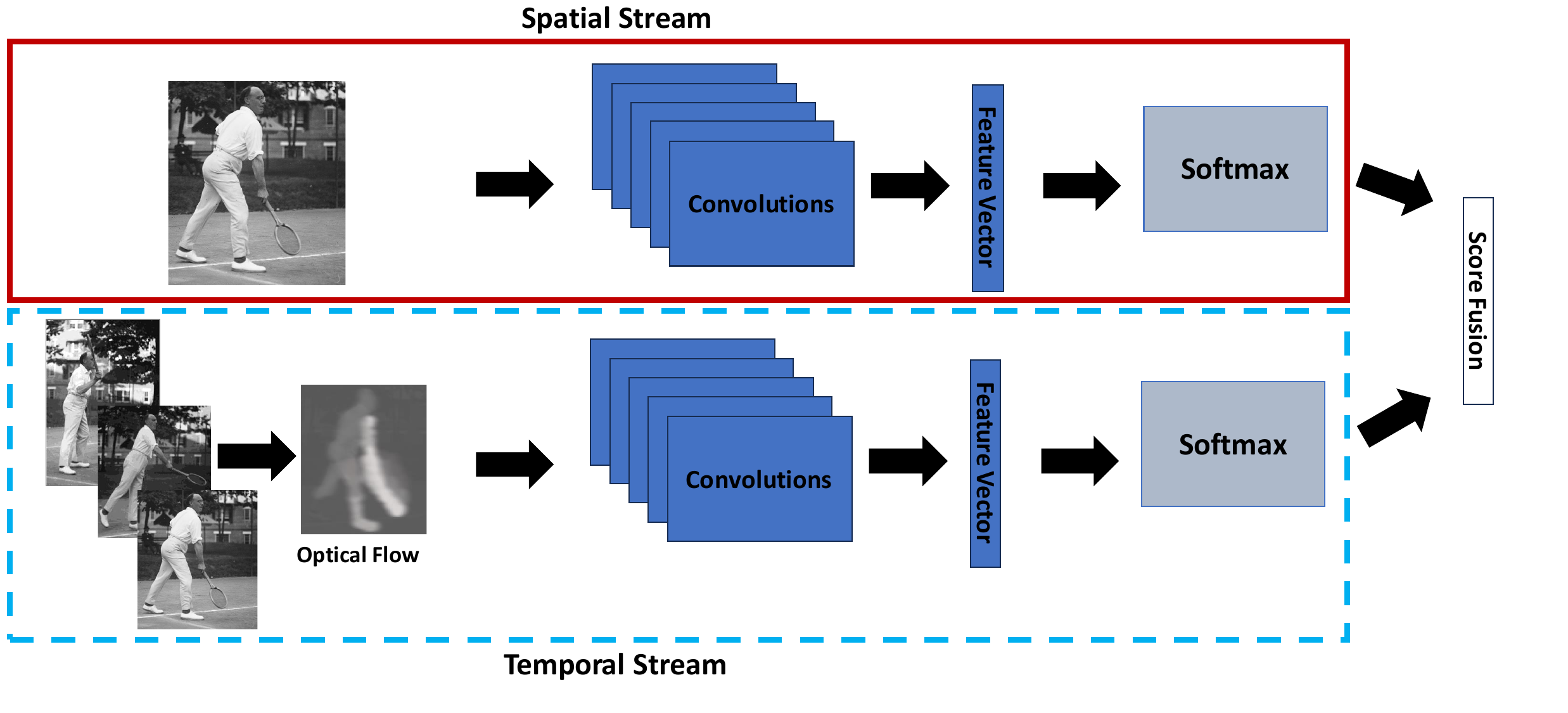}
    \caption{The two-stream architecture for video classification. The spatial stream (framed by a solid red border) illustrates a single image passing through a Convolutional Neural Network to a softmax layer. The temporal stream (delineated by a dashed blue border) shows stacked frames transforming into an optical flow map and passing through a Convolutional Neural Network and softmax layer. Both streams culminate in a joint diagram representing the fusion of their softmax scores into final class probabilities.}
    \label{fig:Two_Stream_Origin}
\end{figure}

The two-stream network architecture \cite{simonyan2014two} was designed to capture spatial and temporal information from video streams.
As shown in Figure~\ref{fig:Two_Stream_Origin}, the system used two Convolutional Neural Networks, one trained on single frames to capture spatial information and one expanded from prior work by Karpathy et al. \cite{karpathy2014large} by training on optical flows derived from sequences of frames. Each network outputs a classification vector of softmax scores fused for the final classification of the video segment. The two-stream architecture marks a major turning point for human action recognition, as previous solutions did not explicitly represent spatial and temporal dimensions. Accordingly, the system achieved state-of-the-art accuracy on the UCF-101 dataset. While the implementation from Simonyan et al.~\cite{simonyan2014two} is straightforward, the definition of two-stream networks for human action recognition from videos is broad. Generally, HAR systems that learn representations for at least spatial and temporal dimensions through distinct data paths are two-stream networks. The original work identified the importance of capturing spatial and temporal representations. It used Convolutional Neural Networks to learn them, but subsequent works have experimented with other network architectures, training regimens, additional network modules, and unsupervised learning.

\subsubsection{Extensions of the Two-stream Model}
Since the advent of two-stream networks, numerous works \cite{simonyan2014two,YaminGoing,SarabuDistinct,YanqinAction,ShengLearning,ChengHuman,Zhongwen,WenbingToward,tran2017two,feichtenhofer2016convolutional,XiaohangUnsupervised,gammulle2017two,Xiao_2022_CVPR} have improved upon the original architecture. Simonyan et al. \cite{simonyan2014two} used two Convolutional Neural Networks for spatial and temporal streams. Still, the state-of-the-art in deep neural networks is constantly evolving, and a diverse and ever-growing range of architectures can be used. Multiple two-stream human action recognition works have specifically experimented with different architectures to enrich the features used for classification.

Yamin et al. \cite{YaminGoing} proposed deepening the simple CNN architecture for both streams to enhance the quality of extracted features. Sarabu et al.~\cite{SarabuDistinct} replaced the simple CNN used in the seminal two-stream work with a ResNet \cite{he2016deep} for the spatial stream and an InceptionV2~\cite{ioffe2015batch} network intended to capture long-range information for the temporal stream. Yanqin et al. \cite{YanqinAction} modified the notion of two-stream networks by extracting features using long-term and short-term networks. They incorporated a 3D CNN for the long-term stream (C3D) and a VGG16 network \cite{simonyan2014very} for the short-term, fusing features before training an SVM \cite{hearst1998support} for classification. 

Drawing attention to their proficiency at time series and sequence-based problems, several works have proposed incorporating Recurrent Neural Networks (RNN) into the two-stream architecture for human action recognition in videos. Yu et al. \cite{yu2017novel}, motivated by the success of RNNs in adjacent recognition tasks and early applications of Long Short-Term Memory (LSTM) modules to action recognition problems, proposed a pooling input LSTM-based architecture. Both spatial and temporal streams were extracted using GoogLEnet \cite{szegedy2015going}, and features were pooled between frames before being fed into LSTM modules. This system aimed to capture long-term temporal information from videos, a quality that authors argued was not adequately addressed by prior RNN-based action recognition works. Yu et al. \cite{ShengLearning} further improved this system by replacing the LSTM modules with Pseudo Recurrent Residual Neural Network Modules. 
Authors drew inspiration from the ResNet architecture methodology \cite{he2016deep} and designed these modules with the intent to capture global long-term features of video clips. Similarly, Cheng et al. \cite{ChengHuman} added LSTM modules at the end of each stream and fused their features together, creating a joint optimization layer for classification. Zhongwen et al. \cite{Zhongwen} adapted the spatial stream to process multiple frames simultaneously using their proposed bidirectional Gated Recurrent Unit (BiGRU), which authors argue allows the network to learn coherent features of actions. An attention module called SimAM was also applied to the temporal stream to better capture long-term features.

Wang et al. \cite{wang2016temporal} soon after proposed the temporal segment network framework, which augmented the two-stream model by parsing multiple segments of a video and penalizing the consensus between segments. By learning simultaneously via segments rather than utilizing a whole video of sequential frames, the framework was designed to learn long-term temporal dependencies of labeled actions.

Li et al.~\cite{li2023fsformer} designed a two-stream Transformer network, which targets short-term actions in video segments, such as walking to a stove, igniting the gas, and heating a pan. Researchers argued that these short-term actions are key cues for longer action recognition tasks. This work adapted the SlowFast~\cite{feichtenhofer2019slowfast} mentality originally applied for learning 3D convolutions. Li et al.~\cite{li2023fsformer} proposed a short-term action differentiated attention module to link features from both slow and fast streams. The authors argued this module enables their system to determine positive and negative short-term actions, which correlate to the overall action recognition label.

While the standard two-stream architecture \cite{simonyan2014two} fused the final classification scores between spatial and temporal streams, some researchers have attempted to unify them beyond simple fusion and outside of recurrent strategies. Wenbing et al.~\cite{WenbingToward} was inspired by the infrequent yet crucial distribution of action cues in videos. They introduced a backpropagation approach that selectively propagates loss for specific segments of video clips. Their method employed KL divergence and introduced the MaxRule algorithm to determine the optimal segments for loss application. By only applying the loss at key points in videos, spatial and temporal streams tended to learn the same cues from different dimensions. Tran et al.~\cite{tran2017two} linked the first convolutional layer of both streams, the early fusion seemingly forcing the networks to learn similar cues. Feichtenhofer et al.~\cite{feichtenhofer2016convolutional} experimented with feature fusion at various stages of the network, learning a pixel-to-pixel correspondence between the two streams.

Other researchers have contributed exciting to the two-stream architecture and incorporated learning strategies beyond supervised learning. Xiaohang et al.~\cite{XiaohangUnsupervised} modified the two-stream model to use a two-stage network for the temporal stream, which was trained with unsupervised learning. Their two-stage temporal stream was trained using brightness and edge-based losses. Gammulle et al.~\cite{gammulle2017two} adapted two-stream networks by adding an LSTM module on top of the convolutional streams and applied unsupervised learning during training. Xiao et al.~\cite{Xiao_2022_CVPR} used semi-supervised learning and cross-modal knowledge distillation to train a two-stream network model. Their method achieved higher accuracy on Kinetics-400, UCF-101, and HMDB-51 datasets than state-of-the-art competitors at the time.

Recent developments in Transformer and skeleton-based models have inspired various modifications to two-stream action recognition systems. Shi et al.~\cite{shi2023novel} viewed the combination of spatial and skeletal information as a multi-modal learning problem. They proposed a two-stream transformer model that incorporates the SlowFast mentality \cite{feichtenhofer2019slowfast}, but uses skeletal heatmaps for the high-frequency stream and normal RGB frames for the low frequency. They were motivated by prior works demonstrating the complimentary use of Skeletal and RGB extracted features but focused on incorporating them effectively using transformer networks. The authors recognized that skeletal features were already refined compared to RGB features; accordingly, they adjusted the skeletal Transformer to use fewer attention layers, increasing accuracy and precision. Table~\ref{Tab:TwoStream-based-models} summarizes the two-stream networks for HAR discussed in this subsection.

\subsubsection{Discussion}
Of all the extensions to the two-stream model mentioned in this survey, one was overwhelmingly impactful to the two-stream methodology. The seminal two-stream architecture, introduced by Simonyan et al. \cite{simonyan2014two}, used softmax score fusion to combine representations for classification from both streams. This simple and intuitive fusion concept had demonstrable value, as evidenced by the system's performance on standard benchmarks. While softmax score fusion served as an excellent initial method for linking spatial and temporal dimensions, other researchers \cite{feichtenhofer2016convolutional, tran2017two} explored advanced fusion schemes. Feichtenhofer et al. \cite{feichtenhofer2016convolutional}, in particular, investigated several essential questions regarding two-stream fusion:
\begin{itemize}
\item How should spatial and temporal networks be fused with respect to space?
\item Where should two networks be fused?
\item How can spatial and temporal networks be fused temporally?
\end{itemize}

\begin{table}[h]
\centering
\small 
\caption{This label pertains to Figure~\ref{fig:HARVennDiagrame}. Classified as a two-stream Network (T-SNs) based on the model architecture involved. For all categories, refer to Figure~\ref{fig:HARVennDiagrame}. \\ \textbf{\textit{Acronyms:}} Two-Stream Networks (T-SNs), Motion (M), 3D Convolutional (3DC), Graph Convolutional Networks (GCNs), Hybrid (H), and Transformer (T).}
\label{Tab:TwoStream-based-models}
\begin{tabular}{p{2.5cm} p{5cm} p{5cm}} 
\hline
\textbf{\textit{T-SNs}} & Model architectures & Paper citations \\
\hline
T-SN.a & T-SN, M, and H Networks & \cite{simonyan2014two,YaminGoing,SarabuDistinct,ShengLearning,WenbingToward,feichtenhofer2016convolutional,XiaohangUnsupervised,gammulle2017two,Xiao_2022_CVPR,yu2017novel,wang2016temporal,yu2019learning} \\
T-SN.b & T-SN, M, 3DC, and H Networks & \cite{YanqinAction} \\
T-SN.c & T-SN, M, H and Attention Networks & \cite{ChengHuman,Zhongwen,tran2017two} \\
T-SN.d & T-SN, M, H, T and attention Networks & \cite{li2023fsformer,shi2023novel} \\
\hline
\end{tabular}
\end{table}

The authors addressed these questions by examining and evaluating several fusion methods against each other and state-of-the-art action recognition systems. Through their examination, authors arrived at their proposed fusion model, spatiotemporal fusion, which creates a correspondence between spatial and temporal streams. By learning correspondence between the two streams, both streams are encouraged to learn similar cues from different representations. This work represents a major milestone in two-stream architectures as it advanced the original work's \cite{simonyan2014two} findings from ``we need information representing spatial and temporal dimensions'' to ``How can we best unify spatial and temporal representations''.

With the recent advancement of Vision Transformer models \cite{dosovitskiy2020image}, multiple works \cite{li2023fsformer, shi2023novel, Zhongwen} have applied attention mechanisms to two-stream and multi-stream architectures to learn such a unified representation.

\subsection{3D Convolutional Networks} \label{subsec:3D CNNs}
\subsubsection{Background}
3D convolutional neural networks (CNNs) were designed to tackle the same problem two-stream networks sought to solve: to incorporate motion information into HAR algorithms. Convolutions have been used for recognition tasks since Fukishima~\cite{fukushima1980neocognitron} proposed them in the seminal work called Neocognitron. However, as CNNs gained popularity, their application to human action recognition was limited to the use of two-dimensional kernels, which alone can only iterate over pixels in a single image. Ji et al.~\cite{ji20123d} recognized that while 2D convolutions have excellent performance at image recognition tasks, they cannot capture the temporal motion information conveyed in videos. To this end, the authors proposed 3D CNNs, which extend the traditional 2D convolutional kernel to process features across multiple frames. Authors validated the effectiveness of their networks on TRECVID \cite{yang2009detecting} and KTH \cite{schuldt2004recognizing} data, which limited the paper's impact as these datasets have very few classes, and in the case of TRECVID, an overwhelming amount of negative or unclassified samples. Training on limited or imbalanced classes is problematic for supervised learning schemes, and systems that succeed at these tasks may not generalize well to large datasets or real-world settings.

Tran et al.~\cite{tran2015learning} investigated the applicability of 3D CNNs to large-scale recognition tasks and proposed a new 3D CNN, called C3D, based on their findings. This work majorly advanced the use of 3D CNNs for action recognition by applying pooling between 3D convolutional layers and exploring kernel depth in the temporal dimension. These researchers used C3D to learn spatiotemporal features from videos and evaluated the effectiveness of these models on action recognition (using Sports1M and UCF101 datasets), action similarity labeling, scene classification, and object recognition tasks. This work took the first major step towards showing that 3D CNNs can be widely applicable to video recognition tasks.

\subsubsection{Extensions of 3D CNNs}
The 3D CNN models hold theoretical promise to capture spatial-temporal features from videos effectively. However, 3D convolutions are more complex than their 2D counterparts and present enormous difficulty to train. 
Carreira and Zisserman~\cite{carreira2017quo} proposed two-stream Inflated 3D ConvNets (I3D). Recognizing the burdensome complexity of training 3D convolutional kernels and the abundance of ImageNet \cite{deng2009imagenet} pre-trained models, researchers devised an inflation and bootstrapping scheme to convert 2D convolutions to 3D and leverage pre-training model on ImageNet and Kinetics datasets. Interestingly, researchers used a two-stream configuration for I3D. As discussed in subsection \ref{subsec:two-stream}, two-stream networks were meant to address capturing spatial and temporal dimensions of action recognition, the same problem 3D convolutions targeted. Looking back, this design choice may seem redundant, but at the time, they showed significant improvements against Two-Stream and 3D-ConvNet systems. Authors demonstrated the effectiveness of \say{Two-Stream inflated 3D CNNs} by evaluating their performance in a transfer-learning task where networks were initially trained on the Kinetics dataset, followed by another period of training and testing on UCF-101 and HMDB-51.

Qiu et al.~\cite{qiu2017learning} proposed pseudo3D (P3D) residual networks. The authors circumvented the complexity of 3D convolutional kernels by designing bottleneck building blocks that link 2D convolutions with a 1D temporal convolution. In addition to reducing complexity, this enabled researchers to initialize the 2D convolutions with pre-trained weights. P3D held significant promise because the building blocks proposed could be arranged in a variety of network architectures, similar to traditional ResNets. Notably, researchers validated P3D on the Sports1-M dataset. Later, Wang et al.~\cite{wang2018non} proposed non-local blocks for neural networks, which can be inserted in between the residual connections of traditional neural networks. Researchers were trying to capture long-term dependencies in video data, a natural complement to action recognition. Non-local blocks were added to I3D to investigate their effect on 3D convolutions. Researchers found the non-local I3D model had superior performance on the Kinetics-400 dataset.

Zolfaghari et al.~\cite{zolfaghari2018eco} targeted the problem of video understanding and specifically utilized 3D convolutions for capturing temporal data. However, they did not start at the image level as prior works did, nor did they focus on only sequential frames. In their proposed systems, ECO Lite and ECO Full, a 3D CNN model was used to process feature maps extracted by traditional 2D CNNs. 2D CNNs randomly sampled frames in video segments and fed the resulting features into part of a 3D ResNet 18. In their ECO Full system, they also processed sequential frames with a 2D Inception-4a network to aid in classifying short actions. While I3D had superior results on UCF101 and HMDB51, ECO Full achieved superior performance on the Kinetics-400 dataset. 
 
Feichtenhofer et al.~\cite{feichtenhofer2019slowfast} introduced SlowFast networks for video recognition. They recognized that not all spatial-temporal relations are alike, and a system that captures temporal information very granularly will not always be suited for actions that can be recognized by sparse frame sampling. In a similar but distinct method compared to ECO~\cite{zolfaghari2018eco}, SlowFast networks use one stream to capture temporal data at high rates and link outputs from another stream, which samples videos at low frame rates. Both streams use 3D convolutional but fuse features through lateral connections after residual blocks, a technique pioneered by two-stream networks \cite{feichtenhofer2016convolutional}. Authors also proposed several variations of their SlowFast network by incorporating non-local blocks \cite{wang2018non}. Impressively, authors were able to train these SlowFast networks from scratch, a feat that most prior works sought to avoid. Their systems showed superior performance on the Kinetics-400 and Kinetics-600 datasets.

The next major advancement came in 2020, also by Feichtenhofer et al., with X3D~\cite{feichtenhofer2020x3d} setting a new bar for 3D CNNs. The work first proposed X2D network architecture, a lightweight 2D network motivated by recent advances in mobile neural networks. X2D is then expanded gradually into a 3D architecture X3D by modifying depth, width, temporal sampling, and fastness (temporal resolution). Because the network's complexity is gradually changing, X3D was able to be trained to be configurable and suitable for a variety of computational resources. These innovations enabled X3D to match the state-of-the-art SlowFast network on Kinetics-400, with a drastic reduction in parameters and computational expense.

More recently, Ou et al. \cite{ou20233d} improved the reasoning of temporal dependency between entities and objects in action recognition systems and proposed 3D Deformable Convolutional Temporal Reasoning (DCTR) neural networks. These researchers targeted interaction information derived from entity-object interactions and introduced temporal and spatial modeling modules that can be added to existing 3D convolutional networks. The researchers validated their work on the UCF101 and HMDB51 datasets and showed consistent performance improvements when applying their modules.

\subsubsection{Discussion}
The 3D CNNs models were slow to be adapted for HAR since they were hard to train and had significantly more parameters than their 2D counterparts. Combined with the historically limited size of action recognition datasets, this posed a serious problem for implementing 3D Convolutional networks. X3D revolutionized the use of 3D convolutions in HAR and made training feasible, achieving reasonable accuracy in comparison to other HAR methods outside of 3D convolutions. Ou et al.~\cite{ou20233d} began exploring 3D Convolutions from a temporal reasoning standpoint, a perspective that has become popular among recent motion modules.

\begin{table}[h]
\centering
\small 
\caption{This label pertains to Figure~\ref{fig:HARVennDiagrame}, classified as 3D Convolutional Networks (3DCNs) based on the model architecture involved. For a detailed view of all categories, refer to Figure~\ref{fig:HARVennDiagrame}. \\ \textbf{\textit{Acronyms:}} Two-Stream Networks (T-SNs), Motion (M), 3D Convolutional (3DC), Hybrid (H), and Transformer (T).}
\label{Tab:3DCN-models}
\begin{tabular}{p{2cm} p{6cm} p{5cm}} 
\hline
\textbf{\textit{3DCN}} & Model architectures involved & Paper citations \\
\hline
3DCN.a & 3DCN Networks & \cite{ji20123d, tran2015learning, qiu2017learning, wang2018non, zolfaghari2018eco, feichtenhofer2020x3d} \\
3DCN.b & 3DCN, HN, MN, T-SN Networks & \cite{carreira2017quo} \\
3DCN.c & 3DCN, H Networks & \cite{ou20233d} \\
\hline
\end{tabular}
\end{table}

As additional temporal modules continue to develop, their use with 3D convolutional networks will likely follow. Though X3D has not been combined with temporal reasoning modules, this remains an area for future work. 

\subsection{Graph Convolutional Neural Networks} \label{subsec:GCNs}
\subsubsection{Background}
Originating from the University of Wollongong, Australia, the groundbreaking Graph Neural Networks (GNNs) model was introduced in 2009 as a solution to the intricate issue of graph learning. This innovative model excels at isolating features from graphs of arbitrary structure and adeptly transposing graph data into a low-dimensional space. Remarkably, it does so while preserving both structure and property information to an exceptional degree. The model paves the way for building a training-specific neural network, further enhancing its versatility and utility\cite{scarselli2008graph}.

GNNs can be classified into two main categories: Spectral GNNs and Spatial GNNs \cite{li2018adaptive}. Spectral GNNs operate by processing input graph signals through an array of learned filters strategically positioned within the graph Fourier domain. This sophisticated approach permits the nuanced manipulation of spectral properties. On the other hand, Spatial GNNs employ a distinctly different methodology, implementing layer-wise updates for each node. This process can be broken down into three key steps. Initially, neighbors are selected based on a designated neighborhood function, such as adjacent nodes. Subsequently, these chosen features are combined with the node's own, using an aggregation function like mean pooling, effectively creating a feature amalgamation. Finally, this merged entity undergoes an activation transformation, often by deploying a Multi-layer Perceptron (MLP), leading to a highly refined feature representation.

The domain of GNNs encompasses several distinct but interconnected categories. This includes the Graph Convolutional Networks (GCNs)~\cite{kipf2016semi}, which employ a form of convolution operation adapted for graph data to capture local spatial information effectively, and the Graph Attention Networks (GATs)~\cite{velivckovic2017graph}, which utilize attention mechanisms to weigh the contributions from neighboring nodes. The Graph Isomorphism Networks (GINs)~\cite{xu2018powerful} embed a form of the Weisfeiler-Lehman graph isomorphism test in the network architecture, enabling greater expressiveness. The GraphSAGE model \cite{liu2020graphsage} offers a unique method for learning node representations in large graphs by sampling and aggregating features from a node's local neighborhood. Lastly, the 3D VSG \cite{looper20223d} model expands the application of GNNs to three-dimensional data, enhancing the ability to handle more complex spatial structures. Each of these categories has significantly contributed to the progression and capabilities of GNNs.

\begin{figure}[hbt!]
    \centering
    \includegraphics[width=7cm]{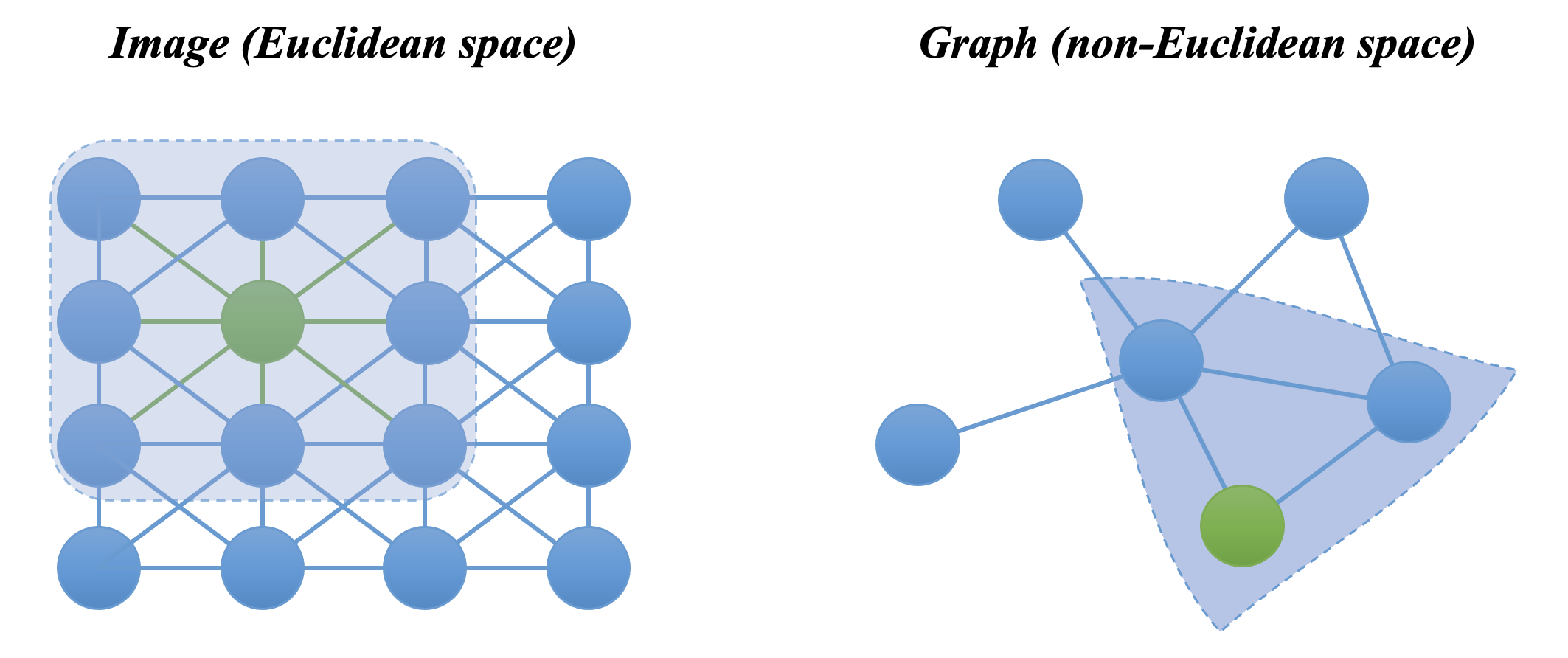}
    \caption{Difference between GCN and CNN.}
    \label{fig:GNC_CNN}
\end{figure}

GCNs are the most commonly used methods for skeleton-based action recognition due to their unparalleled efficacy in modeling non-Euclidean data, unlike the CNN, which performs convolution on a Euclidean space (e.g., images) \cite{peng2020learning}, as shown in Figure~\ref{fig:GNC_CNN}. GCNs allow capturing both spatial and temporal information in a skeleton sequence by using spatial convolutions to capture structural information of the skeleton and temporal convolutions to understand the dynamics of the action. It has achieved remarkable performance and reached state-of-the-art results on benchmarks.

\begin{figure}[hbt!]
    \centering
    \includegraphics[width=7cm]{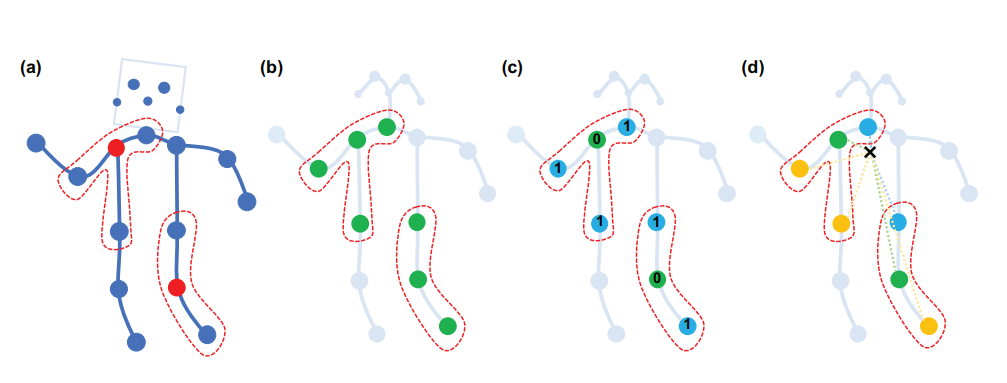}
    \caption{The diagram shows different strategies for structuring convolution operations. It includes: (a) An input skeleton frame with body joints (blue dots) and filter receptive fields (red dashed circles). (b) A uniform labeling strategy with all nodes in a neighborhood sharing the same label (green). (c) A distance partitioning strategy separates the root node (green) and its neighbors (blue). (d) Spatial configuration partitioning where nodes are classified based on their proximity to the skeletal gravity center (black cross), with closer nodes (blue) and farther nodes (yellow) relative to the root node (green) \cite{yan2018spatial}.} \label{fig:strategies for convolution operations}
\end{figure}

The human skeleton is interpreted as a specialized graph data structure within this framework, denoted as $G = (V,E)$. The vertices, represented by $V = {v_1, ..., v_n }$, signify the joints of the human body, amounting to a total of $N$ nodes. The set $E$, on the other hand, encapsulates the bones, symbolized as the connecting edges between the nodes. This unique arrangement embeds a real-world human body's complex interconnections and constraints into a structured form; see Figure~\ref{fig:strategies for convolution operations} for more details. 

Through this representation, the human skeleton graph provides a nuanced, high-fidelity depiction of human body structures that can be efficiently utilized for various machine-learning tasks. The graph's model is bifurcated into two branches - spatial and temporal. The spatial branch aims to mine structural information from the human body, tapping into the intricate details embedded within the nodes or joints, denoted as $V$. In contrast, the temporal branch focuses on the sequential cues derived from adjacent frames, symbolizing the dynamic movements of the bones, represented by the edges, $E$. By leveraging this dual methodology, the model thoroughly comprehends the dynamism inherent in the human body's movements. This, in turn, significantly bolsters the effectiveness and precision of tasks related to human action recognition. 

\subsubsection{GCNs and Extensions of GCNs for HAR}
Some recent works have shown remarkable performance using GCNs in skeleton-based action recognition  \cite{peng2020learning,yan2018spatial,li2019actional,liu2020disentangling}. Yan et al. \cite{yan2018spatial} created ST-GCN, the first graph-based neural network model that can dynamically model skeletons by managing temporal and structural relations. This model captures the dynamism of human movements across time and uses a classifier for precise action categorization, enhancing comprehension of human motion. Li et al.~\cite{li2019actional} proposed an encoder-decoder structure, the A-link inference module, to capture action-specific latent dependencies and extend skeleton graphs for higher-order dependencies. Liu et al.~\cite{liu2020disentangling} introduced a robust feature extractor by proposing a multi-scale aggregation scheme for unbiased joint relationship modeling and a new module, G3D, for unobstructed cross-space-time information flow. G3D, a more advanced version of 3D CNNs, is a unified spatial-temporal graph convolutional operator designed to enhance the modeling of long-range joint relationships and capture complex spatial-temporal dependencies in skeleton-based action recognition.

\begin{figure}[hbt!]
    \centering
    \includegraphics[width=14cm]{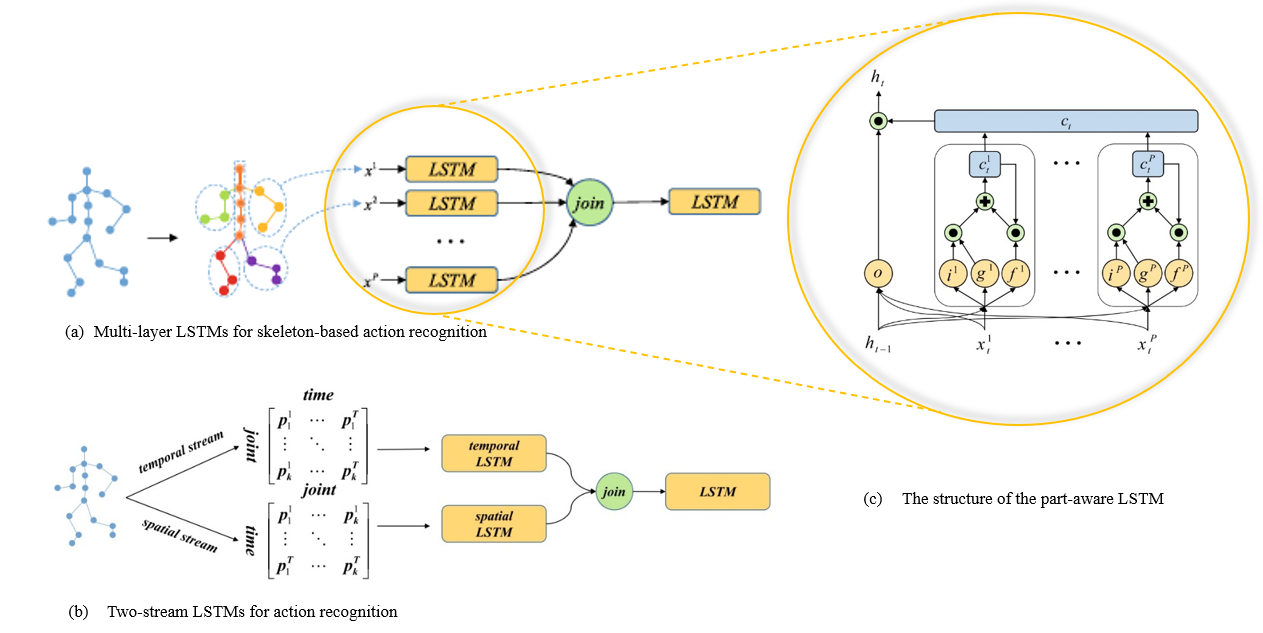}
    \caption{This figure shows how an LSTM splits the human body skeleton into multiple parts and extracts the spatial-temporal features by creating a GCN for each part and then merging them. This Figure is adopted from figures in~\cite{qin2020skeleton}.}
    \label{fig:LSTM for skeleton based HAR}
\end{figure}

Approaching from a different perspective, numerous researchers extol the virtues of integrating LSTMs with GCNs, thereby significantly amplifying the results in action recognition tasks that are based on skeletal structure. 
In an intriguing proposition, Si et al. \cite{si2019attention} introduced an innovative model known as the Attention Enhanced Graph Convolution LSTM Network (AGC-LSTM). This distinctive model is proficient in not only discerning distinctive features in spatial configurations and temporal dynamics but also in investigating the synergistic interplay between the spatial and temporal domains. In a notable study, Qin et al.~\cite{qin2020skeleton} proposed an innovative part-aware LSTM model, which involves segmenting the human body skeleton into multiple parts, extracting the spatial-temporal features, and creating separate GCNs for each part instead of one for the entire body. Each part corresponds to an LSTM network, and the outputs from all these networks are integrated, as illustrated in Figure~\ref{fig:LSTM for skeleton based HAR}. This approach showcases the potential of LSTM in the realm of action recognition, particularly when paired with a part-wise analysis of the human body skeleton.

Xiang et al.~\cite{xiang2022language} introduced an intriguing concept called Language Supervised Training (LST). The LST approach leverages a large-scale language model functioning as a knowledge engine, providing textual descriptions that illustrate the movement of various body parts during an action. Their proposal further includes a multi-modal training scheme, which capitalizes on a text encoder to generate a feature vector for differing body parts, subsequently guiding the skeleton encoder for advanced action representation learning. When building the model's encoder, the authors employ GCN as the backbone network within the LST framework. The skeleton encoder, crafted intricately, comprises multiple GC-MTC blocks, with each block housing a Graph Convolution layer along with a Multiscale Temporal Convolution module. 

Drawing parallels with the LST approach, Xu et al. \cite{xu2023language} proposed an innovative technique, the Language-Assisted Graph Convolution Network (LA-GCN), using the expansive knowledge provided by large language models. Yet, it distinguishes itself by mapping the language models into a predefined global relationship and category relationship taxonomies among nodes. The global relationship feature facilitates the creation of novel bone representations, emphasizing the crucial data housed within each node, while the category relationship element reflects category-prior knowledge ingrained in human neural networks. This knowledge is encoded via the PC-AC module, offering further guidance for distinctive feature learning across various classes. Moreover, in a bid to bolster the efficiency of information transfer in topological modeling, the authors introduce a concept known as multi-hop attention graph convolution.

There has been a recent surge in research efforts aiming to enhance the capabilities of GCNs \cite{lee2022hierarchically, duan2022dg, chi2022infogcn, huang2023graph, rahevar2023spatial}. Among these advances, Lee et al.~\cite{lee2022hierarchically} introduced a novel architecture known as the Hierarchically Decomposed Graph Convolution Network (HD-GCN)  utilizing a unique hierarchically decomposed graph (HD-Graph) to augment the extraction of edges within the graph. The HD-GCN ingeniously decomposes each joint node into several subsets to extract structurally adjacent and distant principal edges, assembling these within an HD-Graph that parallels the semantic spaces of a human skeleton. Duan et al.~\cite{duan2022dg}, on the other hand, proposed the Dynamic Group Spatio-Temporal GCN (DG-STGCN). Comprising two integral modules, the DG-GCN and the DG-TCN, this approach exploits learned affinity matrices to grasp dynamic graphical structures without the need for a predetermined one. The DG-TCN performs group-wise temporal convolutions with fluctuating receptive fields and incorporates a dynamic joint-skeleton fusion module for adaptive multi-level temporal modeling. While, the DG-GCN employs affinity matrices that are learned to grasp the changing structures of graphs, moving away from the dependency on pre-defined structures. On the other hand, DG-TCN executes temporal convolutions across groups using receptive fields of different sizes and integrates a dynamic fusion module for joint-skeleton, which allows for flexible multi-level temporal analysis.

InfoGCN~\cite{chi2022infogcn} is a learning framework for HAR that employs an information bottleneck-based learning objective to coax the model into acquiring informative compact latent representations. It further introduces an attention-based graph convolution that captures the intrinsic topology of human action, providing distinctive information vital for action classification. Huang et al.~\cite{huang2023graph} proposed the SkeletonGCL model, engineered to explore the global context across all sequences. Specifically, SkeletonGCL unites graph learning across sequences by compelling graphs to be class-discriminative. Lastly, Raheva et al.~\cite{rahevar2023spatial} presented a spatial-temporal dynamic graph attention network (ST-DGAT) designed to learn the spatial-temporal patterns of skeleton sequences. They refine the sequence of weighted vector operations in GAT to foster dynamic graph attention, achieving a global approximate attention function and rendering it unequivocally superior to the traditional GAT.

Trivedi et al.~\cite{trivedi2022psumnet} introduced a novel part-based stream processing approach to achieve more detailed and specialized representations for actions that involve specific joint subsets. The input skeleton undergoes processing by the multi-modality data generator, producing data related to joints, bones, joint velocity, and bone velocity. This multi-modal data is subsequently processed by the spatio-temporal relational module, which then undergoes global average pooling and a fully connected layer. The data also flows through the spatio-temporal relational block, further refining the process via the spatial attention map generator. Zhou et al.~\cite{zhou2023learning} proposed an innovative auxiliary feature refinement head that employs spatial-temporal decoupling combined with contrastive feature refinement to derive more discriminative representations of skeletons. This innovation addresses challenges inherent in using skeleton data, particularly the inadvertent omission of vital cues such as related items.

Hu et al.~\cite{hu2022spatial} proposed an STGAT framework that represents a significant advance in action recognition, particularly in its ability to process complex action sequences by addressing both long-term and short-term temporal dependencies. This method has successfully improved the accuracy of action classification, especially in distinguishing similar actions, which was a notable challenge for previous techniques. The Temporal-Channel Aggregation Graph Convolutional Networks (TCA-GCN)~\cite{wang2022skeleton} is designed for skeleton-based action recognition, dynamically learning and efficiently aggregating spatial and temporal topologies across various temporal and channel dimensions. It employs a Temporal Aggregation module for learning temporal features and a Channel Aggregation module to effectively merge spatial and temporal dynamic topological features. 

Liu et al.~\cite{liu2023temporal} introduced the Temporal Decoupling Graph Convolutional Network (TD-GCN) model, innovatively overcoming the constraints of traditional skeleton-based gesture recognition methods. Unlike previous models that rely on a uniform adjacency matrix for all frames, TD-GCN utilizes distinct, channel-specific, and temporal-specific adjacency matrices. This tailored approach significantly enhances the capture of spatiotemporal relationships among skeleton joints. The effectiveness of TD-GCN is demonstrated through its exceptional performance on challenging gesture recognition datasets, including SHREC’17 Track and DHG-14/28, establishing a pioneering application of temporal-dependent adjacency matrices in this field.

\subsubsection{Discussion}

Graph Convolutional Networks (GCNs) have demonstrated remarkable proficiency in capturing spatial information by employing convolutional neural networks and capturing temporal dynamics by integrating either optical flow or 3D convolutional networks, which is pivotal for interpreting and understanding action dynamics in non-Euclidean data domains \cite{zhou2018mict}. The GCNs have the intrinsic capability for concurrently modeling the spatial-temporal body skeleton over the length of the video sequence. GCNs adeptly model the spatial-temporal dynamics of body skeletons in video sequences, capturing complex motion patterns efficiently \cite{yan2018spatial}. 
The GCN-based action recognition model can classify based on the different model architectures that we discussed in the previous section. The first method is Spatio-Temporal Graph Convolution Network (ST-GNC) \cite{yan2018spatial}\cite{duan2022dg}\cite{huang2023graph}\cite{zhou2023learning}. ST-GCN interprets video skeletons through a spatio-temporal graph, using in-frame and between-frame edges for spatial and temporal details. Drawing from 2D convolution principles for its sampling and weighting functions, ST-GCN maintains consistent feature map sizes, unlike traditional CNNs. This innovation has paved the way for developments in STGR, PB-GCN, and ST-Graph sparsification.

The second method, Recurrent-Attention GCN, integrates either recurrent networks or attention mechanisms into its architecture \cite{si2019attention}\cite{qin2020skeleton}\cite{xu2023language}\cite{chi2022infogcn}\cite{rahevar2023spatial}\cite{wang2022skeleton}. This approach utilizes attention networks to highlight key areas of the body skeleton, typically employing recurrent architectures like LSTM or Transformer models to enhance focus on salient features.

The third method is encoder-decoder GCN \cite{li2019actional}\cite{ghosh2020stacked}\cite{li2020dynamic}, which represents an unsupervised learning strategy for transforming nodes or entire graphs into a compact latent vector space, effectively achieving a sophisticated network embedding. Following the encoding phase, the method reconstructs the original graphs from their latent representations. This process simplifies the complex structure of body skeletons into an inherently low-rank format via a graph encoder and enables the decoder network to generate low-rank graphs with potent features. These attributes are particularly beneficial for enhancing the performance of GCN-based action recognition systems, offering a robust framework for accurately identifying and analyzing movements.

The fourth method is a two-stream GCN approach, ingeniously designed to capture the complementary aspects of body skeletons, specifically focusing on both joints and bones \cite{shi2019two}\cite{li2020edge}\cite{luo2019improving}\cite{shi2019skeleton}\cite{tang2020graph}. It leverages the power of a two-stream network using both Spatial and Temporal features and the graph convolutional network to recognize human actions. Some other works propose zero-shot action recognition via a two-stream graph convolutional network \cite{gao2019know}\cite{gao2020learning}.
\begin{table}[h]
\centering
\small 
\caption{This label pertains to Figure~\ref{fig:HARVennDiagrame}, classified as Graph Convolutional Networks (GCNs) based on the model architecture involved. For a detailed view of all categories, refer to Figure~\ref{fig:HARVennDiagrame}. \\ \textbf{\textit{Acronyms:}} Two-Stream Networks (T-SNs), Motion (M), 3D Convolutional (3DC), Graph Convolutional Networks (GCNs), Hybrid (H), and Transformer (T).}
\label{Tab:GCNs-based-models}
\begin{tabular}{p{2.5cm} p{6cm} p{5cm}} 
\hline
\textbf{\textit{GCNs}} & Network Model(s) Used & Paper citations \\
\hline
GCNs.a & GCN Networks & \cite{yan2018spatial, li2019actional, liu2020disentangling, qin2020skeleton, huang2023graph, trivedi2022psumnet, zhou2023learning} \\
GCNs.b & GCN Networks \& Attention & \cite{si2019attention, lee2022hierarchically, rahevar2023spatial} \\
GCNs.c & GCN, Hybrid Networks & \cite{xiang2022language, duan2022dg} \\
GCNs.d & GCN, Hybrid, Attention Networks & \cite{xu2023language, chi2022infogcn} \\
\hline
\end{tabular}
\end{table}

The last method is the miscellaneous GCN \cite{trivedi2022psumnet}\cite{tang2018deep}\cite{li2020skeleton}. In this categorization, the proposed methods are quite naive and in the developing stage for GCN-based action recognition. 

In this section, we have reviewed the advancements of different GCN-based human action recognition models utilizing different mythologies. The evaluation of different GCN-based models is mentioned in the Subsection~\ref{sec:Holistic_Discussion}.

\subsection{Motion Models} \label{subsec:Motion models}
\subsubsection{Background}

Motion, often described as temporal representation, has been a critical component in HAR research from its early stages, providing key insights~\cite{aggarwal1997human}. As mentioned in previous sections, many modern HAR systems have relied on optical flow-based techniques to capture and recognize motion patterns across sequences of frames. Optical flow, first introduced by Horn et al.\cite{horn1981determining}, became a cornerstone of video recognition, evolving through extensive research. With the rise of deep learning, strategies for representing temporal information have emerged. For instance, the TF-Blender framework leverages temporal relations between video frames to improve object detection by aggregating features from neighboring frames, leading to more robust feature representations and improved detection accuracy\cite{cui2021tf}. Liqi et al.~\cite{yan2022video} propose a novel global-local encoder that exploits rich temporal representation for video captioning for human actions. Another " TransFlow " method introduces a pure transformer approach for optical flow estimation, leveraging spatial self-attention and cross-attention mechanisms to capture global dependencies between adjacent frames~\cite{lu2023transflow}. The method improves accuracy in challenging scenarios like occlusion and motion blur. Modern HAR systems have increasingly moved away from strict reliance on temporal representation, dramatically impacting the field and offering more efficient and robust solutions for HAR.

\subsubsection{Extensions to Modern Motion Models for HAR}
Researchers have proposed many novel techniques to capture motion or temporal relationships within videos effectively. In 2015, Wang et al. proposed trajectory pooled deep convolutional descriptors (TDD) \cite{wang2015action}. The system relied on point tracking and extracted feature maps from CNNs to learn a trajectory-constrained pooling function. The output of these TDDs is then used for prediction. While innovative, this method still relies on costly optical flow calculations for effective point tracking.

In 2016, a method termed `Rank Pooling' was introduced, aimed at learning the evolution of appearances over time~\cite{fernando2016rank}. This work, rooted in unsupervised learning, proposed a temporal pooling mechanism that is fit through a ranking function to reorder jumbled video frames sequentially. Video representations are then extracted through a series of fit models, and an SVM is used to classify them. One limitation of this work as designed is that it targets video-wide representations rather than segment-based action recognition. The distinction is that video-wide representations assume a video contains one action label, whereas segment-based action recognition assumes a video may contain more than one action. Despite its initial incarnation not being a deep learning approach, several subsequent methods have incorporated Rank Pooling as a differentiable layer within deep learning frameworks. This innovative idea honed in on the significance of temporal order and demonstrated its value for HAR.

In the same year, another approach named `Shuffle and Learn'~\cite{misra2016shuffle}, was proposed for analyzing video data under a similar framework. This technique utilized deep learning, specifically three instances of AlexNet trained end-to-end, to extract features from three frames within a video and penalize the network if frames were predicted in the wrong order. The design of this technique allowed researchers to employ unsupervised learning by sampling frames from the UCF101 video dataset. The objective was to target human poses in these videos rather than perform object or scene recognition.

Three years later, the hidden two-stream CNN approach was proposed by Zhou et al. \cite{zhu2019hidden}. The authors avoided the costly optical flow computation by using a deep neural network (DNN) to estimate optical flow.
Their network architecture (MotionNet) estimates optical flow from raw video frames to be combined with other DNNs. The paper showcases a two-stream system, with MotionNet preceding the temporal stream’s CNN.
This is interesting because while the paper proposes a two-stream system, its main focus is on estimating optical flow on demand.
Oddly enough, authors test the system with I3D \cite{carreira2017quo}, which inflates 2D network kernels to 3D but does not adopt the same strategy for MotionNet. Conceptually, processing sequential frames simultaneously through MotionNet seems like a natural way to learn optical flow patterns.

A major milestone was struck when Zhou et al.~\cite{zhou2018temporal} proposed the Temporal Relation Network (TRN). They combined concepts from visual question-answering with action recognition to model temporal relations between subjects and objects. This resulted in the TRN module, designed to learn and reason about temporal dependencies across videos at multiple timescales. Authors explicitly argue that temporal reasoning at multiple time scales is essential because many actions have both long and short-term dependencies or cues. They evaluated TRN using several datasets, including Something-Something, Charades, and Jester. These datasets represent an interesting choice because they incorporate more generic actions, like poking something or giving a thumbs up, in contrast to the Sports1M dataset, where there may be a lot of contextual information from background imagery or objects in the scene.

The Temporal Shift Model (TSM)~\cite{lin2019TSM} extends the concept of shift operation to video understanding. Essentially, channels from 2D convolutions can be ``shifted" along the temporal dimension (consecutive frames). This results in feature maps, which are combinations of features from the current and neighboring frames. However, shifting large amounts of data degrades model accuracy, which authors attribute to a reduced capacity for spatial modeling. In the proposed TSM, authors balance spatial and temporal learning by shifting only portions of the channels along the temporal dimension. Additionally, authors only insert the TSM into residual connections within the network, achieving what they term ``multi-level temporal fusion". These implementation choices allow TSM to adapt 2D CNNs for temporal modeling with little additional computational and memory overhead.

Materzynska et al.~\cite{materzynska2020something} proposed Spatial-Temporal Interaction Networks (STIN), which aimed to learn interactions between agents and objects. 
The authors were particularly interested in whether learned verbs or actions could be effectively identified when various objects that were not part of the training set are incorporated into the test set. 
The system relied on the spatial/geometric relationships between detected objects and their temporal relationships over a series of input frames. By predicting the bounding boxes and the positions of objects, the network was able to model these geometric relations between objects and the agents manipulating them. To examine the effectiveness of STINs, the authors introduced a new benchmark, the Something-Else task, a subset of Something-Something V2. In this task, frequent objects are featured exclusively with specific actions, and at test time, the associations are broken (novel combinations of objects and actions are evaluated).

Both the SpatioTemporal and Motion Encoding (STM)~\cite{jiang2019stm} and Temporal Excitation and Aggregation (TEA)~\cite{li2020tea} methods utilize motion and temporal modules to unearth long-term dependencies, both also propose implementing these modules as modifications to ResNet Blocks. However, STM calculates the feature difference between frames (preserving only motion) in their Channel-wise motion modules, whereas TEA designs their Motion Excitation module to preserve background features while emphasizing motion-sensitive features. This design allows TEA to contextualize motion features through the use of spatial-temporal features, learning motion patterns with scene information. TEA further augments the ME module by adding a Multiple Temporal Aggregation module, which performs cascading channel-wise convolutions on the motion excitation information. Recently, multiple works have spawned off of TEA. 

The Multi-Kernel Excitation Network~\cite{tian2022multi} for video action recognition utilized a multi-kernel attention (MKA) module to add the capacity for temporal modeling to 2D convolutional networks. In particular,
the authors added the MKA module to a ResNet and evaluated it on the Something-Something dataset. Similarly, Joefrie et al.~\cite{joe2022video} proposed Motion and Multi-View Excitation and Temporal Aggregation (META), a building block that can be inserted into traditional CNNs.
The goal was to add temporal reasoning to 2D convolutional networks, this time through multiple views. The module relied on three submodules designed for Motion Excitation, Multi-view Excitation (MvE), and Densely Connected Temporal Aggregation. The MvE module uses 2D convolutions combined with the excitation algorithm to form a multi-view feature map. Specifically, authors expand upon the MV-CNN model~\cite{li2018team} by adding the excitation algorithm to the proposed module.


\subsubsection{Discussion}

Motion models are a broad category of HAR systems. There are no clearly defined criteria for the best way to represent moving entities and these properties in a video. As we have seen, approaches have used optical flows and estimators, frame sequence information, geometric relationships, and temporal and excitation modules. As of writing, temporal and excitation modules achieve the state of the art and have the potential for broad application using existing CNNs. It is important to note that these modules utilize a kind of attention selectively to identify critical feature information, a common theme among state-of-the-art HAR systems since attention modules have gained popularity. As motion models have evolved with HAR, much of the focus has been on representing spatiotemporal information and attention. More specifically, excitation modules are the most recent and rapidly developing innovation designed for this task, and they may become the dominant path forward for training these complex systems.

\begin{table}[h]
\centering
\small 
\caption{This label pertains to Figure~\ref{fig:HARVennDiagrame}, classified as Motion Models Networks based on the model architecture involved. For a detailed view of all categories, refer to Figure~\ref{fig:HARVennDiagrame}. \\ \textbf{\textit{Acronyms:}} Two-Stream Networks (T-SNs), Motion Networks (MN), 3D Convolutional (3DC), Graph Convolutional Networks (GCNs), Hybrid (H), and Transformer (T).}
\label{Tab:MNs-models}
\begin{tabular}{p{2.5cm} p{6cm} p{5cm}} 
\hline
\textbf{\textit{MN}} & Model architectures involved & Paper citations \\
\hline
MNs.a & Motion Models & \cite{fernando2016rank, zhou2018temporal, lin2019TSM} \\
MNs.b & Motion Models \& Hybrid Networks & \cite{misra2016shuffle} \\
MNs.c & Motion Models \& Hybrid Networks \& 3DC & \cite{zhu2019hidden, materzynska2020something} \\
MNs.d & Motion Models \& Hybrid Networks \& 3DC \& Attention & \cite{yang2022spatio} \\
\hline
\end{tabular}
\end{table}

\subsection{Transformer Models} \label{sec:Transformer Models}
\subsubsection{Background}

Translating sentences from a source language into a target language presents an intricate puzzle, often testing the limits of human intelligence. We handle this complexity through the intuitive partitioning of sentences into manageable fragments, processing each individually, taking into account how the fragments impact one another. Reflecting on this human approach, Bahdanau et al. \cite{bahdanau2014neural} designed the attention mechanism to enhance the encoder-decoder model's efficacy in machine translation. The attention mechanism's fundamental principle is to adaptively seek pertinent segments of the source sentence when producing the translated sequence. This strategy eliminates the need for explicit alignment of and target fragments of these segments.

In the realm of vision, the human brain instinctively identifies objects and actions within images or videos by selectively concentrating its focus, akin to machine translation. Xu et al.~\cite{xu2015show} harnessed the power of the attention mechanism for the intricate task of image caption generation. They crafted two specialized attention techniques exclusively for image caption generators. The first, referred to as `soft attention', is a deterministic method trainable by conventional back-propagation techniques. The soft attention mechanism does not single out a specific part of the image. Instead, it assigns varying weights to each part of the image, creating a weighted representation. Consequently, a ``new'' image is formed where each portion carries a unique weight, leading to a deterministically differentiable output. The second, termed `hard attention,' is a stochastic method, trainable by either maximizing an approximate lower bound or utilizing reinforcement learning \cite{williams1992simple}. In the hard attention mechanism, a specific portion of the image is selected as the sole focus for generating a particular output. For instance, an image captioning model may focus on a particular object within an image to produce a corresponding word output.

As articulated by Vaswani et al. \cite{vaswani2017attention}, the Transformer architecture was principally designed for language comprehension undertakings, taking a bold departure from the conventional recurrence in neural networks. The architecture places complete reliance on self-attention mechanisms to establish global dependencies between inputs and outputs. Alternatively referred to as intra-attention, self-attention has carved a niche for itself in the realm of Natural Language Processing (NLP), significantly elevating performance levels across a diversified array of tasks. These range from reading comprehension and abstractive summarization to textual entailment, not forgetting the crafting of task-independent sentence representations, as evidenced in many studies \cite{cheng2016long, parikh2016decomposable, paulus2017deep, lin2017structured}.

In the domain of vision, the self-attention mechanism has been effectively deployed to improve image classification. Parmar et al. \cite{parmar2018image} harnessed the power of self-attention within local neighborhoods for each query pixel instead of a global scale, employing 1D/2D local attention. This local attention selectively hones in on a defined `neighborhood' or a particular area of interest. Such a focus might be a specific object within an image or a compact, localized region of the input space. Proving highly efficient in scenarios where contextual information is primarily confined to small, specific areas, local attention alleviates the necessity for a comprehensive scan of the entire input space. Dosovitskiy et al. \cite{dosovitskiy2020image} demonstrated that the Vision Transformer model (ViT), utilizing global attention, outperforms local attention. The model uses the encoder component of the original Transformer architecture. Global attention operates by considering all input positions in its calculations, effectively adopting a comprehensive perspective of the data. This strategy ensures that even the most subtle interconnections across remote elements are factored into the decision-making process. Interestingly, the authors revealed that reliance on CNNs is no longer a necessity for processing images. In fact, a pure Transformer, when applied directly to sequences of image patches, exhibits substantial prowess in image classification tasks.

\subsubsection{Extensions of Transformer Network for HAR}
Video understanding bears a close resemblance to NLP. Both tasks are executed in a sequential manner and require context consideration for disambiguation. Researchers began using Transformers for HAR due to its impressive ability to model intricate dependencies in text data and its adaptability across diverse tasks. Vaswani et al.~\cite{vaswani2017attention} defined an attention function as mapping a query $Q$ and memory (key $K$ and value $V$) of dimension $d_k$ to an output. The output is computed as a weighted sum of $V$, such that the weights are calculated from the product of $Q$ and $K$. 
In the task of translation, $Q$ is the source word or the word being translated, and $K$ and $V$ are linear projections of the input and output sequences created. Action Transformer~\cite{girdhar2019video} is one of the earlier works that used a Transformer architecture for HAR such that $Q$ is the person being classified, and the context in the clip is the memory, projected into keys and values. Given a video clip, Action Transformer creates a spatio-temporal feature representation using a trunk network, commonly an initial pre-trained layer of I3D. A Region Proposal Network processes the central frame of the feature map, generating bounding box proposals. These proposals and the location-embedded feature map navigate through a head network to generate a unique feature, which is then employed to refine the bounding box and classify actions. A series of Action Transformer units, integral to the head network, is responsible for generating the classifying features.

Graph Convolutional Networks discussed in section~\ref{subsec:GCNs}, have been used to recognize human actions from video based on the skeletal structure due to their ability to represent non-Euclidean data effectively. However, the spatial-temporal Graph Convolutional Networks (ST-GCN) may derive rich representations ineffectively due to the fixed representations of the human body and actions. In addition, ST-GCN performs poorly in a local neighborhood because the convolutions are based on a standard 2D convolutional and lack correlations between body joints. The two-stream Spatial-Temporal Transformer (ST-TR) model \cite{plizzari2021spatial,plizzari2021skeleton} overcomes the limitations of ST-GCN by combining a spatial self-attention model, which is used to extract spatial features and represent the relationships between parts of a human body or to learn intra-frame interactions, with a temporal self-attention model, which is employed to study the dynamics of joints or to capture inter-frame motion dynamics. The two streams are fused by summing up their softmax output scores to calculate the final classification. The authors report that adding joint and bone information as input enhances the model's performance significantly. 

The GCN-based HAR models are inefficient in real-time due to vast computation on the dense skeleton representation for extracting features of neighboring nodes. Shi et al.~\cite{shi2021star} solve this problem by processing variable lengths of skeleton action sequences without the need for further pre-processing using a Spare Transformer-based Action Recognition (STAR) model. STAR comprises Spatial-Temporal (ST) Transformer blocks followed by a context-aware attention layer and a Multi-Layer Perceptron (MLP) head for classification. Each ST Transformer block adheres to the standard Transformer encoder architecture consisting of a multi-head self-attention (MSA) layer, a skip connection, and a feed-forward network. A spatial encoder captures the correlation between skeleton joints, and a temporal encoder captures the correlation of joints along the time dimension. In contrast to the GCN-based HAR models, the Transformer-based model can handle inputs of different lengths and subjects. START~\cite{shi2021star} is more effective in terms of model size, computations, and latency but less accurate than ST-GCN~\cite{yan2018spatial} and ST-TR~\cite{plizzari2021spatial}. Most Transformer-based HAR models use the same strategy to handle the skeleton in spatial and temporal dimensions, which do not share the same mechanism. Spatial-Temporal Specialized Transformer~\cite{zhang2021stst} (STST) for skeleton-based action recognition enhances the Transformer-based HAR model by separately modeling pose information at the frame level using the Spatial Transformer Block and capturing the actions of the entire skeleton using the Directional Temporal Transformer Block, respectively.

HAR is enhanced by the combination of spatio-temporal video and skeleton data, but presenting features for cross-modal data is challenging. Ahn et al.~\cite{ahn2023star} propose a spatio-temporal cross (STAR)-Transformer, which is able to learn cross-model features successfully. Video frames and corresponding skeleton sequences are fed to a shared CNN-based model to extract local and global feature maps. The global feature map is transformed into a global grid token. The local feature map is concatenated with the joint heatmap to produce a joint map token. To fuse the two modalities, the concatenation of a global grid token and a joint map token is fed into the STAR transformer to predict the action. The STAR-transformer model employs spatio-temporal attention and follows the standard Transformer architecture. Specifically, the encoder consists of a full spatio-temporal attention (FAttn) and a zigzag spatio-temporal attention, while the decoder has an FAttn module and a binary spatio-temporal attention module. Kim et al.~\cite{kim2022cross} introduce a 3D deformable Transformer with adaptive spatio-temporal receptive fields and a cross-modal learning scheme. The model includes a backbone network and Transformer blocks. The backbone network utilizes an RGB modality and a pose modality to provide visual feature maps. The 3D deformable Transformer consists of three attention modules: the 3D deformable attention for the RGB images, the joint stride attention for the poses, and the temporal stride attention for the prediction based on the output of the two previous attentions. The Transformer block has the ability to fuse cross-modal features using the modalities with cross-modal tokens. After multiple iterations, the model combines with the last cross-model tokens for classification.

Zhang et al. \cite{zhang2021co} improve the Transformer-based HAR models using a co-training Transformer paradigm (CoVeR), such that the model is simultaneously trained on video and image datasets to improve representations. CoVeR learns robust spatial and temporal representation via simultaneous learning across multiple action recognition tasks. Local redundancy and complex global dependency between frames are challenges when working with high-dimensional videos. Li et al. \cite{li2022uniformer} report that the GCN may effectively decrease the local redundancy issue by aggregating local context, while the Transformer may capture the global dependency using a self-attention mechanism. Consequently, they introduced a UniFormer network \cite{li2022uniformer}, which leverages the benefits of 3D convolution and spatiotemporal self-attention by learning local relations in the shallow layers and global relations in the deep layers. The UniFormer network is based on the Transformer architecture. It consists of a stack of UniFormer blocks, each of which includes three modules: a dynamic position embedding, a multi-head relation aggregator (MHRA), and a feed-forward network. Unlike the standard Transformer-based HAR models that separate the spatial and temporal attention, UniForm \cite{li2022uniformer} encodes spatiotemporal context in all layers.

\begin{figure}[hbt!]
    \centering
    \includegraphics[width=14cm]{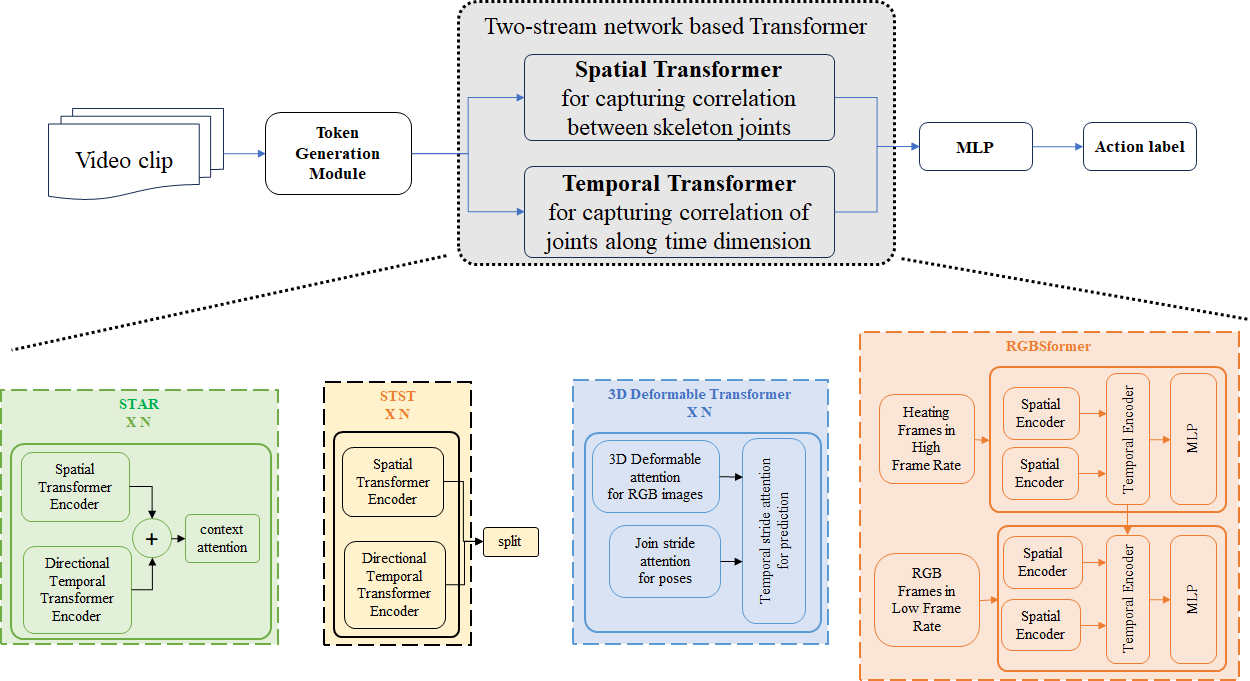}
      \caption{Two-stream network-based Transformer for human action recognition. The figure on the top illustrates the general architecture of the two-stream network-based Transformer, while the figures on the bottom present various pipelines of the Transformer-based two-stream network. Figure adopted from \cite{ahn2023star}\cite{zhang2021stst}\cite{kim2022cross}\cite{shi2023novel}.}
    \label{fig:TransformerBasedHAR}
\end{figure}

Shi et al. \cite{shi2023novel} exploit both the skeleton and RGB modalities by developing RGBSformer, which follows a two-stream Transformer-based framework (section~\ref{subsec:two-stream}) and slow-fast pathway \cite{feichtenhofer2019slowfast}. RGB videos are utilized to obtain skeleton heatmaps. The heatmap frames at high frame rates are fed to the spatial Transformer encoder with higher temporal resolution and lower spatial resolution compared with the RGB frames at low frame rates. The output of the spatial Transformer encoder of the skeleton stream is fused into the RGB stream. This fusion is fed to the temporal Transformer encoder and the MLP head. The model predicts the final action based on the average score from two streams. To fuse the skeleton information to the RGB stream, RGBSformer uses two fusion methods, including the score fusion based on average scores from the two streams and the classification token fusion based on concatenating tokens and applying score fusion at the classification head.

We highlight some of the Transformer-based HAR approaches discussed in Figure~\ref{fig:TransformerBasedHAR} and summarize the attention mechanisms employed in the studies presented in this subsection in Table~\ref{Tab:Transformer-attentionSum}.

\begin{table}[!ht]
\centering
\small 
\caption{Transformer-based models and attention mechanisms}
\label{Tab:Transformer-attentionSum}
\begin{tabular}{p{3cm} p{4.5cm} p{5cm}} 
\hline
\textbf{Model} & \textbf{Attention Mechanisms} & \textbf{Purpose} \\ 
\hline
Action Transformer~\cite{girdhar2019video} & Self-attention & Learned during the training process \\ 
\hline
\multirow{2}{*}{ST-TR~\cite{plizzari2021spatial}} & Spatial self-attention & Understand interactions between skeleton joints \\ 
& Temporal self-attention & Study joint dynamics and capture discriminant features \\ 
\hline
\multirow{2}{*}{STAR~\cite{shi2021star}} & Sparse self-attention & Extract correlations among connected joints via sparse matrix multiplications \\ 
& Segmented linear self-attention & Provide joint correlations in a time-series context \\ 
\hline
\multirow{2}{*}{STST~\cite{zhang2021stst}} & Forward and backward self-attention & Model skeleton sequences in spatial and temporal dimensions separately \\ 
& 3D deformable attention & Create cross-attention tokens \\ 
\hline
\multirow{2}{*}{3D Deformable Transformer~\cite{kim2022cross}} & Local joint stride attention & Spatially combine attention and pose tokens \\ 
& Temporal stride attention & Support temporal expression learning without using all tokens simultaneously \\ 
\hline
STAR-Transformer~\cite{ahn2023star} & Zigzag and binary spatio-temporal attention & Learn cross-modal features \\ 
\hline
\end{tabular}
\end{table}

\subsubsection{Vision Transformer-based Models for HAR}
\begin{figure}[hbt!]
    \centering
    \includegraphics[width=\linewidth]{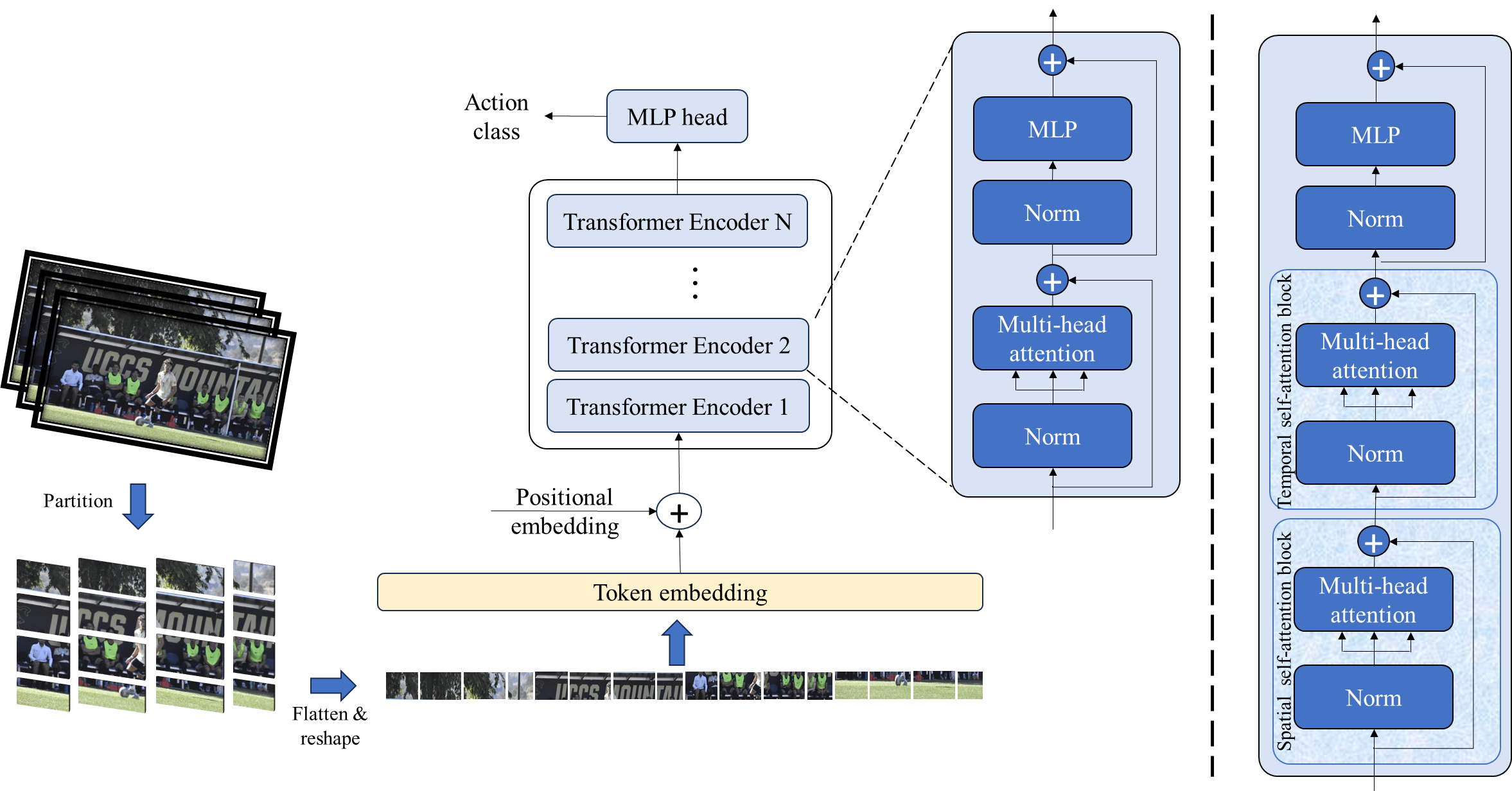}
    \caption{Human action recognition model inspired by the Vision Transformer model for images. The right side is a Transformer Encoder block, which splits the standard multi-head attention into spatial and temporal self-attention blocks. The Figure adopted from \cite{vaswani2017attention} \cite{arnab2021vivit}.}
    \label{fig:VisionTransformerBasedHAR}
\end{figure}

In the context of video analysis, ViT divides a video into fixed-size, non-overlapping patches, each flattened into a vector and enriched with positional embeddings. This setup allows the Transformer to capture complex spatial and temporal patterns across the video sequence, as shown in Figure~\ref{fig:VisionTransformerBasedHAR} for human action recognition~\cite{liang2022dualformer}.

A ViT-based and convolution-free HAR model, TimeSformer~\cite{bertasius2021space}, learns spatiotemporal features directly from a sequence of frame-level patches. Given an input clip, the authors decompose each frame into $N$ fixed-size non-overlapping patches, flatten these into vectors, add them to positional embeddings, and then feed these to the standard Transformer blocks to capture the spatiotemporal relationships using a divided spatiotemporal attention mechanism. TimeSformer \cite{bertasius2021space} outperforms the GCN-based approaches regarding speed, accuracy, and handling longer videos. ViViT \cite{arnab2021vivit} adopts ViT for HAR from input video, focusing on modeling global attention on the spatial-temporal video tokens. ViViT handles many tokens and uses model variants that factorize components of the Transformer encoder over the spatial and temporal dimensions. ViViT converts an input video into tokens using standard uniform frame sampling (based on non-overlapping patches from 2D frames) and tubelet embedding (based on non-overlapping tubes from 3D spatio-temporal input volume). The Transformer fuses the uniform frame sampling embeddings, whereas the tubelet embeddings are fused during tokenization.

Similarly, DualFormer~\cite{liang2022dualformer} processes the input clip into tokens, which then serve to learn the visual representations. DualFormer has four stages, each including a four stack of DualFormer blocks. A DualFormer block employs self-attention mechanisms using a local-global stratification strategy to capture long and short-range information within the clip. The action is predicted by applying a global average pooling layer followed by a linear classifier. Yan et al. \cite{yan2022multiview} introduce Multiview Transformer (MTV) based on ViViT using separate standard Transformer encoders specialized for different representations or ``views'' of the input video with lateral connections for cross-view information fusion. The experiments show that processing multiple views improves accuracy and computational cost more than increasing the depth of a single view as SlowFast \cite{feichtenhofer2019slowfast}. 

The Action Transformer (AcT) model~\cite{mazzia2022action} is also based on ViT. AcT exploits 2D skeletal representations of short-time sequences. The frames of a given video input sequence are pre-processed to estimate poses, which are then projected linearly to the model's dimension. The projected tokens and class tokens are input tokens, each of which is added to a corresponding positional embedding and fed to the Transformer encoder. An MLP head predicts the action based on the last class output token. Chen et al.~\cite{chen2022mm}  introduce the Multi-Modal Video Transformer (MM-ViT), a Vision Transformer (ViT)-based model designed for enhanced learning through a multi-modal approach. Unlike other Vision Transformer-based models focused solely on decoding RGB frames, MM-ViT excels in handling compressed videos. It innovatively integrates different data types, including appearance (from I-frames), motion (via motion vectors and Residuals), and audio. This integration is achieved by factorizing self-attention across spatial, temporal, and modality dimensions, demonstrating a comprehensive and efficient approach to video analysis.

Multiscale Vision Transformers (MViT) \cite{fan2021multiscale} extends ViT for a video domain by employing multiscale feature hierarchies. MViT is based on the channel resolution scale stages, each of which composes Transformer blocks with particular space-time resolution and channel dimensions. MViT expands the channel capacity and pools the resolution between different stages. MViTv2~\cite{li2022mvitv2} notably improves upon the MViT model by integrating decomposed location distance, which introduces positional information, into the Transformer block through shift-invariant positional embedding. Additionally, they incorporate residual pooling connections, further refining the model's architecture. They employ a standard dense prediction framework to demonstrate the model's applicability: Mask R-CNN~\cite{he2017mask} combined with Feature Pyramid Networks (FPN)~\cite{lin2017feature}. This approach is effectively applied in tasks like object detection and instance segmentation, showcasing the model's versatility and efficiency. However, Ryali et al. \cite{ryali2023hiera} report that these hierarchical Vision Transformer models, such as MViT2, are effective but slower overall because of the lack of inductive bias after the patch operation. 

UniFormerV2 model \cite{li2022uniformerv2} inherits UniForm but follows the ViT paradigm. The UniFormerV2 uses local UniBlocks by inserting the local temporal MHRA before the spatial Vision Transformer block, global UniBlocks by applying a global cross MHRA, and multi-stage fusion blocks. Sun et al. \cite{sun2023integrating} improve the fusion accuracy using the VT-BPAN model based on spatial lightweight Vision Transformer, bilinear pooling, and attention network. Specifically, the RGB features and skeleton features are fused by the bilinear pooling method before feeding to the VT-BPAN module, followed by the attention module and the MLP head for final classification.

Inspired by ImageMAE~\cite{he2022masked}, Tong et al.~\cite{tong2022videomae} introduce VideoMAE, a method that applies a customized high-ratio masking strategy to video tubes. Their work demonstrates that video-masked autoencoders (VideoMAE) serve as data-efficient learners for self-supervised video pre-training (SSVP). The model's design challenges video reconstruction as a self-supervision task, thereby promoting the extraction of more effective video representations during the pre-training process. Next year, Wang et al. \cite{wang2023videomae} enhance the computational and memory requirements for HAR by devising a video-masked autoencoder (VideoMAE V2). The model utilizes a masking map for both the encoder and decoder. VideoMAE V2 is trained using a progressive pipeline on billion-level video transformers and then post-pre-trained on the label hybrid dataset.

Sun et al.~\cite{sun2023masked} introduce Masked Motion Encoding (MME), a pretraining approach designed to reconstruct appearance and motion information, thereby capturing temporal dynamics. MME aims to tackle two critical challenges for enhancing representation performance by effectively representing potential long-term motion across multiple frames and extracting fine-grained temporal details from sparsely sampled videos. Drawing inspiration from the human ability to recognize actions by observing changes in object positions and shapes, they propose reconstructing a motion trajectory that captures these changes in the masked regions.

Another architecture by Hiera \cite{ryali2023hiera}, a simple hierarchical Vision Transformer using vanilla ViT blocks, adds spatial bias using a visual pretext task through masked autoencoders, resulting in a more accurate and faster model during the inference and training stages. Piergiovanni et al.~\cite{piergiovanni2023rethinking} introduce a sparse video TubeViT based on ViT. TubeViT can learn from images and videos simultaneously by sparsely sampling different-sized tube patches from the video to create learnable tokens for Vision Transformer. Table~\ref{Tab:ViTSum} overviews the ViT-based approaches for HAR, attention mechanisms used, and types of input video.

Srivastava et al.~\cite{srivastava2024omnivec} used Vision Transformer Encoders as backbone networks to extract multi-modality features as input, including ViT for image and depth directly, ViViT for video. AST is used for audio, and a standard BERT transformer is used for textual data. The same authors in~\cite{srivastava2024omnivec2} extend the capabilities of the original OmniVec by introducing a broader range of modality support, including advanced data types like X-ray, infrared, and hyperspectral data. OmniVec2 enhances this with modality-specific tokenizers, a shared transformer architecture, and cross-attention mechanisms, allowing for more sophisticated multimodal and multitask learning.

\begin{table}[!ht]
\caption{Vision Transformer-based models and attention mechanisms. $T$ denotes the number of frames of dimension $H$ in height, $W$ in width, $C$ channels, and $D$ dimension.}\label{Tab:ViTSum}
\centering
\begin{tabular*}{\textwidth}{lp{1.75cm}p{3.35cm}p{2.25cm}p{2.cm}}
\midrule
\textbf{Model}& \textbf{Attention mechanisms} &  \textbf{Used to}& \textbf{Input video $X$ in}&\textbf{Tokens}\\ \midrule 
TimeSformer\cite{bertasius2021space} &divided space-time attention & temporal attention and spatial attention are applied one after the other&$R^{H\times W\times3\times T}$& patches\\ \midrule

&factorised self-attention& ﬁrst only compute self-attention spatially then temporally& &uniform frame sampling ~~~~~ and\\ \cmidrule(lr){2-3}

ViViT~\cite{arnab2021vivit}& factorised dot-product attention&compute
attention weights for each token separately over the spatial and
temporal-dimensions using different heads&$R^{T\times H\times W\times C}$& tubelets\\ \midrule

&local-window based MSA &extract short-range interactions among nearby tokens&&\\ \cmidrule(lr){2-3} 
DualFormer~\cite{liang2022dualformer}& global-pyramid based MSA& capture long-range dependencies between the query token and the coarse-grained global pyramid contexts&$R^{T\times H\times W\times 3}$& patches\\ \midrule

MTV \cite{yan2022multiview}&cross-view attention &combine information between different views to perform self-attention on all tokens&$R^{T\times H\times W\times C}$&tubelets\\ \midrule

Act~\cite{mazzia2022action} &self-attention&assemble information into a condensed, high-dimensional representation&$R^{T\times H\times W\times C}$& pose tokens\\ \midrule

MM-ViT~\cite{chen2022mm}&cross-modal attention (merged/ co-attention/ shift-merge)&learn the inter-model interactions&compressed video $V$ including $T$ sampled $I$-frames, motion vectors, and residuals $H \times W$& patches\\ \midrule

MViT~\cite{fan2021multiscale,li2022mvitv2}&muilti-head pooling attention&incorporate decomposed relative positional embeddings and residual pooling connections&$ R^{D\times T\times H\times W}$ &patches\\ \hline

Hiera~\cite{ryali2023hiera}&mask unit attention&local attention within a mask unit&$R^{D\times T\times H\times W}$&patches\\  \hline

TubeViT~\cite{piergiovanni2023rethinking}&pooling attention&&image \& video &tubes\\
\end{tabular*}
\end{table}

\subsubsection{Vision Transformer-Based Transfer Models for HAR} 

In recent years, pre-trained vision-language models have been utilized for HAR in videos. A baseline transfer approach embeds the video and category description into a pre-aligned feature space and selects the category closest to the video. The X-CLIP model~\cite{ni2022expanding} learns to align the video representation and corresponding text representation by jointly training a video encoder and a pre-trained text encoder. The pre-trained text encoder is expanded with a video-specific prompting scheme. The video encoder consists of a cross-frame communication transformer and a multi-frame integration transformer for modeling temporal aspects of the video. Wu et al.~\cite{wu2023bidirectional} introduce a two-stream BIKE framework utilizing the cross-modal bidirectional knowledge from CLIP~\cite{radford2021learning} pre-trained ViT. BIKE consists of an attribute branch that employs the video-attribute association mechanism to obtain relevant phrases as auxiliary attributes and a video branch that uses the video concept spotting mechanism to measure the similarity between frames and categories. The same author, Wu et al., in \cite{wu2023revisiting}, enhances the vision-language pre-trained model for HAR by replacing the linear classification with different knowledge from the CLIP model~\cite{radford2021learning} to generate a semantic target of efficient transferring learning.

Most existing HAR approaches use transfer techniques with videos and text modalities. Building upon the foundational architecture introduced by BIKE \cite{wu2023bidirectional}, Chaudhuri and Bhattacharya \cite{chaudhuri2023vilp} have innovatively adapted the CLIP pre-trained Vision Transformer (ViT) to capture pose information. This is achieved through a comprehensive multimodal strategy integrating video, text, and pose data. Specifically, their approach includes a pose encoder designed to extract skeleton information, which markedly enhances the model's accuracy when combined with a temporal saliency generation scheme. 

Li et al. \cite{li2023unmasked} introduce a progressive pre-training method for temporal-sensitive Video Foundation Models using UnMasked Teacher (UMT). The low-semantics video tokens are masked, while unmasked tokens are aligned with Image Foundation Models. Specifically, video data is used to mask video modeling, and then public vision-language data is utilized for multi-modality learning. The UMT not only reduces training sources but also speeds up convergence. Videos may have redundant content, leading to unnecessary computations for feature extraction. Pan et al. \cite{pan2023svt} propose a Supertoken Video Transformer (SVT) using a Semantic Pooling Module (SPM) to merge latent representations into supertokens based on semantic similarity. SVT increases the proportion of salient information and reduces the redundancy inherent in video.

In addition to Vision Transformer-based models like CLIP, recent advancements in transfer-based models which could be applied to HAR (as CLIP was) incorporate instruction tuning and multimodal prompt tuning techniques, demonstrating notable improvements in model adaptability and efficiency. One remarkable development is Visual Instruction Tuning~\cite{liu2024visual}, as demonstrated by the LLava model. LLava introduces the concept of fine-tuning large vision-language models using multimodal instructions. This involves providing natural language instructions alongside image or video inputs, which helps the model interpret the context with more nuance, leading to better generalization. In particular, authors argue the instruction tuning paradigm enables the model to follow human intent. In the context of HAR, an adapted version of LLava may be able to recognize new HAR tasks without requiring task-specific retraining. The core of this technique lies in aligning visual and textual embeddings using a multi-stage fine-tuning process, where pre-trained visual and text encoders are jointly optimized to perform cross-modal reasoning. Some pre-published work has already started to explore the use of LLava for HAR \cite{lu2024enhancing}.

Building directly on LLava, MMPT: Multimodal Prompt Tuning~\cite{wang2024mmpt} offers a zero-shot approach to transfer learning by exploiting pre-trained multimodal models. MMPT visual and textual prompts optimized through a lightweight tuning process. Instead of updating the entire network, MMPT focuses on fine-tuning only the prompt tokens and an interaction layer while freezing most network parameters. This drastically reduces computational overhead and enables the model to rapidly adapt to unseen tasks. For example, in HAR, MMPT could generate task-specific prompts that describe actions (e.g., 'person running' or 'object falling') and tune the visual encoder to focus on the most relevant parts of the input, leading to improved activity recognition.

Another significant contribution is E\textsuperscript{2}VPT~\cite{han20232vpt}, which enhances model adaptation using prompt-tuning through attention and prompt pruning. E\textsuperscript{2}VPT modifies encoder attention layers, adding new key-value pairs which help the model adapt to new data. Additionally, authors introduced a prompt-pruning technique to eliminate unnecessary prompts, improving efficiency. E\textsuperscript{2}VPT achieves state-of-the-art performance on multiple recognition tasks. This work, which has made significant strides in prompt-tuning-driven model adaptation, could reasonably be adapted for HAR tasks in future work.

Han et al.~\cite{han2024facing} investigated the trade-off between visual prompt tuning and full fine-tuning for video-based tasks. Their findings revealed that while full fine-tuning improves marginal performance, visual prompt tuning offers a more computationally efficient alternative. By optimizing only a subset of the model’s parameters (i.e., prompt embeddings), visual prompt tuning allows models to retain general knowledge from pre-trained weights. If adapted to HAR, this work could provide efficient general knowledge transfer to new actions or settings. We provide the main strategies of the transfer-based HAR approaches in Table~\ref{Tab:TransferBased-sum}.

\begin{table}[!ht]
\centering
\small 
\caption{Transfer-based HAR methods}
\label{Tab:TransferBased-sum}
\begin{tabular}{p{2.5cm} p{2cm} p{4cm} p{4.5cm}} 
\hline
\textbf{Model} & \textbf{NLP Model} & \textbf{Key Components} & \textbf{Highlights} \\ 
\hline 
X-CLIP~\cite{ni2022expanding} & CLIP & \begin{itemize} 
    \item Cross-frame communication transformer 
    \item Multi-frame integration transformer 
\end{itemize} & \begin{itemize}
    \item Replaces spatial attention with cross-frame attention 
    \item Jointly trains a video encoder and text encoder 
\end{itemize} \\ 
\hline
BIKE~\cite{wu2023bidirectional} & CLIP & \begin{itemize} 
    \item Attribute branch 
    \item Video branch 
\end{itemize} & \begin{itemize}
    \item Uses visual encoder of CLIP as video encoder and textual encoder of CLIP for category and attribute encoders 
    \item Trains video encoder first, then attribute encoders 
\end{itemize} \\ 
\hline
Wu et al.~\cite{wu2023revisiting} & CLIP & \begin{itemize} 
    \item Visual encoder 
    \item Textual encoder 
\end{itemize} & \begin{itemize}
    \item Uses pre-trained visual encoder to extract visual embeddings, performs LDA to create LDA coefficients, and fine-tunes the pre-trained encoder 
    \item Transfers textual semantic knowledge from a pre-trained textual encoder 
\end{itemize} \\ 
\hline
ViLP~\cite{chaudhuri2023vilp} & BIKE & \begin{itemize} 
    \item 3 modalities: video encoder, text encoder, pose encoder 
\end{itemize} & \begin{itemize}
    \item Uses BIKE's video branch for video and text encoding 
    \item CLIP pretrained ViT for video representation 
    \item CLIP's text encoder for textual context modeling 
\end{itemize} \\ 
\hline
UMT~\cite{li2023unmasked} & ViT & \begin{itemize} 
    \item Progressive pre-training with unmasked teacher 
\end{itemize} & \begin{itemize}
    \item Trains temporal-sensitive Video Foundation Models 
    \item Masks most low-semantic video tokens, aligning unmasked tokens with Image Foundation Models 
\end{itemize} \\ 
\hline
SVT~\cite{pan2023svt} & ViT & \begin{itemize} 
    \item Semantic pooling module 
\end{itemize} & \begin{itemize}
    \item Merges latent visual token embeddings based on distances 
\end{itemize} \\ 
\hline
\end{tabular}
\end{table}

\subsubsection{Discussion}
Transformer-based methods are highly effective for Human Action Recognition tasks, as they adeptly manage variable input lengths and diverse subjects \cite{shi2021star}. They can learn the context from other people and objects surrounding them to localize and classify actions. ViT outperforms CNN-based models in capturing long-range dependencies by replacing CNN’s inductive biases of locality with global relation modeling through MSA \cite{ryali2023hiera, dosovitskiy2020image}. However, as the number of tokens increases, the computational cost of MSA rises quadratically \cite{liang2022dualformer}. To mitigate this, various attention mechanisms have been proposed, including divided space-time attention \cite{arnab2021vivit}, local window-based attention \cite{liu2022video}, and dual-level MSA \cite{liang2022dualformer}. Interestingly, Piergiovanni et al. \cite{piergiovanni2023rethinking} found that factorized attention techniques are ineffective for networks pre-trained on images. ViT also demands substantial amounts of training data to achieve optimal results. To address this, models may be initialized with pre-trained image networks or integrated with multi-modal learning. Another way to enhance attention mechanisms for HAR is to apply recent advancements like CLUSTERFORMER~\cite{liang2024clusterfomer}, which is a novel clustering-based approach within the Transformer framework. It employs recursive clustering and feature redistribution; CLUSTERFORMER learns robust, adaptable representations and enhances interpretability to vision models.

Despite their advantages, Transformer-based video HAR approaches face high computational costs due to the large number of spatiotemporal tokens involved. To enhance efficiency, several strategies have been proposed, such as using low-rank approximations, factorizing attention mechanisms \cite{chen2022mm}, reducing resolution \cite{fan2021multiscale,li2022mvitv2}, and learning spatial biases through pretext tasks \cite{ryali2023hiera}. Additionally, researchers leverage powerful vision-language models, pre-trained on vast image-text pairs, applying knowledge transfer to improve HAR performance. We summarize the Transformer approaches for HAR discussed in this section in Table~\ref{Tab:Transformer-based-models}.

\begin{table}[h]
\centering
\small 
\caption{This label pertains to Figure~\ref{fig:HARVennDiagrame}, classified as a Transformer Network (TN) based on the model architecture involved. For a detailed view of all categories, refer to Figure~\ref{fig:HARVennDiagrame}. \\ \textbf{\textit{Acronyms:}} Two-Stream Networks (T-SNs), Motion Networks (MN), 3D Convolutional (3DC), Graph Convolutional Networks (GCNs), Hybrid (H), and Transformer (T).}
\label{Tab:Transformer-table}
\begin{tabular}{p{2.5cm} p{5cm} p{6cm}} 
\hline
\textbf{\textit{Transformer}} & Model Architectures Involved & Paper Citations \\
\hline
TN.a & TN \& Attention & \cite{plizzari2021spatial, plizzari2021skeleton, shi2021star, zhang2021stst, ahn2023star, zhang2021co, bertasius2021space, yan2022multiview, mazzia2022action, fan2021multiscale, li2022mvitv2, ryali2023hiera, wang2023videomae, piergiovanni2023rethinking, ni2022expanding, pan2023svt} \\ 
\hline
TN.b & TN, Attention \& 3DC & \cite{girdhar2019video, kim2022cross, liang2022dualformer, li2022uniformer, li2022uniformerv2} \\ 
\hline
TN.c & TN, Attention \& T-SN & \cite{shi2023novel, srivastava2024omnivec, srivastava2024omnivec2, wu2023bidirectional} \\ 
\hline
TN.d & TN \& T-SN \& 3DC & \cite{sun2023integrating} \\ 
\hline
\end{tabular}
\end{table}

\subsection{Hybrid Networks} \label{sec:Hybrid Models}
Our SMART-Vision taxonomy, as depicted in Fig~\ref{fig:HARVennDiagrame}, is a powerful visualization for analyzing and understanding the interworking of ideas and components from previous sections. It shows that Hybrid Networks can be formed from combinations of several basic HAR approaches in the taxonomy, thereby equipping the reader with the pertinent knowledge to incorporate alternate data modalities into Hybrid Networks, such as the combination of pose or joint data when using GCNs combined with masked auto-encoder transformer-based models that use RGB data.

While this paper has already presented some works in prior sections (such as transformers) based on their dominant design choices, we investigate how the synergistic combination of design choices led to more effective models. So, while some works may have been mentioned already in prior sections, hybrid models are, by definition, not exclusive in our taxonomy. The focus of this section is the discussion of why hybrid approaches have made contributions across the taxonomy, not necessarily a recounting of individual hybrid systems.

\subsubsection{Background}
Hybrid models for human action recognition are versatile conglomerate systems that adapt to the needs of the task at hand. They consist of multiple systems or components from action recognition systems described in previous sections. Common examples of hybrid models are combinations of attention mechanisms with Graph Convolutional Networks, 3D Convolutional Networks, or Skeletal information. Hybrid models combine the advantages of multiple systems to yield superior overall performance. They leverage either additional information extracted from inputs (such as skeletal pose estimation) or enhance the processing of extracted information (such as attention). Some systems are inherently hybrid, such as two-stream models (subsection \ref{subsec:two-stream}), which combine motion and spatial RGB information, while others have extended and combined existing frameworks \cite{li2020spatio}

\subsubsection{Extensions of Hybrid Networks for HAR}

Ye et al.~\cite{ye2020dynamic} proposed a Dynamic GCN system, which modifies GCNs by incorporating a Context-encoding Network (CeN) CNN. This CeN is incorporated inside a modified Graph Convolutional layer (GConv), which can be dropped into GCNs. The CeN is used to predict an adjacency matrix, which forms a context-enriched graph. The context-enriched graph undergoes a dynamic graph convolutional operation and is combined with a static graph, forming the proposed GConv Layer. Using the GConv Layer, Dynamic GCN leverages a simple CNN to learn better features for Graph Convolutions and achieves competitive state-of-the-art performance on NTU+RGB+D Skeleton-Kinetics and NTU-RGB+D 120 datasets.

EfficientGCN \cite{song2022constructing} combines attention with GCN blocks at numerous levels. EfficientGCN is a follow-up work stemming from ResGCN \cite{song2020stronger}. One of the main motivations of this work is to reduce the number of parameters while maintaining performance.
For context, EfficientGCN models have almost \textcolor{red}{$1/15_{th}$} the number of parameters as DynamicGCN and the B4 configuration demonstrates superior performance on the NTU 60 dataset. EfficientGCN extracts data into three input streams (Joint, Bone, and Velocity) and combines them before feeding into a mainstream. Each stream uses a combination of attention and GCN Blocks to process the input features. The attention mechanisms themselves are novel modules proposed by the authors to capture spatial-temporal joint information.

Chi et al.~\cite{chi2022infogcn} made use of attention modules for GCN-based HAR when they proposed InfoGCN. A module that sets InfoGCN apart from other GCN solutions is a Self-Attention-based Graph Convolution (SA-GC) module. The SA-GC module uses self-attention to create relationships between joint information and combines the resulting self-attention map with an evolving topology learned over time. This allows the SA-GC module to form a context-dependent topology, which the authors argue can better represent action. The authors validated these claims by demonstrating superior performance against numerous GCN-based competitors on the NTU RGB+D 60/120 and NW-UCLA datasets.

InternVideo~\cite{wang2022internvideo} combines attention, self-supervised pre-training, multi-modal learning, and the concept of a masked auto-encoder. The authors recognized the importance of the base image-based models and sought to create a base video model useful for downstream tasks. The network consists of two main modules: a Multi-Modal Video Encoder (UniformerV2) and a Masked Video Encoder (VideoMAE). These modules were inherently linked through a cross-modal attention module, which extends the multi-head cross-attention mechanism. Recently by the same author Wang et al.~\cite{wang2024internvideo2}, InternVideo2 has been introduced, which is an enhanced version of InternVideo. The core design of the model is a progressive training approach that unifies masked video modeling, crossmodal contrastive learning, and next token prediction, scaling up the video encoder size to 6B parameters. They prioritize spatiotemporal consistency by semantically segmenting videos and generating video-audio-speech captions, which improves the alignment between video and text. The authors evaluated both models on various downstream tasks and achieved state-of-the-art performance on several HAR datasets, including Kinetics 400-600-700, Something-Something V1/2, ActivityNet, HACS, and HMDB51.

Srivastava et al. \cite{srivastava2024omnivec} introduce a unified architecture called "OmniVec" for multi-task learning across different modalities, such as visual, audio, text, and 3D. The approach leverages self-supervised pre-training followed by sequential task training by employing task-specific encoders and a shared trunk. The same authors propose OmniVec2~\cite{srivastava2024omnivec2}, which extends the capabilities of the original OmniVec by introducing a broader range of modality support, including advanced data types like X-ray, infrared, and hyperspectral data. While OmniVec primarily focused on task-specific encoders and a shared trunk for joint learning across modalities, OmniVec2 enhances this with modality-specific tokenizers, a shared transformer architecture, and cross-attention mechanisms, allowing for more sophisticated multimodal and multitask learning. 

A combination of GCN and single stream base, called HybridNet \cite{yang2023hybridnet}, consists of three modules for HybridNet: a GCN-based feature extractor, a Gluing Module, and a CNN-Based feature processing module. The GCN-based feature extractor is adapted from an adaptive graph convolution module. The authors modified the kernel sizes and included a residual-like operation to pass shallow features to the following GC blocks in the module. One output stream is passed to a fusion classifier, while another is passed forward to the Glueing module and a CNN-based feature processing module. The Glueing module uses a local and global branch to model adjacent and distant joints in the GCN feature map. The output of these branches is concatenated and fed into the CNN-based feature processing module. The CNN-based feature processing module is a single-path multi-convolution-bottleneck architecture that relies on average pooling.

This is used to generate discriminative features from the GCN-encoded joints. The output is fused with the original GCN-based feature extractor output before going into a classifier. The authors validated HybridNet against a wealth of competitors on the NTU-RGB+D 60/120 and Kinetics-Skeleton datasets, where they achieved superior performance.

PoseConv3D \cite{duan2022revisiting} combines skeletal action recognition with 3D convolutional neural networks. The network works in two main phases, first extracting 2D poses and forming human-joint heatmaps, and then stacking these heatmaps before feeding them into a 3D CNN. The authors also applied techniques from SlowFast two-stream networks to enhance the information extracted from the stacked temporal dimension. 
Inspired by the recent success of Zhu et al.~\cite{zhu2019deformable}, Li et al.~\cite{li2020spatio} introduced 3D spatial-temporal deformable ConvNets by extending 2D deformable ConvNets into a 3D variety using an attention mechanism. In particular, a submodule with two data paths, which use spatial and temporal attention, is created. The attention mechanisms allow the selection of disjoint frames for the 3D convolutional operation, breaking free of the rigid constraints of traditional 3D convolutions. The authors argue that this allows spatial and temporal representations to be learned simultaneously. The submodule was designed to be a drop-in replacement for 3D convolution modules. The authors evaluated the work by modifying ResNet101 and comparing it against several competitors, most interestingly other networks that use 3D convolutions. The system showed superior performance on the UCF-101 and HMDB-51 datasets with a modest increase in complexity.

Multi-modal works in HAR sometimes combine pose estimation with RGB data, a task for which it is intuitive to use CNNs and GCNs. A notable work in 2020 showed that this can be invaluable when working with 3D convolutional networks. Researchers from the University of Nice proposed VPN \cite{das2020vpn}, a Video-Pose Network that combines feature maps from 3D convolutions with an attention network. This hybrid model can be placed on top of existing 3D convolutional networks and trained end-to-end.
The attention network is based on GCNs, and a purpose-built embedding loss aligns the spatial representations from the VPN with the 3D feature map from the convolutional backbone. These representations are then modulated and fed into a classifier.

Bruce et al.~\cite{bruce2022mmnet} proposed MMNet, a model-based multimodal network that also focused on the fusion between skeleton and RGB data.
This work combines a standard ResNet CNN with two GCNs. Rather than fusing the embeddings, the GCNs identify what the authors call ``spatial-temporal regions of interest,'' which focus the RGB input over time.
They argue that this is a practical way to extract effective features without incurring the computational costs or training difficulties needed for 3D convolutional networks. After training, the system combines outputs from the softmax layer for all three networks to produce recognition probabilities. One interesting work combined poses estimation and CNNs for what authors called Dynamic Motion Representation \cite{asghari2020dynamic}. The DynaMotion system used 3D heatmaps extracted by Mask R-CNN, which are fed into a shallow 3D CNN. The authors used a Mask R-CNN model pre-trained on the COCO dataset. They do not rely on depth or skeletal inclusive datasets and instead are trained and tested on the classic HAR datasets, including HMDB51 and UCF101.

\subsubsection{Discussion}
Hybrid models are combinations of base HAR systems. Accordingly, they are an amorphous category of HAR systems. As HAR continues to evolve, effective systems will be naturally combined with each other, forming new hybrid systems. Just as the advent of transformer networks spawned a new category of HAR systems, new systems will continue to be combined. Because it is intuitive to fuse well-performing systems, Hybrid models will remain a recurring theme in HAR.

\subsection{Additional Novel Work} \label{sec:Additional Recent Novel Work}
Numerous research papers in HAR have emerged recently, enhancing the ones already discussed. While most of these studies employ traditional methods for action recognition and primarily focus on enhancing accuracy by refining existing techniques, a few recent investigations have ventured into pioneering strategies. We delve into these papers here as they diverge from the categories previously outlined in this document. The study by Gao et al.~\cite{gao2020listen} introduced approaches to discern actions from untrimmed videos by synergizing audio representation with the video content. IMGAUD2VID employs sound as a strategic preview mechanism, effectively eliminating both short-term and long-term visual redundancies.

Furthermore, the authors present IMGAUD-SKIMMING, a method that iteratively pinpoints pivotal moments in untrimmed videos, minimizing long-term temporal redundancies and thus optimizing efficiency. Jain et al.~\cite{jain2020actionbytes} proposed the ActionBytes method to address the challenge of localizing actions within extensive untrimmed videos. Distinctly, it gleans insights from short-trimmed videos, diverging from conventional methods that typically rely on annotated untrimmed videos for training. This approach serves as a technique that emphasizes and regularizes action boundaries throughout the training process.

Recent research from Stanford University has introduced a technique that adeptly records alterations in objects and their interrelationships during actions  \cite{ji2020action}. While numerous prior studies have advocated using scene graphs to predict tasks based on static images, this research elevates an existing action recognition model by integrating scene graphs as spatial-temporal feature repositories. In Geometry Consistency inspired Key Point Leaning (GC-KPL) \cite{weng20233d}, the author proposed a new approach for learning 3D human joint locations from point clouds without human labels. This is achieved with the novel unsupervised loss formulations that account for the structure and movement of the human body.

Typically, large-scale HAR datasets encompass many classes. Teng et al. \cite{li2019large} introduce a pioneering approach to action search using a hyperbolic geometric graph termed \say{hyperbolic space network}. This technique is rooted in a shared hyperbolic space that bridges action hierarchies with videos. Using the potency of tree-shaped regions, the hyperbolic space methodology can seamlessly embed any hierarchy without compromising information, reflecting its aptitude to represent hierarchical data with multiple nodes at each level. This approach surpasses a parallel method that assimilates transferable visual features by encoding the semantic interplay between source and target classes \cite{li2019large}. The researchers ingeniously map both the actions and videos into this shared realm, aligning them based on their embeddings within the hyperbolic space.

Advances in research have pivoted towards harnessing 3D motion data for human action recognition. One such innovative approach is the 3D Dynamic Voxels (3DV), designed specifically for the representation of 3D motion \cite{wang20203dv}. The essence of this method is its proficiency in capturing 3D motion patterns, which is paramount for depth-based 3D action recognition. The fundamental principle of the 3DV technique is to transcode the 3D motion data from a video into a structured voxel set. Intriguingly, each voxel inherently encompasses joint 3D spatial and motion attributes. These voxels are then channeled into the Pointnet++ network \cite{qi2017pointnet++}, renowned for its streamlined architecture and efficacy in deep feature extraction.

\subsection{Comprehensive Evaluation and Performance Comparison} 
\label{sec:Holistic_Discussion}

\begin{table}[h]
\caption{This table shows the state-of-the-art models in vision-based human action recognition vision-based for Two-Stream, 3D Convolutional, and Motion Networks.}
\label{Tab:TS-3D-MN-based}
\begin{tabular*}{\textwidth}{@{\extracolsep\fill}lcccccccl}
\toprule%
& & \multicolumn{2}{@{}c@{}}{\textbf{\textit{Two-Streams}}} \\ \cmidrule{1-8}
Model Name  & Year & UCF-101\footnotemark[1]  & HMDB-51\footnotemark[2]& K400\footnotemark[3]& Backbone & Pre-train Data & Code \\
\midrule
TS-Netowrk\cite{simonyan2014two}& 2014 &88.0\%& 59.4\%& -& -& - &\href{https://github.com/feichtenhofer/twostreamfusion}{\faGithubSquare}\\
TS-Fusion\cite{feichtenhofer2016convolutional}& 2016 & 93.5\%  & 69.2\% & -& VGG-16& ImgNet &\href{https://github.com/tomar840/two-stream-fusion-for-action-recognition-in-videos}{\faGithubSquare}\\
TSN\cite{wang2016temporal}& 2016 &94.2\% &69.4\% & - & ResNet-50 & ImgNet &\href{https://github.com/yjxiong/caffe?utm_source=catalyzex.com}{\faGithubSquare}\\
TS-FCAN\cite{tran2017two}& 2017 &93.4\%  & 68.2\%& - & - & - &\href{https://github.com/antran89/two-stream-fcan?utm_source=catalyzex.com}{\faGithubSquare}\\
RHN\cite{yu2017novel}& 2017 & 93.2\%  & 71.8 &  - & - & - & - \\
Fu-2\cite{gammulle2017two}& 2017 & 94.6\%  &  - & - & VGG-16 & ImgNet & - \\
DTS-ConvNets\cite{YaminGoing}& 2018 & 95.1\%  & -  &-& ResNet-101& ImgNet & -\\
IP-LSTM+IDT\cite{ShengLearning}& 2019 & 91.4\%  & 68.2\%& - & GoogLeNet & ImgNet& -\\
TS-LSF CNN\cite{YanqinAction}& 2020 & 94.8\%  & 70.2\%& - & VGG16 & ImgNet & - \\
Distinct-TS\cite{SarabuDistinct}& 2020 & 95.0\%  & 67.9\%& - & R101+I-V2& ImgNet& -\\
PBNets\cite{WenbingToward} & 2020  & \textbf{95.4\%}  & 72.5\% & - &BN-Inception& ImgNet & - \\
UF-TSN\cite{XiaohangUnsupervised} & 2021  & 91.2\%  & 62.5\% & - &ResNet-152/18& ImgNet & - \\
BS-2SCN\cite{Zhongwen} & 2022  & 90.1\%  & 71.3\% & - & ResNet & ImgNet& -\\
TG\cite{Xiao_2022_CVPR} & 2022  & 92.1\%  & \textbf{75.9\%} & \textbf{69.2\%}&R3D-18& ImgNet&\href{https://github.com/lambert-x/video-semisup}{\faGithubSquare} \\
\cmidrule{1-8}
& & \multicolumn{2}{@{}c@{}}{\textbf{\textit{3D Convolutional}}} \\ 
\cmidrule{1-8}
Model Name  & Year & UCF-101\footnotemark[1] & HMDB-51\footnotemark[2] & K400\footnotemark[3] & Backbone & Pre-train Data & Code \\
\midrule
C3D\cite{tran2015learning}& 2014 &90.4\% &  - & - & 3D VGG-11 &  Sports-1M&\href{https://github.com/Ontheway361/C3D}{\faGithubSquare}\\
P3D ResNet\cite{qiu2017learning}& 2017 &93.7\% &  - & - & - & ImgNet&\href{https://github.com/ZhaofanQiu/pseudo-3d-residual-networks?utm_source=catalyzex.com}{\faGithubSquare}\\
ECO\cite{zolfaghari2018eco}& 2018&94.8\% &72.4\%  & - & BNInception& K400&\href{https://github.com/mzolfaghari/ECO-efficient-video-understanding?utm_source=catalyzex.com}{\faGithubSquare}\\
Two-Stream I3D\cite{carreira2017quo}& 2018& \textbf{97.0\%} & \textbf{80.2\%} & 74.2\% & Inception-V1 & ImgNet &\href{https://github.com/ahsaniqbal/Kinetics-FeatureExtractor?utm_source=catalyzex.com}{\faGithubSquare}\\
NL I3D\cite{wang2018non}& 2018& - & - &  93.3\% & ResNet-101 & ImgNet &\href{https://github.com/facebookresearch/video-nonlocal-net?utm_source=catalyzex.com}{\faGithubSquare}\\
SlowFast\cite{feichtenhofer2019slowfast}& 2019& - & - &  \textbf{93.9\%}& -& -&\href{https://github.com/facebookresearch/SlowFast}{\faGithubSquare}\\
X3D\cite{feichtenhofer2020x3d}& 2020& - & \textbf{94.6\%}& - & R50/101+NL & - &\href{https://github.com/facebookresearch/SlowFast}{\faGithubSquare} \\
DCTR\cite{ou20233d}& 2023& 95.0\%& 72.9\% & - & - & - & -\\

\cmidrule{1-8}
& & \multicolumn{2}{@{}c@{}}{\textbf{\textit{Motion Models}}} \\ 
\cmidrule{1-8}
Model Name  & Year & UCF-101\footnotemark[1] & HMDB-51\footnotemark[2] & K400\footnotemark[3] & Backbone & Pre-train Data & Code \\
\midrule
TDD\cite{wang2015action}& 2015 &91.5\% &65.9\% & - & - & -&\href{https://wanglimin.github.io/tdd/index.html?utm_source=catalyzex.com}{\faGithubSquare}\\
Rank Pooling-US\cite{fernando2016rank}& 2016 &65.8\% & - & - &-&- &\href{https://bitbucket.org/bfernando/videodarwin/src/master/?utm_source=catalyzex.com}{\faGithubSquare}\\
TOV-US\cite{misra2016shuffle}& 2016 & 50.9 \% & - & - & -& - & -\\
TRN\cite{zhou2018temporal}& 2018 & 83.8\% & - & - & -& - &\href{https://github.com/zhoubolei/TRN-pytorch?utm_source=catalyzex.com}{\faGithubSquare}\\
MotionNet-US\cite{zhu2019hidden}& 2019 & \textbf{97.1\%} & \textbf{78.7\%}& - & I3D & -&\href{https://github.com/bryanyzhu/Hidden-Two-Stream?utm_source=catalyzex.com}{\faGithubSquare}\\
TSM\cite{lin2019TSM}& 2019 & 95.9\% & 73.5\%& - & ResNet-50  & ImgNet+K400 & \href{https://github.com/MIT-HAN-LAB/temporal-shift-module}{\faGithubSquare}\\
STM\cite{jiang2019stm}& 2020 & 96.2\% & 72.2\%& - & ResNet-50 & ImgNet+K400 &\href{https://github.com/joaanna/something_else?utm_source=catalyzex.com}{\faGithubSquare}\\
TEA\cite{li2020tea}& 2020 & 96.9\% & 73.3\%& \textbf{92.5\%} & ResNet-50 & ImgNet &\href{https://github.com/Phoenix1327/tea-action-recognition?utm_source=catalyzex.com}{\faGithubSquare}\\
\end{tabular*}
\footnotetext[1]{UCF-101 Dataset: a benchmark of human action recognition \cite{soomro2012ucf101}}
\footnotetext[2]{HMDB-51 Dataset: a benchmark of human action recognition \cite{kuehne2011hmdb}}
\footnotetext[3]{Kinetics-400: a benchmark of human action recognition \cite{kay2017kinetics}}
\end{table}

\begin{table}[h]
\caption{This table shows the state-of-the-art models in skeleton-based human action recognition using Graph Convolutional Networks}\label{Tab:GCN-based_results}
\begin{tabular*}{\textwidth}{@{\extracolsep\fill}lccccccc}
\toprule%
& & \multicolumn{3}{@{}c@{}}{\textbf{\textit{Graph Convolutional Networks (GCN)}}} \\ \cmidrule{1-8}

& &\multicolumn{2}{@{}c@{}}{NTU RGB+D\footnotemark[1]} & \multicolumn{2}{@{}c@{}}{NTU RGB+D 120\footnotemark[2]} \\ \cmidrule{3-4}\cmidrule{5-6}%
Model Name  & Year & XSub & XView & XSub & XView & Parameters & Code \\
\midrule
ST-GCN\cite{yan2018spatial}& 2018 & 90.7 & 96.5 & 86.2 & 88.4 & - &\href{https://github.com/yysijie/st-gcn}{\faGithubSquare}\\
TCA-GCN\cite{wang2022skeleton}& 2022 & 92.8 & 97.0 & 89.4 & 90.8 & - &\href{https://github.com/OrdinaryQin/TCA-GCN}{\faGithubSquare} \\
STGAT\cite{hu2022spatial} & 2022 & 92.8 & 97.3 & 88.7 & 90.4 &- &\href{https://github.com/hulianyuyy/STGAT}{\faGithubSquare} \\
PSUMNet\cite{trivedi2022psumnet} & 2022 & 92.9 & 96.7 & 89.4 & 90.6 & - &\href{https://github.com/skelemoa/psumnet}{\faGithubSquare} \\
LST\cite{xiang2022language}  & 2022 & 92.9&97& 89.9& 91.2 & - &\href{https://github.com/martinxm/lst}{\faGithubSquare}\\
InfoGCN\cite{chi2022infogcn} & 2022& 93.0 & 97.1 & 89.8 & 91.2 & - &\href{https://github.com/stnoah1/infogcn}{\faGithubSquare}\\
DG-STGCN\cite{duan2022dg}  & 2022 & 93.2 & \textbf{97.5} &  89.6 & 91.3 & - &\href{https://github.com/kennymckormick/pyskl}{\faGithubSquare}\\
HD-GCN\cite{lee2022hierarchically} & 2022 & 93.4 & 97.2& 90.1& 91.6 & - &\href{https://github.com/Jho-Yonsei/HD-GCN}{\faGithubSquare}\\
SkeletonGCL\cite{huang2023graph}& 2023& 93.1 & 97.0 & 89.5 & 91.0 & - &\href{https://github.com/oliverhxh/skeletongcl}{\faGithubSquare}\\
TD-GCN\cite{liu2023temporal} & 2023 &92.8 & 96.8 & - & - & - &\href{https://github.com/liujf69/TD-GCN-Gesture}{\faGithubSquare} \\
LA-GCN\cite{xu2023language} & 2023 & \textbf{93.5} & 97.2 & \textbf{90.7} & \textbf{91.8} & - &\href{https://github.com/damnull/lagcn}{\faGithubSquare}  \\
\\ \cmidrule{1-8}
\end{tabular*}
\footnotetext[1]{Note: Here, we review the state-of-the-art models.}
\footnotetext[1]{NTU RGB+D Dataset: a benchmark of human action recognition skeleton based \cite{shahroudy2016ntu}.}
\footnotetext[2]{NTU RGB+D 120 Dataset: a benchmark of human action recognition skeleton based \cite{liu2019ntu}.}
\end{table}
In this section, we review the performance of the different networks we mentioned previously on most HAR vision-based datasets, including UCF-101, HMDB-51, K400, Kinetics-400, Kinetics-600, NTU RGB+D 60, and NTU RGB+D 120. In pioneering work on HAR, handcrafted features played a pivotal role. Researchers had relied on handcrafted features since the mid-80s, and they continued to hold substantial value till the advent of deep learning models to enhance performance. Researchers who applied the deep learning concept for the first time in human action recognition got rid of handcrafted features with single-stream networks \cite{karpathy2014large}. The single-stream network achieved \textbf{65.4\%} on the UFC-101 dataset \cite{soomro2012ucf101} while using the two-stream model for the first achieved \textbf{88.0\%} by leveraging the motion information \cite{simonyan2014two}. The two-stream networks proved that capturing spatial and temporal feature information is essential in recognizing human action in videos when dealing with a sequence of images.

Building on the insights from the subsection~\ref{subsec:two-stream}, numerous studies have used the two-stream network for HAR. Most of these works have shown great results by being evaluated on the UCF-101, HMDB-51, and Kinetics-400 datasets, as shown in Table~\ref{Tab:TS-3D-MN-based}. PBNets \cite{WenbingToward} reached the highest accuracy by evaluated on UCF-101 with \textbf{95.4\%}, while TG models \cite{Xiao_2022_CVPR} have achieved the highest accuracy on HMDB-51 with \textbf{75.9\%} and on Kinetics-400 with \textbf{69.2\%}.

Researchers used the 3D Convolutional (3DC) network in parallel and demonstrated their effectiveness on the same datasets, as mentioned in section~\ref{subsec:3D CNNs}. In 2018, Carreira et al. \cite{carreira2017quo} integrated two-stream networks with 3DC, further enhanced by pre-training on the ImageNet dataset. This approach led the Two-Stream I3D model to achieve remarkable accuracy levels \textbf{97.0\%} on UCF-101 and  \textbf{80.2\%} on HMDB-51. Since then, the exploration of 3DC networks has expanded, including the evaluation of the X3D model \cite{feichtenhofer2020x3d} in 2020, which reached a  \textbf{94.6\%} accuracy rate on Kinetics-400, and the SlowFast model \cite{feichtenhofer2019slowfast} in 2019, achieving  \textbf{93.9\%}. Using the Motion Networks, as shown in section~\ref{subsec:Motion models}, also shows remarkable works. Once again, using the two-stream network with the motion network shows good results evaluated on these datasets. MotionNet model \cite{zhu2019hidden} achieved \textbf{97.1\%} on UCF-101 and \textbf{78.7\%} on HMDB-51. None of the results of the models mentioned above involve Transformer or GCN networks.

\begin{table}[h]
\caption{Accuracy comparison of the Transformer-based HAR models}\label{Tab:Transformer-based-models}
\begin{tabular*}{\textwidth}{@{\extracolsep\fill}lcccccccccc}
\toprule%
& &\multicolumn{2}{@{}c@{}}{NTU-60} & \multicolumn{2}{@{}c@{}}{NTU-120} &
\\\cmidrule{3-4}\cmidrule{5-6}%
Model Name  & Year & XSub & XView & XSub & XView & Parameters& Code \\
\midrule
Action Transformer~\cite{girdhar2019video} & 2019  & - & - &- &-&-&\href{https://rohitgirdhar.github.io/ActionTransformer/}{\faGithubSquare} \\
ST-TR~\cite{plizzari2021spatial}  & 2021 & 89.9 & 96.1 & 84.3 & 86.7 &-&\href{https://github.com/Chiaraplizz/ST-TR}{\faGithubSquare}  \\
ST-TR-agcn~\cite{plizzari2021skeleton} & 2021 &90.3&96.3  &85.1&87.1&-  &\href{https://github.com/Chiaraplizz/ST-TR}{\faGithubSquare}  \\
STAR-128~\cite{shi2021star} & 2021 & 83.4 & 89.0 & 78.3 & 80.2 &1.26M&\href{https://github.com/imj2185/STAR}{\faGithubSquare}  \\
STST~\cite{zhang2021stst}  & 2021 & 91.9 & 96.8 &-  &-  &-& -  \\
3D-D Transformer\cite{kim2022cross} &2022  & 94.3 & \textbf{97.9} & \textbf{90.5} & \textbf{91.4}&-&- \\
RGBSformer~\cite{shi2023novel} & 2023 & 91.1 & - & 85.7 &-& -&-\\
STAR-Transformer~\cite{ahn2023star} & 2023 & 92.0 & 96.5 & 90.3 & 92.7 &-&-   \\
ViLP~\cite{chaudhuri2023vilp} & 2023 & - &-&-&-&-&- \\
VT-BPAN~\cite{sun2023integrating} & 2023  & \textbf{95.4} &97.7 &86.7 &88.6  &-&-  \\
\midrule
& & \multicolumn{3}{@{}c@{}}{\textbf{\textit{Transformer Models}}} \\ \cmidrule{1-8}
& & \multicolumn{2}{@{}c@{}}{Kinetics-400} & \multicolumn{2}{@{}c@{}}{Kinetics-600} &
\\\cmidrule{3-4}\cmidrule{5-6}%
Model Name  & Year & Top-1 & Top-5 & Top-1 & Top-5 &Parameters& Code \\
\midrule
CoVeR~\cite{zhang2021co} & 2021 &87.2 &-&87.9&-&431.0M&-  \\ 
MViT-B~\cite{fan2021multiscale} & 2021 &81.2&95.1&84.1&96.5 &36.60M&\href{https://github.com/facebookresearch/SlowFast}{\faGithubSquare} \\
ST-TR~\cite{plizzari2021spatial} & 2021  &37.0&59.7&-&-&-&\href{https://github.com/Chiaraplizz/ST-TR}{\faGithubSquare}  \\ 
ST-TR-agcn~\cite{plizzari2021skeleton} & 2021 &38.0&60.5&-&-&-&\href{https://github.com/Chiaraplizz/ST-TR}{\faGithubSquare}  \\ 
STST~\cite{zhang2021stst} & 2021 & 38.3&61.2&-&-&-& -  \\
TimeSformer-L~\cite{bertasius2021space} &2021 &80.7&94.7&82.2&95.6&121.4M&\href{https://github.com/facebookresearch/TimeSformer}{\faGithubSquare} \\
ViViT-H~\cite{arnab2021vivit} & 2021 &84.9&95.8&85.8&96.5&310.8M&\href{https://github.com/google-research/scenic}{\faGithubSquare}  \\ 
AcT-M~\cite{mazzia2022action} & 2022  & - & - & - &- &2.743K&- \\
DualFormer-B~\cite{liang2022dualformer} & 2022 &82.9&95.5&85.2&96.6 &86.8M&\href{https://github.com/sail-sg/dualformer}{\faGithubSquare} \\
MM-ViT IV~\cite{chen2022mm} & 2022 &-&-&83.5&96.8&158.6M&-\\ 
MTV-H~\cite{yan2022multiview} & 2022 &89.1&98.2&89.6&98.3 &1000+M&\href{https://github.com/google-research/scenic}{\faGithubSquare} \\
MViTv2-L~\cite{li2022mvitv2} & 2022 &86.1&97.0&87.9&97.9&217.6M &\href{https://github.com/facebookresearch/mvit}{\faGithubSquare} \\
Swin-L~\cite{liu2022video} &2022 &84.9&96.6&85.9&97.1&200.0M&\href{https://github.com/SwinTransformer/Video-Swin-Transformer}{\faGithubSquare} \\
UniForm-B~\cite{li2022uniformer} & 2022 &83.0&95.4&84.9&96.7 &-&\href{https://github.com/Sense-X/UniFormer}{\faGithubSquare} \\
UniFormV2~\cite{li2022uniformerv2} & 2022 &90.0&98.4&90.1&98.5 &354.0M&\href{https://github.com/OpenGVLab/UniFormerV2}{\faGithubSquare} \\
X-CLIP-L/14~\cite{ni2022expanding} & 2022  &87.7&97.4& 88.3& 97.7&-& \href{https://github.com/microsoft/VideoX/tree/master/X-CLIP}{\faGithubSquare}  \\
BIKE~\cite{wu2023bidirectional}  & 2023 &88.6&98.3&-&-&230.0M& \href{https://github.com/whwu95/BIKE}{\faGithubSquare} \\
Hiera-H~\cite{ryali2023hiera} & 2023 &87.8&-&88.8&-&672.0M &\href{https://github.com/facebookresearch/hiera}{\faGithubSquare}   \\
L-SPM8/14/18~\cite{pan2023svt} & 2023 &85.0&96.5&-&-&304.0M&-  \\
Text4Vis~\cite{wu2023revisiting} & 2023 & 87.8&97.6&-&-&230.7M&\href{https://github.com/whwu95/Text4Vis}{\faGithubSquare} \\
TubeViT-H~\cite{piergiovanni2023rethinking} & 2023 &90.9&\textbf{98.9}&\textbf{91.8}&98.0&635.0M &\href{https://sites.google.com/view/tubevit}{\faGithubSquare}  \\
UMT-L~\cite{li2023unmasked}& 2023  &90.6&98.7&90.5&\textbf{98.8}&304.0M& \href{https://github.com/OpenGVLab/unmasked_teacher}{\faGithubSquare}  \\
VideoMAE~\cite{tong2022videomae} & 2022 &87.4&97.6&-&-& 633.0M& \href{https://github.com/MCG-NJU/VideoMAE}{\faGithubSquare}  \\
VideoMAE V2~\cite{wang2023videomae} & 2023 &90.0&98.4&89.9&98.5& 1B& \href{https://github.com/OpenGVLab/VideoMAEv2}{\faGithubSquare}  \\
OmniVec~\cite{srivastava2024omnivec} & 2024 & 91.1 &- & - & - & - & - \\
OmniVec2~\cite{srivastava2024omnivec2} & 2024 &	\textbf{93.6} & - & - & - &-& - \\
InternVideo2~\cite{wang2024internvideo2} & 2024 &92.1& - & - & - &-& \href{https://github.com/OpenGVLab/InternVideo/tree/main/InternVideo2}{\faGithubSquare} \\
\\ \cmidrule{1-8}
\end{tabular*}
\end{table}

Research in HAR has witnessed remarkable progress in utilizing Graph Convolutional Networks (GCNs) to gather information on the human skeleton structure. As discussed in section~\ref{subsec:GCNs},  researchers use the GCN approaches to capitalize on the structural nuances of the human body's skeleton to encode dynamic joint locations, offering a more nuanced representation of human movements. Unlike traditional methodologies, GCN algorithms excel in deciphering direct and indirect correlations within these complex skeletal connections, unveiling previously overlooked patterns. The performance of GCN-based techniques is usually evaluated using NTU RGB+D \cite{shahroudy2016ntu} and NTU RGB+D 120 \cite{liu2019ntu} benchmarks due to the skeleton information provided in these datasets. As delineated in Table~\ref{Tab:GCN-based_results}, the LA-GCN model exemplifies the pinnacle of achievement within this domain, leveraging the Large Language Model to attain unparalleled accuracy. It has achieved groundbreaking accuracy rates of \textbf{93.5\%} and \textbf{97.2\%} on the NTU RGB+D 60 dataset, alongside \textbf{90.7\%} and \textbf{91.80\%} on the NTU RGB+D 120, setting new benchmarks for state-of-the-art results in HAR through GCNs.

As mentioned in section~\ref{sec:Transformer Models}, the Transformer network architecture shows great results using the skeleton data on the NTU RGB+D 60/120 datasets \cite{shahroudy2016ntu} \cite{liu2019ntu} as well as using RGB-based dataset on the Kinetics series (400, 600, and 700). The VT-BPAN model achieved the best performance on the NTU RGB+D 60 dataset with an accuracy of \textbf{95.4\%} for the Cross-Subject (XSub), which is training data and validation data collected from different subjects. The 3D Deformable provided \textbf{97.9\%} for the Cross-View (XView), which is training data and validation data 
collected from different camera views. However, the LA-GCN model still wins with the NTU RGB+D 120. The TubeViT-H and UMT-L Models \cite{piergiovanni2023rethinking}\cite{li2023unmasked} obtain the best results on the Kinetics series of datasets. The TubeViT-H model produced  \textbf{90.9\%} accuracy for the XSub and \textbf{98.9\%} the XView on Kinetics-400 and \textbf{91.8\%} for the XSub on Kinetics-600, which the UMT-L Model provide \textbf{98.8\%} for XView on Kinetics-600.

Our comprehensive survey has reviewed numerous papers in HAR using different network architectures, and most of these methods aim to extract spatial and temporal features from image sequences to enhance performance. The two-stream network advances this objective by capturing the spatial features from individual frames and temporal features from sequences separately before fusing these streams. Similarly, 3D convolutional networks extend the convolution operation into the time dimension, facilitating the extraction of patterns across spatial and temporal dimensions from sequential data. GCNs capture the spatial and temporal relationships through dynamic interactions by modeling the human body as a graph, where nodes and edges represent body parts and their connections. The Transformer model is utilized for RGB-based and skeleton information by capturing the long-range dependencies and intricate relationships, thus offering a nuanced comprehension of spatial and temporal dynamics. This is achieved by harnessing the model's sequencing capabilities through pixel-based or skeletal data.

However, HAR's landscape can improve using new networks, methods, and datasets. After over a decade of prolific research in this domain, the question remains: \say{What is next?}. Researchers have now started leveraging Large Language Models (LLMs) to improve results, and the LA-GCN~\cite{xu2023language} has successfully achieved state-of-the-art performance by incorporating these advancements. Another question to pose is, \say{Could providing additional information be helpful?} Several studies emphasize hand pose information's significance in enhancing outcomes, with PBNets~\cite{WenbingToward} achieving state-of-the-art performance on the UCF-101 dataset, particularly due to its comprehensive coverage of musical instrument classes that require detailed hand movement analysis. Researchers are increasingly adopting Explainable Artificial Intelligence (XAI) techniques to gain insights into what models have learned from the data \cite{mersha2024explainable}. Researchers should select an architecture that aligns with their specific problem requirements and provides critical information to enhance the accuracy and generalizability of HAM models.

\section{Open-HAR} \label{sec:Open World Recognition}
Open-HAR is an emerging problem area in the HAR field, challenging networks through the introduction of novel action classes which must be detected and rejected (in the Open-Set setting) or detected and learned (in the Open-world setting). Relatively few works have proposed solutions to Open-HAR problems. We provide a comprehensive overview of the problem, novelty selection, and existing approaches for Open-HAR systems.
\subsection{Background}
Open-world recognition \cite{bendale2015towards} was first introduced by Bendale et al. in 2015 and applied to Image Recognition using a new open-world evaluation protocol. Open-world recognition is a natural extension of Open-set recognition \cite{scheirer2012toward, scheirer2014probability}, where samples at test-time can come from known and unknown classes (those which were not seen during training). This concept has evolved into Open-world recognition through the addition of a labeling process and incremental learning stage wherein unknown samples are detected, labeled, and learned. Open-set only requires a system to detect and reject unknown samples, but Open-world has the additional constraint of learning new classes from which the unknowns belong to. 
For systems to be effective at the detection of unknown open-set samples, they must strike a balance between empirical risk and open-space risk \cite{bendale2015towards}. Open-world recognition is an emerging evaluation setting within HAR.

\begin{figure}[hbt!]
\centerline{\includegraphics[width=13cm]{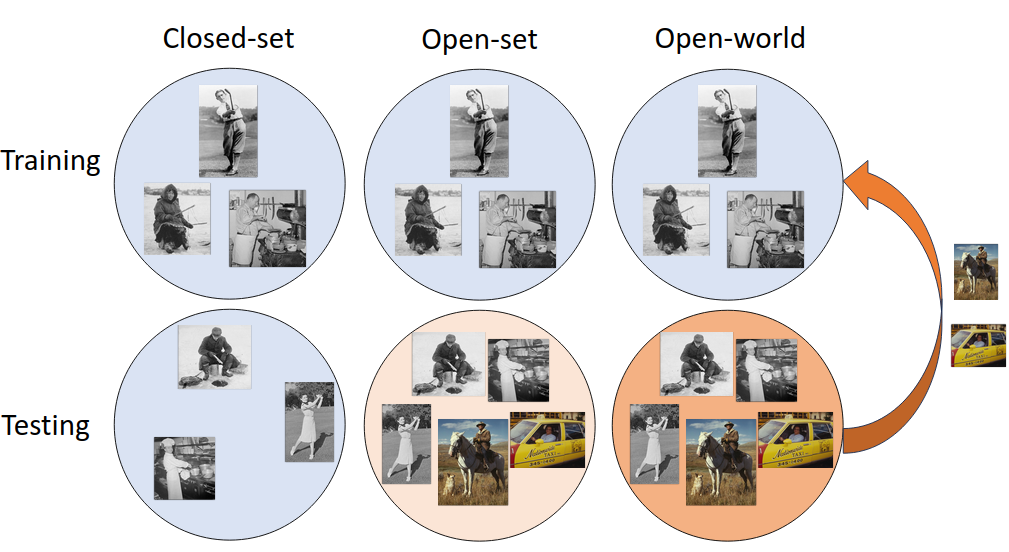}}
\caption{Closed-set classification (left) assumes only samples belonging to known classes will appear at test time (ice-fishing, golfing, cooking). Open-set requires systems to reject samples from unknown classes at test time (horseback riding, driving). Open-world learning (right) is an iterative process where unknowns encountered at test time must be detected and learned.}
\label{fig:OpenHAR}
\end{figure}

The vast majority of HAR systems do not experiment with incremental learning, continual learning, Open-set, or Open-world recognition.
While labeled HAR datasets (as discussed in Section \ref{sec:datasets}) have grown to include a wide variety of actions and sample data, no dataset can comprehensively describe all possible human actions.
Of course, this is true for open-world evaluations across most recognition problems, including Image Recognition, Object Detection, and Facial Recognition.
However, an open World (like an open set) is a practical problem with a grounded objective: deploying robust (multi-class) recognition systems that can handle novelty. A dataset that encompasses the entirety of open-set classes for a problem is a theoretical ideal; instead, Open-set and Open-world systems can be evaluated using unknowns that are representative of the (sometimes only circumstantially related) novelty they may encounter once deployed.
However, no large-scale protocols for Open-set and Open-world evaluations have been widely adopted, which hinders the development and comparison of HAR systems designed for these tasks.

\subsection{A Foreword on Novelty}
Boult et al.~\cite{boult2021towards} proposed major refinements of the Open-world recognition problem via the formalization of novelty. 
Notably, the authors identified three major spaces where novelty occurs: World novelty, Observation novelty, and Agent novelty. 
Prijatel et al.~\cite{prijatelj2022human} related these and other observations from the foundational Open-world recognition and novelty theory works \cite{bendale2015towards, boult2021towards, shrivastava2023novelty} to the HAR space. 
This work is one of the first major steps in transferring the knowledge and theory of Open-world recognition from the general domain into HAR-specific problem spaces. 
But both these works offload the distinction between novel and known samples to a ``Dissimilarity function."
While the unified novelty framework is useful for describing the \textit{kind} of novelty in terms of which space or combinations thereof it occurs in as determined by regret and dissimilarity measures, it offers no concrete suggestion to a practitioner wondering ``What is a novel class?" and for good reason.
As Prijatel et al.~\cite{prijatelj2022human} bluntly state, ``Fundamentally, novelty is anything that is new to something else."
The vagueness of novelty is necessary in most cases because the specificity of ``known" semantically deteriorates as it is discussed with increasing granularity.

Suppose a recognition system is trained to recognize all the animals one might find in a zoo --llamas, bears, tigers, etc-- they are all regarded as known.
Given this simple example, it is natural to assume novelty here is ``not animals one might find in a zoo."
This relation appears concrete, but the circular dependency between novelty and known is only grounded insofar as ``animals one might find in a zoo" is defined.
Does the category ``bears" include all subtypes of bears (black, polar, grizzly, and others)?
From a purely taxonomic standpoint, perhaps, but what of the extinct California grizzly bear?
Do knowns include everything taxonomically defined as a bear (Ursidae) that has ever been in a zoo or only in modern zoos?
What about tigers? A Thylacine is a marsupial, yet colloquially referred to as a Tasmanian Tiger; whether such an animal is known or novel, tiger or not tiger, depends on which infinitely granular definition of ``animals one might find in a zoo" is assumed.
Of course, we can easily answer these questions by examining the contents of the known dataset's tiger or bear class and looking for these animals, but this is entirely impractical.
The task of choosing a dataset of unknowns to test a system's robustness, as would a real-world deployment, is seldom approached on a picture-by-picture basis.
Instead, the question is often \textit{what \textbf{kind} of dataset is known? What \textbf{kinds} of classes are in this dataset? Do those unknown classes \textbf{overlap} with these known classes?}
This is the root of the semantic issues with selecting unknowns, the loss of specificity when trying to describe what a given network saw during training.

This problem is compounded when applied to HAR, which must recognize actions over time instead of recognizing nouns.
By necessity, the exact meaning of ``human action" is entirely vague.
If ``swimming" is a known class of human action, does this include only well-recognized swimming techniques (freestyle, backstroke, butterfly), or does it extend to the flailing motions someone might make when swimming for the first time?
What about when one becomes exhausted from swimming and can no longer effectively keep their head above water?
Are they swimming or drowning? Where does one end and the other begin?
If someone is playing soccer, a common human action is kicking, but kicking to score a goal is entirely different than kicking to pass the ball to another player.
Indeed, both ``actions" involve kicking, but one involves a lateral motion where the ball contacts the side of the shoe, while the other involves a medial motion where the ball contacts the shoe laces.

This problem of semantic decomposition is highly related to fuzzy sets \cite{zadeh1965fuzzy}, where a membership function determines class inclusion. 
At least one relevant work \cite{wu2022open} has shown that fuzzy minmax networks \cite{simpson1992fuzzy} can be used for Open-set recognition.
However, that work evaluated unknown rejection utilizing unknowns, which were very different from the known training set and did not delve into the semantic concerns of dataset construction, which we've discussed here.

Without precisely defining what a known action is, we cannot concretely define what novel actions are. It is no wonder that using language that is implicitly imprecise to define what is known results in frustratingly arbitrary bounds. The usual approach adopted by OSR works \cite{rudd2017extreme, neal2018open, perera2020generative, zhou2021learning,yang2020convolutional, vaze2021open} is to divide an existing dataset into known and unknown classes, that is, to train with only a fraction of classes in a given dataset and reserve the remaining fraction for simulating unknowns at test time. By splitting a dataset into knowns and unknowns, we can guarantee both are thematically related by the methodology used to design the original dataset. This entirely avoids the problem of defining knowns (beyond the literal content of the dataset) and exploits the fact that categories in single-label datasets are inherently unique to each other as far as classification is concerned. 

\subsection{Approaches to Open-Set HAR}

The multi-class Open-Set Recognition (OSR) function is a key component of any Open-world system. This function enables a system to detect novel samples and differentiate them from known samples of known classes. 
Bao et al.~\cite{bao2021evidential} proposed Deep Evidential Action Recognition (DEAR), which presents an Open-set HAR system. 
DEAR uses an evidential neural network with uncertainty calibration to augment an action recognition backbone, forming a system that is robust to unknown samples. 
Zhao et al. \cite{zhao2023open} proposed an open-set HAR system based on evidential learning, Multi-Label Evidential Learning (MULE), motivated by a multi-actor/action problem setting. 
This is a subtly different problem than single-label action recognition, where classes are inherently mutually exclusive. 
It opens the door to simultaneous unknown novelty where an actor could perform known actions while simultaneously performing novel ones. 
MULE makes several improvements over DEAR in terms of evidential learning, which enables it to handle action recognition from one or more actors and actions. 
Feng et al. ~\cite{feng2023spatial} realized the success of Guo et al. \cite{guo2021conditional} in adapting capsule (CAPS) networks for OSR and extended them to Open-set HAR by proposing Spatio-Temporal Exclusive CAPS network (STE-CapsNet). 
Authors utilized DEAR as a backbone and extended the routing mechanism employed in CAPs networks to utilize spatio-temporal information needed to understand video-based representations. 
For evaluation, authors adapted STE-CapsNet to various successful HAR networks and compared them against other Open-set recognition methods. In nearly every setting, STE-Caps showed improvement in terms of F1 Score.

Gutoski et al. ~\cite{gutoski2021deep} examined how triplet-loss affects I3D-based networks for HAR in the OSR setting. While the basis is somewhat similar to Shu et al. \cite{shu2018odn}, this work focuses on the effect of triplet-loss rather than the overall effectiveness of an OSR system. Triplet loss traditionally utilizes sets of three samples, an anchor, a positive, and a negative for a given class, and enforces a distance-driven penalty. By using a distance-based loss, a network's feature space can ideally relate samples at inference time using the distance metric. Some of the earliest deep learning approaches to Open-set recognition are based on this distance between samples in feature space \cite{rudd2017extreme,bendale2016towards}. Indeed, the authors took note and added an Extreme Value Machine (EVM) on top of their triplet-loss trained network, underpinning the distance-based assumptions the EVM relies on. Notably, authors avoided tuning with their test set and adopted a combination of hyperparameters proposed by the EVM \cite{rudd2017extreme} and their own intuition. They validated their work by splitting the UCF101 dataset into known and unknown, then evaluating the performance of variations of I3D network training with an (identically configured) EVM using mainly Macro-F1 score. The authors compared results using robust statistical testing and found significant improvement in the triplet-loss trained network. Lee et al.~\cite{lee2022sensor} later proposed mixup triplet learning for Malhalanobis distance (MTMD), which uses mixup samples with triplet loss and a Malhalanobis distance metric to shape the decision boundaries of known classes. A number of OSR works attempt to synthesize samples outside known class decision boundaries \cite{zhou2021learning, kong2021opengan, neal2018open} Authors here use mixup to create triplets; mixup is a natural augmentation to apply for Open-set; as mixup authors note, it can reduce undesirable behavior when predicting samples outside the training set \cite{zhang2017mixup}. PROSER \cite{zhou2021learning} proposed a similar approach about a year prior, using manifold-mixup \cite{verma2019manifold} instead, and presented some interesting arguments in favor of the switch.

Yu et al. ~\cite{yu2020action} approached OSR from a representation-matching perspective. Several OSR works have proposed using prototypes \cite{yang2020convolutional, shu2020p, lu2022pmal}, reducing the multi-class classification side of the problem to representation matching.
However, Yu et al. are the first to propose such a method with 3D convolutions. The authors propose a 3D convolution model inspired by autoencoders, reconstructing a series of frames from the input video. A second branch of the model takes the frames --represented by encoded features-- and uses softmax to classify them into known action labels. These representations are also used to build an ``action dictionary," which essentially acts as a prototype catalog, used at inference time for sample matching. While this work labels itself as open-set, it does not draw from the widely accepted open-set and open-world works \cite{scheirer2012toward, bendale2016towards, bendale2015towards}, and instead considers open-set to be an unconstrained evaluation. In common terminology, this work explores open-set recognition, but the evaluation is a blend between OSR and transfer learning in HAR, starting with the design of an OSR function.
Authors design their evaluation with an initial learning phase on Kinetics-600, then 30\% of known classes in the UCF101 and HMDB51 datasets are reserved as unknowns, while the remaining 70\% are used to build an action dictionary. Performance was reported in terms of mean class accuracy.

Wu et al. ~\cite{wu2022open} explored the use of a fuzzy min-max neural network (FMMNN) \cite{simpson1992fuzzy} as a post-training module that adapts a 3D CNN for OSR. Post-training OSR algorithms can usually be implemented on new backbone networks in a straightforward manner because they do not interfere with the training procedure, which often varies from network to network. The first attempt to adapt deep neural networks for OSR \cite{bendale2016towards} was also a post-training algorithm. Wu et al. selected R(2+1)D \cite{tran2018closer} as their backbone, leveraging it as a feature extractor. The features are then fed into an FMMNN, which (among other qualities) can learn non-linear decision boundaries between classes and represent new classes without experiencing catastrophic forgetting \cite{simpson1992fuzzy}. These qualities make FMMNNs seem like ideal candidates for OSR or even Open-World Learning (OWL). However, as with all existing post-training OSR algorithms, the model's descriptive power cannot increase beyond what was initially learned from training data. That is, the convolutions and relations that were learned from the backbone's training process determine the separability of known and incrementally learned classes. While the implications of this limitation are not well studied, it does raise doubts about what \say{kind} of unknowns post-training OSR algorithms can detect. Nonetheless, authors extend the existing FMMNN to use an ``Open Fuzzy Membership Function". The authors evaluated their 3D-CNN-adapted FMMNN system by training on mouse actions and evaluating open-set performance on human actions (UCF101).

Yang et al. ~\cite{yang2023leveraging} investigated attribute learning for open-set HAR. Utilizing UCF101 and attribute labels generated by the THUMOS challenge, the authors trained a single-stream CNN with a spatial, temporal attention module with class and attribute level losses. The proposed system relied on learning a class-attribute relation matrix to determine novelty. By design, this work takes an interesting approach to OSR. While most approaches discussed herein rely on detecting new classes (i.e., splitting a dataset into known/unknown classes), this work attempts to learn attributes associated with each class. It verges on the question \say{what is a novel action?} Although the evaluation was standard (train on UCF101, test unknowns from HMDB51 and another dataset), and attributes are leveraged to enrich class learning, the design of this network is ideal for addressing more subtle novelty. For example, if eating rice is a known action and a network is trained on images from traditional American and Chinese dining, where attributes may be a spoon or chopsticks when shown an image from Malaysian dining (where it is customary to eat rice with the right hand), should the network detect novelty? This kind of entity interaction novelty was specifically explored by the Something-Else dataset \cite{materzynska2020something}, which would've been interesting to apply here.

Yang et al. ~\cite{yang2019open} approached OSR from a negative generation perspective. Like multiple OSR works \cite{ge2017generative, ditria2020opengan, kong2021opengan, moon2022difficulty}, authors sought to use GANs to generate unknown samples. In this work, the use is straightforward; GAN-generated samples are used to ``fool" the HAR network. Due to the complexities of image-based resolution, using GANs in OSR has traditionally been limited to small-scale datasets or resolutions. This would naturally compound for video-based HAR, but here, authors focus on micro-Doppler signals, which are less complicated than RGB images. Like many early OSR papers, the evaluation is based on the F1 measure and varied openness, a balance of known training and known/unknown testing classes. Si et al. ~\cite{si2022open} also examined the OSR micro-Doppler HAR problem. They proposed a simplified version of OpenMax \cite{bendale2016towards} using cosine distance between class centroids and samples to detect unknowns. Authors also train their underlying CNN with Large Margin Cosine Loss \cite{wang2018cosface} to enforce a distance metric relationship in feature space, a shared goal with \cite{gutoski2021deep}

Zhai et al. ~\cite{zhai2023soar} proposed scene-debiasing open-set action recognition (SOAR). Their motivation stems again from the unrelenting question \say{what is a novel action?} Authors specify ``scene bias" as an obstacle to open-set HAR systems, when an unknown action occurs within a background which is similar to the background of a known class, the sample may be labeled as known. Although the authors do not discuss novelty at length, the problem of scene bias is clear. If swimming is a known action consisting of videos of people swimming in pools, is a video of someone swimming in an ocean novel? From the design of SOAR, authors seek to remove bias from networks by adversarial training against scene backgrounds. In other words, they consider a known action in a novel environment not to be novel. Likewise, a novel action in a known environment is novel. Nonetheless, the authors adopted the evaluation protocol from DEAR \cite{bao2021evidential} and showed some improvement in performance.

Zhang et al. ~\cite{zhang2023learning} were motivated by the concept of reconstruction models for OSR. They recognized the limitations of supervised learning and proposed utilizing reconstruction loss, an unsupervised learning technique, to enhance the feature representation of evidential neural networks. This goal of combining reconstruction loss with supervised loss was also explored in \cite{yu2020action}. The system consisted of an action recognition backbone, a reconstruction model, and an uncertainty-aware classifier. By comparing AUROC scores, the authors concluded that such a system utilizing a normalizing flow reconstruction model and evidential deep learning for uncertainty-aware classification were the most effective configurations. Du et al. \cite{du2023reconstructing} propose a multi-feature view graph autoencoder for open-set HAR. Just as prior papers were motivated by reconstruction models, Du et al. leverage graph-based feature reconstruction. Multiple feature views are generated by using max and average pooling methods on a 3D CNN backbone. These features form an undirected, unweighted graph representing similarity and temporal distance. Then, the graph autoencoder estimates an adjacency matrix and is penalized against the ground-truth adjacency matrix found from the original features.

\subsection{Approaches to Open-World HAR}
While HAR systems have yet to widely adopt the Open-world problem as a necessary obstacle to real-world deployment, significant development of the theory has already taken place. The seminal work for Open-world learning (OWL) in computer vision \cite{bendale2015towards} defined the requirements for an Open-world learning system as having three distinct characteristics: a multi-class Open-set recognition function, a labeling process for labeling detected unknown data, and an incremental learning function which learns new classes. 
The authors then proposed a system, Nearest Non-Outlier, which satisfies these requirements and evaluated it on a popular large-scale image recognition dataset.

Shu et al. ~\cite{shu2018odn} first extended open-world learning to the HAR domain.
Authors proposed a process for Opening Deep Networks (ODN), which consists of a triplet thresholding scheme for known acceptance and unknown rejection, a method for extending the classification weights for incrementally learned classes by augmenting weights of known classes, and a learning rate rule which enables incrementally added classes to be learned quicker than known base classes.
The accompanying evaluation used UCF101, selecting 50 known classes and incrementally adding the remainder as unknown.
While ODN represents HAR's first foray into the open-world, the evaluation proposed is not well defined.
Authors mention reporting accuracy in multiple settings, but accuracy over incremental learning phases gives no indication of how reliably unknowns are being detected and submitted to the labeling process.
Separately, no mention is made of what backbone network ODN adapted, which is a major barrier to reproducibility. Authors later extended this work to include prototypes of known classes and distance-based unknown detection in P-ODN \cite{shu2020p}

Prijatel et al.~\cite{prijatelj2022human} formalized Open-world HAR by adapting the unified novelty framework \cite{shrivastava2023novelty}.
They also defined an Open-world learning experimental protocol (KOWL-718) based on the popular Kinetics datasets. 
This is a major improvement over the initial UCF101 splits proposed in \cite{shu2018odn}, as KOWL-718 includes roughly 7x the number of classes. 
The authors also provided a clear methodology for how Kinetics dataset versions were combined to form KOWL-718. To evaluate systems using this protocol, authors used the unsupervised OWL task \cite{jafarzadeh2020open}, which differs from the originally proposed OWL task \cite{bendale2015towards} as no external labeling method is used for unknowns. 
In this problem setting, after initial training, a system must identify unknowns amongst known samples, cluster them to determine what unknown class samples belong to, and incrementally learn these classes. 
The authors adapted two strong HAR backbones (X3D and TimeSformer) for OWL and evaluated several variants on KOWL-718 to characterize the performance of existing HAR works as novelty is added over time. 
Performance was reported regarding arithmetic NMI, Mathews correlation coefficient, accuracy, and novelty reaction time. 
It should be noted that \cite{jafarzadeh2020open} specifically argued against NMI, particularly when used in settings where there are only a small number of samples per class. While KOWL-718 has many classes, at test time, some increments have fewer samples than the total number of classes, so the NMI reported here should be interpreted with caution.

Gutoski et al. ~\cite{gutoski2023unsupervised} also proposed an unsupervised OWL HAR system. The system itself is a modified combination of prior works, I3D \cite{carreira2017quo}, distance metric triplet learning for HAR \cite{gutoski2021deep}, and the DM-EVM \cite{gutoski2021incremental}. 
These components essentially act as a representation extractor, open-space risk minimizer, and incremental learner. 
Like Shu et al. \cite{shu2018odn}, authors proposed an evaluation protocol based on splits of the classic HAR dataset UCF101. 
They also used a variety of performance metrics, improving upon \cite{shu2018odn, shu2020p} where only accuracy was reported. Different metrics were employed for the different phases of open-world unsupervised learning.
For the open-set phase --the identification of unknown samples amongst known classes-- Youden's index \cite{youden1950index} was used.
Youdens index is not an ideal metric for open-set recognition because it is inherently binary, describing only the balance between known detection and unknown rejection. Still, the challenge in open-set is balancing unknown rejection while maintaining multi-class classification accuracy. For the unsupervised clustering task, the authors used the Normalized Mutual Information (NMI) score. For the incremental learning task, authors used two metrics, Forgetting and Inter-Task Intransigence, which are traditional metrics for incremental learning  \cite{gutoski2021incremental}. Both measures rely on NMI and describe the change in NMI over incremental learning tasks. As argued by \cite{jafarzadeh2020review}, NMI is not well-suited for open-world evaluations, where there may be many classes but few samples in each stage, or unknowns may be misclassified as known or clustered into subsets of unknown classes. To this end, Jafarzadeh et al. \cite{jafarzadeh2020review} proposed the open-world metric (OWM), but OWM has yet to be adopted in OWL-HAR.

\subsection{Discussion}
Open-HAR is an emerging problem area with scattered evaluation protocols and metrics. Nearly all works discussed here (except ~\cite{yang2023leveraging}) simulate novelty in HAR by splitting a dataset into known and unknown classes. Though we have not yet seen \textbf{consistent} adoption of any Open-set evaluation, many works discussed use UCF101, and we have seen a couple of works adopt the DEAR ~\cite{bao2021evidential}, evaluation protocol for open-set using datasets UCF101, HMDB51 and MiT-v2 \cite{feng2023spatial, zhai2023soar}.
Open-world recognition has only been explored by a few HAR works, with the largest-scale and most well-defined being KOWL-718 \cite{prijatelj2022human} Metrics used in both open-set and open-world evaluations for HAR also vary, utilizing AUROC, F1, closed-set accuracy, NMI, Youden's index \cite{youden1950index}, forgetting, and inter-task intransigence, yet none have adopted OWM \cite{jafarzadeh2020review}. Metrics are particularly troubling for Open-HAR because metrics for general Open-set or Open-world problems are still developing.

In order for Open-HAR to grow, a consistent evaluation protocol and a set of metrics must emerge. KOWL-718 has the potential to provide a large-scale common baseline for open-world HAR systems, but only if future works adopt it. To our knowledge, no such large-scale evaluation protocol exists for Open-set HAR; instead, UCF101 is commonly used, where merely 101 classes exist.

\section{Datasets used in HAR Experiments} \label{sec:datasets}

To produce an effective Deep Learning (DL) model, it is essential to use a large-scale dataset for training. Large datasets have significantly improved research quality in Human Action Recognition due to the wide range of action categories associated with various objects. Examples of these action categories are shown in Table \ref{tab:Examples} with different data modalities, illustrating the diverse range of objects connected with the actions. Numerous datasets with various modalities have been used in HAR due to the ease of collecting videos and their availability. Data modalities can be divided into two main categories based on the representation of the data, including visual and non-visual modalities, as outlined by Sun et al. \cite{sun2022human}. 

\begin{figure}[hbt!]
\centerline{\includegraphics[width=13cm]{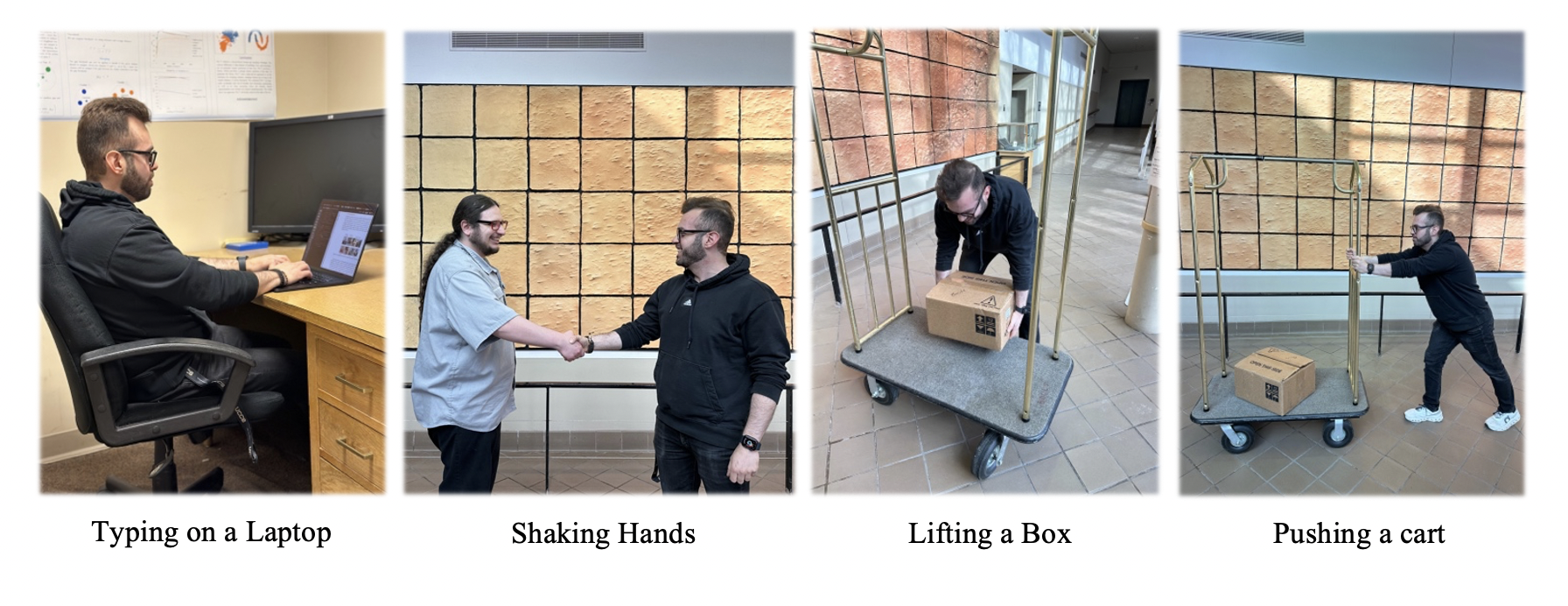}}
\caption{Examples of Human Action Recognition from videos performed in our lab at the University of Colorado Colorado Springs (UCCS).}
\label{fig:MP2}
\end{figure}

Visual modalities, which have been the focus of works surveyed in this paper, encompass RGB data \cite{kay2017kinetics}, skeletal data \cite{yousefi2017survey}, depth information \cite{xia2012view}, infrared sequences \cite{gao2016infar}, point cloud data \cite{cheng2016orthogonal}, and event streams \cite{calabrese2019dhp19}; these are characterized by their distinct visual attributes that enable the depiction of human actions within video footage \cite{liu2016benchmarking}. Utilizing these modalities has proven to be particularly effective in HAR due to their ability to convey intricate motion details and spatial dynamics \cite{poppe2010survey}.

Non-visual modalities like audio \cite{ofli2013berkeley}, acceleration data \cite{kwapisz2011activity}, radar signals \cite{chakraborty2022diat}, and WiFi data \cite{wang2019temporalunet}, distinct from visual forms, offer privacy-friendly options in specific HAR scenarios. However, these have not been leveraged much in the literature surveyed in this paper. However, these modalities are increasingly favored for their ability to protect participant identity while still providing valuable information. Advancements in deep learning now allow for the automatic extraction of features from raw sensor data sourced from devices like accelerometers, gyroscopes, magnetometers, and microphones \cite{zhu2018deep}. Some HAR research leverages both self-collected and widely available public datasets, such as USI-HAD, UCI Smartphone, and Heterogeneous datasets, to enhance experiment validity and applicability \cite{jiang2015human,almaslukh2017effective,yao2017deepsense}.

\begin{table}[ht]
  \caption{Stimuli Category Explanations}  \label{tab:Examples}
  \begin{tabular}{|c|c|c||c|c|c|} 
  \hline
  \multicolumn{3}{|c||}{Visual} & \multicolumn{3}{|c|}{Non-Visual}   \\
  \hline
    
    Modality & Action & Example & Modality & Action & Example \\
    \hline
    \textbf{RGB} & Man biking \cite{soomro2012ucf101} & \includegraphics[width=0.50in]{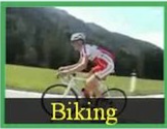} &   \textbf{Audio} & Jumping \cite{ofli2013berkeley} & \includegraphics[width=0.50in]{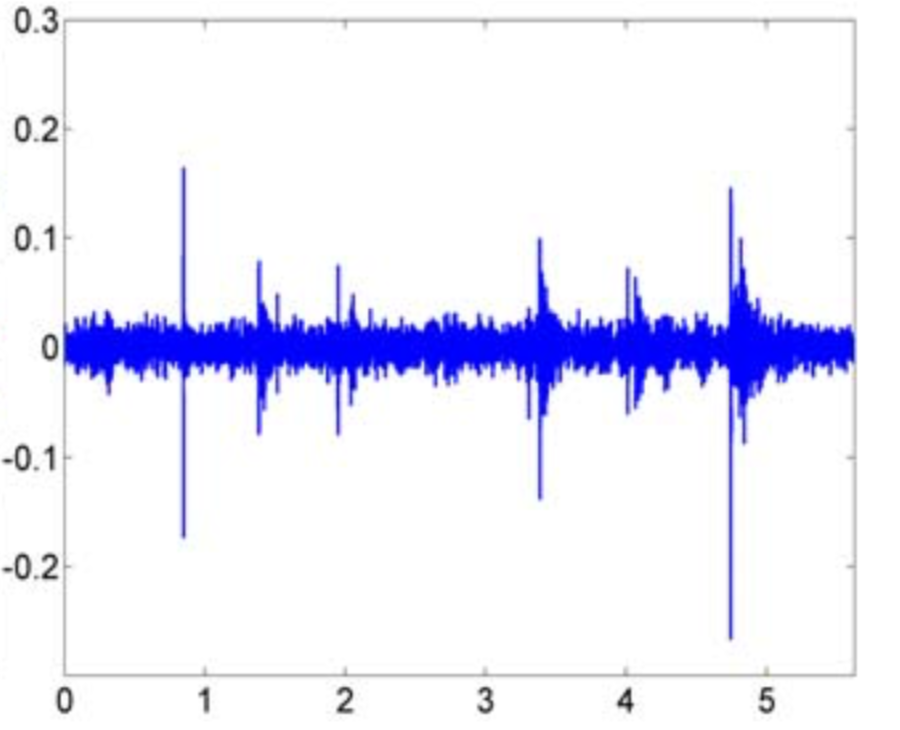}\\
    \hline
    \textbf{Depth} & Person waving \cite{xia2012view} & \includegraphics[width=0.50in]{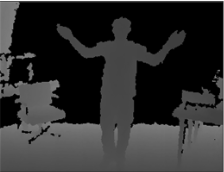} & 
    \textbf{Radar} & Jogging \cite{chakraborty2022diat} & \includegraphics[width=0.50in]{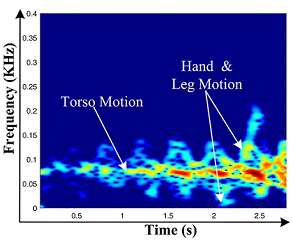}  \\
    \hline 
    \textbf{Skeleton} & Looking at watch \cite{liu2017pku} & \includegraphics[width=0.50in]{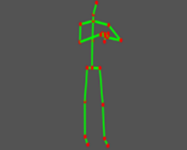} & \textbf{Wifi} & Walking \cite{yousefi2017survey} & \includegraphics[width=0.50in]{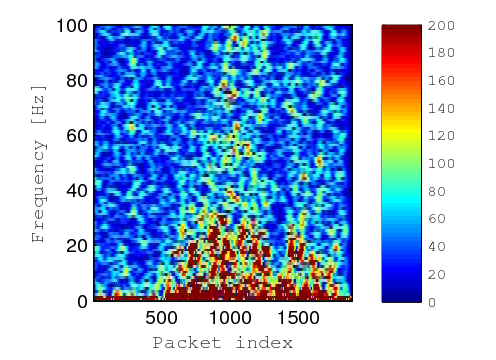} \\ 
    \hline
    
    \textbf{Event Stream} & Walking \cite{calabrese2019dhp19} & \includegraphics[width=0.50in]{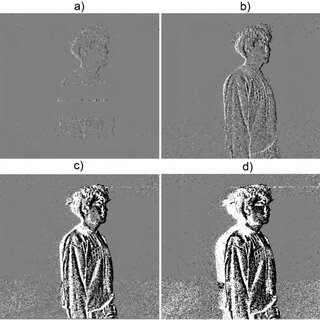} & \textbf{Acceleration } & Walking \cite{kwapisz2011activity} & \includegraphics[width=0.60in]{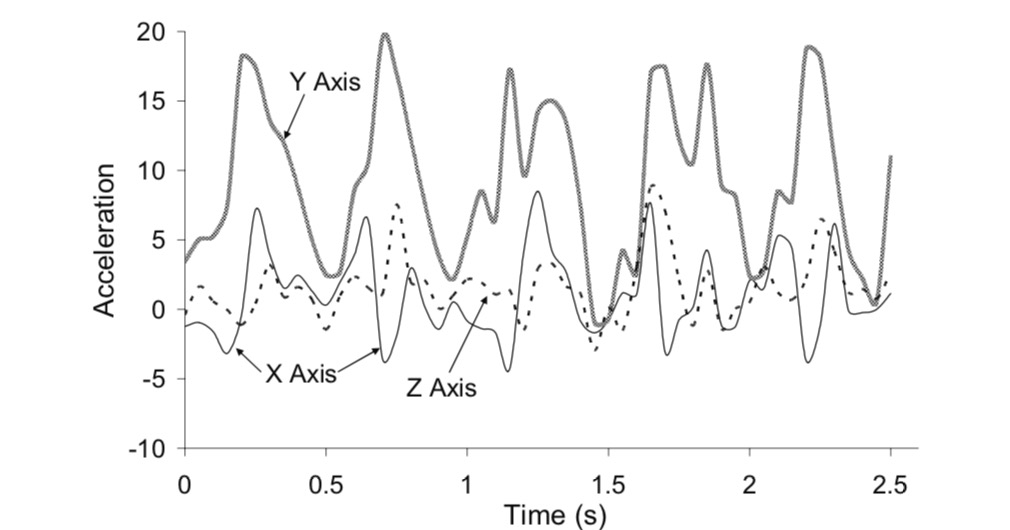} \\
    \hline
    \textbf{Point Cloud} & Bending over \cite{cheng2016orthogonal} & \includegraphics[width=0.50in]{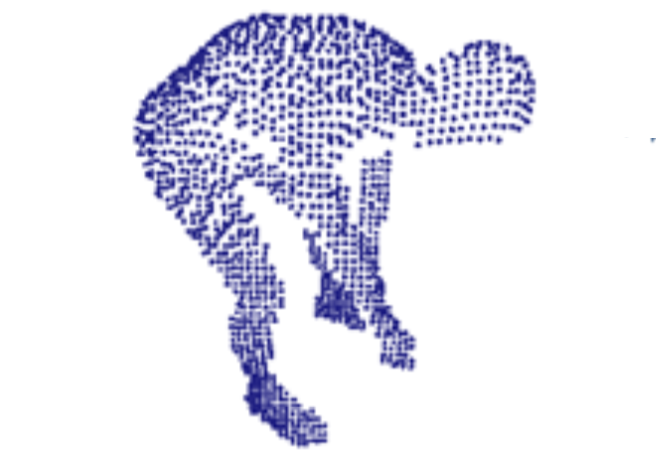} &  &  &\\
    \hline
    \textbf{Infrared Seq} & Pushing \cite{gao2016infar}& \includegraphics[width=0.50in]{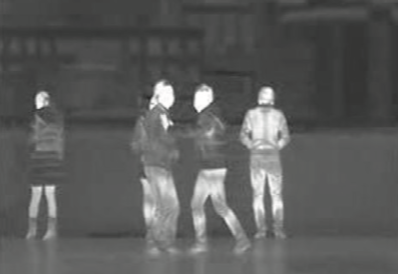} &  &  &\\
    \hline
     
  \end{tabular}
  \caption*{Examples of action categories highlighting the diverse range of objects associated with the actions in HAR.} 
\end{table}

Various data representations possess distinct characteristics with their respective strengths and weaknesses. Several studies \cite{rahmani20163daction, jiang2017learningspatiotemporal, wang20203dv, ghosh2019spatiotemporal, liang2019audiobased, zeng2014convolutional, kim2015humandetection, wang2019temporalunet}  have utilized datasets with various representations and modalities for HAR. In this section, we discuss these briefly, specifically focusing on visual datasets.

\textbf{RGB Representation:} RGB data representation, captured through RGB cameras, mirrors human visual perception by offering detailed insights into colors and textures, essential for comprehensive analysis in domains such as visual surveillance \cite{lin2008human}, autonomous navigation \cite{lu2020driver}, and sports analysis \cite{soomro2015action}. However, it faces limitations such as sensitivity to viewpoint changes, background clutter, and variations in lighting conditions, necessitating advanced preprocessing and feature extraction for effective HAR. Despite these challenges, RGB videos, preferred for capturing the dynamic nature of human movements, remain central to HAR research \cite{chaquet2013survey}, with static images playing a lesser role \cite{delaitre2010recognizing, yao2010grouplet, sharma2012discriminative}. The advent of deep learning has significantly advanced RGB video analysis for HAR, introducing powerful architectures that have become the research focus, complemented by techniques to integrate motion information for improved accuracy \cite{tran2017two}.
\textbf{Skeleton Representation:} Skeleton sequences capture the trajectories of human body joints, effectively representing meaningful human movements. It is commonly used with Graph Convolutional Networks (GCNs) and Transformers. Consequently, skeleton data serves as a fitting modality for HAR. This data type can be obtained through pose estimation algorithms applied to RGB videos \cite{sun2019deep} or depth maps \cite{shotton2011real}. In addition, such data can be collected using motion-capture systems. However, it is important to note that human pose estimation is typically sensitive to variations in viewpoint.

\textbf{Depth Representation:} Depth maps are visual representations in which pixel values encode distance information from a particular viewpoint to the scene's points. This modality demonstrates resilience against color and texture variations, offering dependable 3D structural and geometric shape details of human subjects, making it suitable for HAR. Different types of devices are used to obtain depth images, including active sensors and passive sensors like stereo cameras \cite{chen2013survey}.

\textbf{Infrared Representation:}Infrared sensors are commonly utilized for detecting human movements at night, attributed to their capability to function independently of external ambient lighting. There are two types of methods of infrared sensing: active and passive. Active infrared sensors operate by emitting infrared light, which is then detected upon reflection from objects. Conversely, passive infrared sensors function by detecting the infrared radiation naturally emitted by objects without the need to emit any light themselves.

\textbf{Point Cloud Representation:} Point cloud data comprise many points that indicate an object's spatial distribution and surface attributes in relation to a spatial reference structure \cite{shleibik20233d}. Numerous techniques are available for acquiring 3D point cloud data, among which image-based 3D reconstruction methods stand out \cite{wang20203dv}. The point cloud representation significantly excels in depicting spatial outlines and the 3D geometric shape of subjects as a three-dimensional data modality, making it exceptionally suitable for HAR.

\textbf{Event Streams:} Neuromorphic cameras, or event cameras, have emerged as a promising technology for capturing changes in illumination on a pixel-by-pixel basis, generating asynchronous events with high temporal resolution \cite{innocenti2021temporal}. Unlike traditional video cameras that capture entire frames, event cameras respond only to changes, making them highly effective for tracking fast-moving objects without the motion blur associated with conventional RGB cameras. These cameras offer advantages such as high dynamic range, low latency, reduced energy consumption, and motion blur elimination, making them ideal for Human Activity Recognition. They excel by focusing on foreground motion, thereby reducing data redundancy. Despite their benefits, the data they produce is sparse and asynchronous, which presents unique challenges. Key models include the Dynamic Vision Sensor \cite{lichtsteiner2008128} and the Dynamic and Active-Pixel Vision Sensor \cite{berner2013240}, showcasing the technology's potential for advanced visual sensing.

\begin{table}[h]
\caption{A list of Single and Multi-Modal Datasets used for HAR}
\label{datasets2}
\begin{tabular*}{\textwidth}{@{\extracolsep\fill}lccccccccc}
\toprule
Dataset & RGB & S & D & IR & PC & ES & Au & Ac & Gyr\\
\midrule
UAV-Human \cite{li2021uav} & \checkmark & \checkmark & \checkmark & \checkmark & \checkmark & \checkmark & $\times$ & $\times$ & $\times$ \\
HMDB51 \cite{kuehne2011hmdb} & \checkmark & $\times$ & $\times$ & $\times$ & $\times$ & $\times$ & $\times$ & $\times$ & $\times$ \\
Kinetics-400 \cite{kay2017kinetics} & \checkmark & $\times$ & $\times$ & $\times$ & $\times$ & $\times$ & $\times$ & $\times$ & $\times$ \\
Kinetics-600 \cite{carreira2018short} & \checkmark & $\times$ & $\times$ & $\times$ & $\times$ & $\times$ & $\times$ & $\times$ & $\times$ \\
Kinetics-700 \cite{carreira2019short} & \checkmark & $\times$ & $\times$ & $\times$ & $\times$ & $\times$ & $\times$ & $\times$ & $\times$ \\
EPIC-KITCHENS-55 \cite{damen2018scaling} & \checkmark & $\times$ & $\times$ & $\times$ & $\times$ & $\times$ & \checkmark & $\times$ & $\times$ \\
UCF-101 \cite{soomro2012ucf101} & \checkmark & $\times$ & $\times$ & $\times$ & $\times$ & $\times$ & $\times$ & $\times$ & $\times$ \\
HMDB-51 \cite{kuehne2011hmdb} & \checkmark & \checkmark & \checkmark & \checkmark & $\times$ & $\times$ & $\times$ & $\times$ & $\times$ \\
THUMOS Challenge 15 \cite{gorban2015thumos} & \checkmark & $\times$ & $\times$ & $\times$ & $\times$ & $\times$ & $\times$ & $\times$ & $\times$ \\
ActivityNet \cite{caba2015activitynet} & \checkmark & $\times$ & $\times$ & $\times$ & $\times$ & $\times$ & $\times$ & $\times$ & $\times$ \\
Something-Something-v1 \cite{goyal2017something} & \checkmark & $\times$ & $\times$ & $\times$ & $\times$ & $\times$ & $\times$ & $\times$ & $\times$ \\
Something-Something-v2 \cite{goyal2017something} & \checkmark & $\times$ & $\times$ & $\times$ & $\times$ & $\times$ & $\times$ & $\times$ & $\times$ \\
NTU RGB+D \cite{shahroudy2016ntu} & \checkmark & \checkmark & \checkmark & \checkmark & $\times$ & $\times$ & $\times$ & $\times$ & $\times$ \\
NTU RGB+D 120 \cite{liu2019ntu} & \checkmark & \checkmark & \checkmark & \checkmark & $\times$ & $\times$ & $\times$ & $\times$ & $\times$ \\
MSRDailyActivity3D \cite{wang2012mining} & \checkmark & \checkmark & \checkmark & $\times$ & $\times$ & $\times$ & $\times$ & $\times$ & $\times$ \\
Northwestern-UCLA \cite{wang2014cross} & \checkmark & \checkmark & \checkmark & $\times$ & $\times$ & $\times$ & $\times$ & $\times$ & $\times$ \\
UWA3D Multiview \cite{rahmani2014hopc} & \checkmark & \checkmark & \checkmark & $\times$ & $\times$ & $\times$ & $\times$ & $\times$ & $\times$ \\
UWA3D Multiview II \cite{rahmani2016histogram} & \checkmark & \checkmark & \checkmark & $\times$ & $\times$ & $\times$ & $\times$ & $\times$ & $\times$ \\
InfAR \cite{gao2016infar} & $\times$ & $\times$ & $\times$ & \checkmark & $\times$ & $\times$ & $\times$ & $\times$ & $\times$ \\
DvsGesture \cite{amir2017low} & $\times$ & $\times$ & $\times$ & $\times$ & $\times$ & \checkmark & $\times$ & $\times$ & $\times$ \\
DHP19 \cite{calabrese2019dhp19} & $\times$ & \checkmark & $\times$ & $\times$ & $\times$ & \checkmark & $\times$ & $\times$ & $\times$ \\
MMAct \cite{kong2019mmact} & \checkmark & \checkmark & $\times$ & $\times$ & $\times$ & $\times$ & $\times$ & \checkmark & \checkmark \\
UTD-MHAD \cite{chen2015utd} & \checkmark & \checkmark & $\times$ & $\times$ & $\times$ & $\times$ & $\times$ & \checkmark & \checkmark \\
PKU-MMD \cite{liu2017pku} & \checkmark & \checkmark & \checkmark & \checkmark & $\times$ & $\times$ & $\times$ & $\times$ & $\times$ \\
UCFKinect \cite{ellis2013exploring} & $\times$ & \checkmark & $\times$ & $\times$ & $\times$ & $\times$ & $\times$ & $\times$ & $\times$ \\
HAA500 \cite{chung2021haa500} & \checkmark & $\times$ & $\times$ & $\times$ & $\times$ & $\times$ & $\times$ & $\times$ & $\times$ \\
NEU-UB \cite{kong2017max} & \checkmark & $\times$ & \checkmark & $\times$ & $\times$ & $\times$ & $\times$ & $\times$ & $\times$ \\
AVA \cite{gu2018ava} & \checkmark & $\times$ & \checkmark & $\times$ & $\times$ & $\times$ & $\times$ & $\times$ & $\times$ \\
\midrule
\end{tabular*}
\caption*{S: Skeleton, D: Depth, IR: Infrared, PC: Point Cloud, ES: Event Stream, Au: Audio, Ac: Acceleration, Gyr: Gyroscope.}
\end{table}

A wide array of datasets has been created to train and evaluate various HAR models. Among these, some datasets are categorized as a single modality, primarily utilizing RGB data, which is the most common form used for HAR methods. In contrast, multimodal datasets encompass RGB data along with additional information such as skeleton (S), depth (D), infrared (IR), point cloud (PC), event stream (ES), audio (Au), acceleration (Ac), and gyroscope (Gyr) data, offering a more comprehensive dataset for nuanced activity recognition. We have included Table ~\ref{datasets2}, which presents a comparison of various datasets, detailing the diverse types of information each dataset provides.

In this section, it is also pertinent to highlight notable RGB-based datasets for image-based HAR, including HICO \cite{lin2014microsoft}, HICO-DET \cite{chao2018learning}, and V-COCO \cite{lin2014microsoft}. These datasets are instrumental in depicting a broad spectrum of human activities through RGB data, thus enriching image-based HAR research with valuable activity insights. Furthermore, video datasets for HAR also can be divided into controlled and uncontrolled categories. Controlled datasets, such as the Weizmann and KTH \cite{gorelick2007actions, schuldt2004recognizing}, are generated under predefined conditions with actors performing specified actions. In contrast, uncontrolled or 'in the wild' datasets are derived from naturally occurring videos, including films, surveillance footage, or YouTube content, providing a more diverse range of human activities.

Most popular vision-based datasets in this field of HAR are mentioned in Table~\ref{datasets3}. Among these, significant benchmarks are designed for the analysis of RGB images derived from videos, including UCF101~\cite{soomro2012ucf101}, HMDB-51~\cite{kuehne2011hmdb}, Kinetics-400~\cite{kay2017kinetics}, Kinetics-600~\cite{carreira2018short}, Kinetics-700~\cite{carreira2019short}, EPICKITCHENS-55~\cite{damen2018scaling}, THUMOS Challenge 15~\cite{gorban2015thumos}, ActivityNet~\cite{caba2015activitynet}, and Something-Something-v1~\cite{goyal2017something}. These datasets focus on leveraging RGB video data for HAR tasks. Some of these datasets incorporate RGB images along with other modalities, such as skeleton data, depth information, and more, to provide a more comprehensive understanding of human activities, such as NTU RGB+D~\cite{shahroudy2016ntu} and NTU RGB+D 120~\cite{liu2019ntu}, which enhance the depth and complexity of HAR model evaluation and development. 

\begin{figure}[hbt!]
    \centering
    \includegraphics[width=10cm]{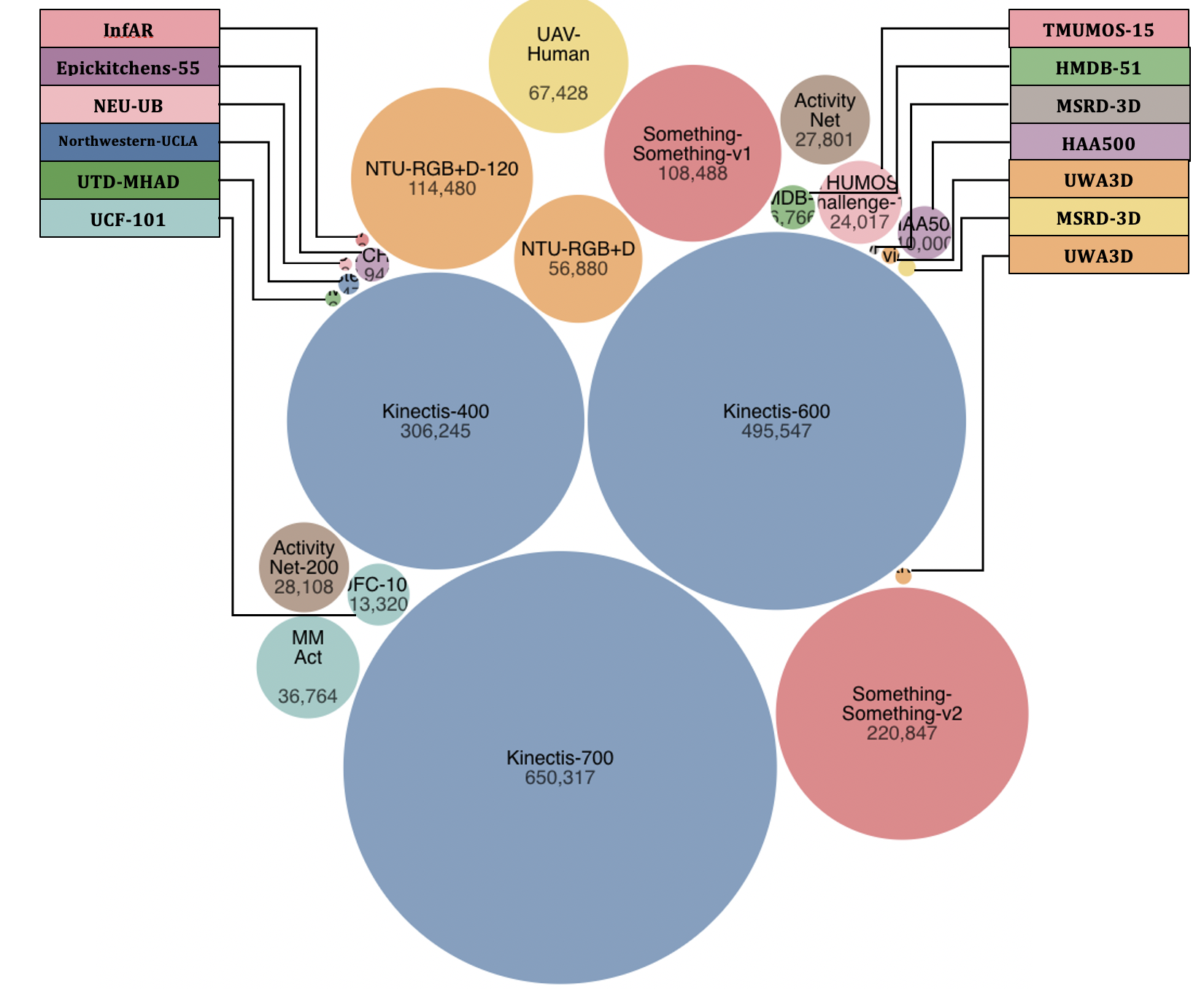}
    \caption{Human Action Recognition Datasets based on the data simple numbers}
    \label{fig:HAR_dataset_graph}
\end{figure}

The large circles in the bubble diagram Figure~\ref{fig:HAR_dataset_graph} depict the most common datasets utilized for HAR with many samples and classes. These datasets were used with a variety of network architectures, contributing to the advancement of HAR mythologies. Recently, RGB-based networks, including single-stream, two-stream, 3D Convolutional, motion-based, and Transformer networks, use Kinetics-400/600/700 ~\cite{kay2017kinetics}\cite{carreira2018short}\cite{carreira2019short}\cite{kuehne2011hmdb} for evaluation due to the various number of samples. Other RGB-based datasets like UCF101 \cite{soomro2012ucf101}, and HMDB-51 \cite{kuehne2011hmdb} are still used for the problem, as shown in Figure~\ref{Tab:TS-3D-MN-based}.

\begin{table}[h]
\caption{Some HAR Benchmark Datasets with Various Data Modalities}
\label{datasets3}
\begin{tabular*}{\textwidth}{@{\extracolsep\fill}lcccccccc}
\toprule%
Dataset &Year &\#Classes &\#Subject &\#Samples &\#Viewpoint &Code  \\
\midrule
HMDB-51 \cite{kuehne2011hmdb}&2011  &51 &- &6,766 &- &\href{https://serre-lab.clps.brown.edu/resource/hmdb-a-large-human-motion-database/#dataset}{\faGithubSquare}\\
UCF-101 \cite{soomro2012ucf101} &2012 &101 &- &13,320 & -& \href{https://www.crcv.ucf.edu/data/UCF101.php}{\faGithubSquare} \\
MSRDailyActivity3D \cite{wang2012mining} &2012 & 16 &10& 320&1&\href{https://wangjiangb.github.io/my_data.html}{\faGithubSquare} \\
UCFKinect \cite{ellis2013exploring} &2013 &16 &16 &12,80 &1& - \\
jHMDB \cite{jhuang2013towards} &2013 & 16 &10& 928&1&\href{http://jhmdb.is.tue.mpg.de/}{\faGithubSquare} \\
UCFKinect \cite{ellis2013exploring} &2013 &16 &16 &12,80 &1& - \\

Northwestern-UCLA \cite{wang2014cross}&2014 & 10 &10& 1475&3 &\href{https://wangjiangb.github.io/my_data.html}{\faGithubSquare}\\
UWA3D Multiview \cite{rahmani2014hopc}&2014 & 30 &10& 900&4  & - \\
UWA3D Multiview II \cite{rahmani2016histogram}&2015 & 30 &10& 1075&4 &\href{https://ieee-dataport.org/documents/uwa-3d-multiview-activity-ii-dataset}{\faGithubSquare}\\
UTD-MHAD \cite{chen2015utd}&2015& 27 & 8& 861&1 &\href{https://personal.utdallas.edu/~kehtar/UTD-MHAD.html}{\faGithubSquare}\\
THUMOS Challenge 15 \cite{gorban2015thumos} &2015 &101&- &24,017 &-&\href{http://www.thumos.info/}{\faGithubSquare}\\
ActivityNet \cite{caba2015activitynet} &2015 &203&- &27,801 &-&\href{http://activity-net.org/}{\faGithubSquare}\\
InfAR \cite{gao2016infar} &2016 & 12 &40& 600&2&\href{https://app.dimensions.ai/details/publication/pub.1008631317}{\faGithubSquare}\\
NTU RGB+D \cite{shahroudy2016ntu} &2016 & 60 &40& 56880 &80&\href{https://github.com/shahroudy/NTURGB-D}{\faGithubSquare}\\
DvsGesture \cite{amir2017low}&2017& 17 &29& -&- &\href{https://research.ibm.com/publications/a-low-power-fully-event-based-gesture-recognition-system}{\faGithubSquare}\\
NEU-UB \cite{kong2017max} &2017& 6 & 20& 600&-&- \\
PKU-MMD \cite{liu2017pku}&2017& 51 & 66& 1076&3 &\href{https://www.icst.pku.edu.cn/struct/Projects/PKUMMD.html}{\faGithubSquare}\\
Something-Something-v1 \cite{goyal2017something} &2017 &174 &- &108,488&-&\href{https://arxiv.org/abs/1706.04261}{\faGithubSquare}\\
Something-Something v2\cite{goyal2017something} &2017&174 &- &220,847 &-& \href{https://developer.qualcomm.com/software/ai-datasets/something-something}{\faGithubSquare}\\
Kinectis-400 \cite{kay2017kinetics} &2017 & 400 & -& 306,245& -&\href{https://github.com/cvdfoundation/kinetics-dataset}{\faGithubSquare}\\
Kinectis-600 \cite{carreira2018short} &2018 & 600 & -& 495,547& -&\href{https://github.com/cvdfoundation/kinetics-dataset}{\faGithubSquare}\\
EPICKITCHENS-55 \cite{damen2018scaling} &2018 & - & 32& 39,45 & - &\href{https://epic-kitchens.github.io/2020-55.html}{\faGithubSquare}\\
DHP19 \cite{calabrese2019dhp19} &2019& 33 &17& -&4 &\href{https://sites.google.com/view/dhp19/home}{\faGithubSquare}\\
MMAct \cite{kong2019mmact} &2019& 37 & 20& 36764&4+Egocentric&\href{https://mmact19.github.io/2019/}{\faGithubSquare}\\
Kinectis-700 \cite{carreira2019short} &2019 & 700 & -& 650,317& -&\href{https://github.com/cvdfoundation/kinetics-dataset}{\faGithubSquare}\\
NTU RGB+D 120 \cite{liu2019ntu} &2019 & 120 &106& 114,480&155&\href{https://github.com/shahroudy/NTURGB-D}{\faGithubSquare}\\

HAA500 \cite{chung2021haa500}&2021& 500 & -& 10000&-&\href{https://www.cse.ust.hk/haa/}{\faGithubSquare}\\
UAV-Human \cite{li2021uav} &2021 &155 &119 &67,428 &- &\href{https://github.com/sutdcv/UAV-Human}{\faGithubSquare}  \\
ActivityNet-200 \cite{kay2017kinetics} & 2022 & 200 & 141 & 28,108 & 19,994 &\href{https://github.com/kennymckormick/pyskl}{\faGithubSquare} \\
Ego4d \cite{grauman2022ego4d} & 2022 & - & - & - & - & \href{https://ego4d-data.org/}{\faGithubSquare} \\
\midrule
\end{tabular*}
\footnotetext[1]{S: Skeleton, D: Depth, IR: Infrared, PC: Point Cloud, ES: Event Stream, Au: Audio, Ac: Acceleration, Gyr: Gyroscope, EMG: Electromyography.}
\end{table}

In the early stages of HAR, essential RGB-based datasets were utilized, such as UCF-101 \cite{soomro2012ucf101}, which played a key role in advancing human action recognition. UCF-101 comprises more than 13,000 video clips, divided into 101 action-related categories. These categories cover various activities, including playing an instrument, sports, and daily activities. The clips are collected from YouTube, thus providing a realistic and diverse backdrop for each action, which is critical for constructing robust HAR systems. This diversity guarantees that models trained on UCF101 may generalize effectively across multiple circumstances.

The Kinetics series RGB-based dataset by Google DeepMind provides three benchmarks: Kinectis-400 \cite{kay2017kinetics}, Kinectis-600 \cite{carreira2018short}, and Kinectis-700 \cite{carreira2019short}, where the number refers to the number of classes. The Kinetics series dataset provides various classes, including sports, musical activities, daily routines, interpersonal interactions, object manipulations, and occasional animal engagements, catering to a broad spectrum of human actions. The dataset provides rich vision feature information, making it perfect for training for diverse network architectures capable of tackling large amounts of vision data and complex representations, including two-stream, 3D convolutional, motion, and vision-base transformer networks.

Other benchmarks, such as NTU RGB+D \cite{shahroudy2016ntu} and NTU RGB+D 120 \cite{liu2019ntu}, provide skeletal information in addition to RGB data with full depth, IR, and mask information; these characteristics have proven essential for various HAR tasks. These datasets are primarily used with graph neural networks (GNNs) and the Skeleton-Vision-based Transformers model because skeleton data provides a structured representation of human actions, making it useful for learning.

Ego4D is also an extensive egocentric video dataset featuring 3,670 hours of daily life activities across various scenarios, including household, outdoor, workplace, and leisure environments~\cite{grauman2022ego4d}. The footage, captured by 931 unique camera wearers, spans 74 locations in 9 countries worldwide, making it one of its most diverse and comprehensive collections.

Other vision-based datasets not mentioned in other survey papers are OOPS! and Mix reality datasets \cite{pei2021mars}\cite{zhang2021open}. OOPS! \cite{epstein2020oops} is a challenging dataset for recognizing unintentional action in the video.  It consists of 20,338 videos from YouTube failed compilation videos, up to over 50 hours of data. The IMU dataset \cite{pei2021mars}, a Mixed Reality resource, is constructed utilizing virtual Inertial Measurement Units (IMUs). This dataset synthesizes IMU data by leveraging the motion capture dataset AMASS \cite{mahmood2019amass} and the DIB dataset \cite{huang2018deep}, creating a rich foundation for virtual motion analysis. Additionally, the CAPture the Essence of Activities (CAPE) dataset \cite{zhang2021open} meticulously documents the movements of 16 subjects (9 males and seven females) across six activities. These activities are captured using the advanced Perception Neuron Studio \cite{PerceptionNeuronStudio2023} and a wireless motion capture system equipped with 17 inertial body sensors, offering a detailed and diverse dataset for activity analysis.

\subsection{Discussions}

This section has presented a large number of datasets containing samples of videos where humans engage in various types of actions. The number of dataset samples varies from a few hundred to hundreds of thousands. The activities portrayed are usually those found in videos uploaded to internet sites like YouTube or in movies. although the same videos have been created in late settings for specific derived actions. As HAR models start to perform well on existing datasets, more difficult datasets are created so that better models that perform better on these difficult datasets can be developed.

\section{Research Challenges and Limitations}
\label{sec:research challenges}
\begin{figure}[hbt!]
\centerline{\includegraphics[width=13cm]{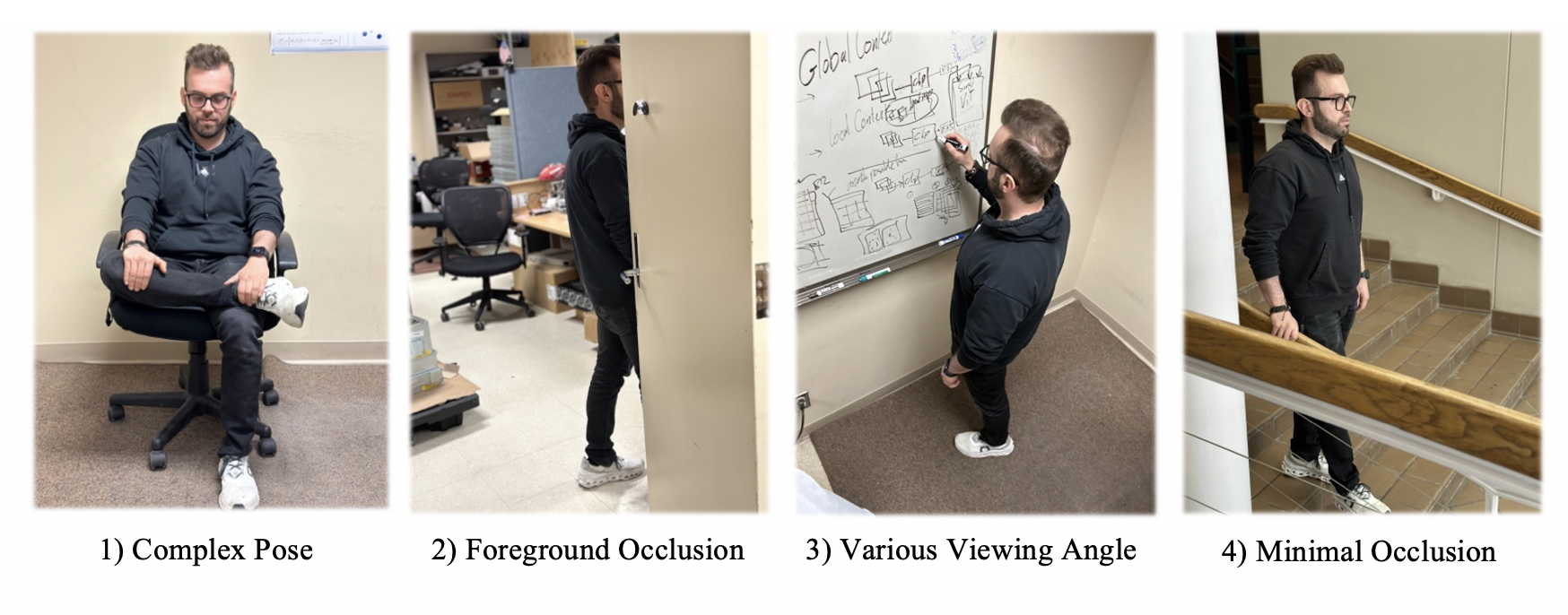}}
\caption{Examples of challenging human activities from videos recorded in our laboratory at the University of Colorado Colorado Springs (UCCS).}
\label{fig:MP}
\end{figure}

Recognizing actions and movements from videos presents a significantly greater challenge than doing so from static images. With images, it is relatively straightforward to employ a classifier model to predict the class or action. In contrast, videos involve dealing with a sequence of images that collectively depict a progression of motion. Further complicating matters, certain human actions can be notoriously difficult to discern even for images, as exemplified in Figure~\ref{fig:MP}. In the first image of Figure~\ref{fig:MP}, which depicts a complex pose, it is quite challenging to discern whether the subject is sitting, attempting to stand, or performing a stretching exercise.

Human action recognition continues to grapple with numerous issues and challenges. Foremost among these are matters of accuracy, efficiency, and generalization. These obstacles persist despite the considerable progress made, underscoring the complexity of the problem and the need for ongoing research efforts.

\subsection{Accuracy}
\label{sec:Accuracy}
Many techniques have been mentioned in this comprehensive study using different network architectures to improve recognition accuracy and address challenges, including complex actions, diverse environmental conditions, and using different benchmarks. However, there is always room for enhancing the accuracy of recognition models \cite{saoudi2023advancing}. In this context, accuracy refers to the model's ability to identify the actions presented in video sequences correctly. The higher the accuracy, the better the model is at distinguishing between different actions and minimizing the number of incorrect predictions.

The quest for improved accuracy is a central theme in most research efforts. Researchers employ various strategies to boost model accuracy, including experimenting with novel network architectures, exploring advanced feature extraction techniques \cite{abdelrazik2023efficient}, and fine-tuning training methodologies \cite{surek2023video, lim2023fine}. Exploring ways to optimize hyperparameters and leveraging ensemble methods are also considered; ensembles can improve a model's predictive accuracy by aggregating multiple models' predictions \cite{tyagi2023proposed}. While accuracy is undeniably important, it is not the sole benchmark of a model's performance. Other factors, such as model robustness, computational efficiency, and the ability to generalize to unseen data, also play crucial roles in determining the effectiveness of a HAR system \cite{mazzia2022action}. While accuracy is undeniably important, it is not the sole benchmark of a model's performance. Other factors, such as model robustness, computational efficiency, and the ability to generalize to unseen data, also play crucial roles in determining the effectiveness of a HAR system \cite{mazzia2022action}.

\subsection{Efficiency}

Another crucial issue in human action recognition pertains to the computational efficiency of models, particularly when processing extensive video data \cite{schiappa2023large}. The substantial computational requirements of large videos can complicate the deployment of models, necessitating the development of more efficient methods. Researchers have begun exploring ways to optimize the input videos for the models in some studies by providing video encoders to leverage motion vectors, which can help reduce computational demands \cite{islam2023representation}. Other techniques, like computing the RGB differences of consecutive frames or integrated optical flow networks to replace the need for external and distinct optical flow computation in streaming scenes, have been employed to enhance efficiency \cite{khan2023enhanced}.

A notable breakthrough in the pursuit of efficiency has been achieved within the realm of 3D convolutional neural networks. Approaches like pseudo-3D Conv and decoupled 3D CNN have effectively reduced the computational overhead of 3D Conv by substituting them with a separated 2D-3D Conv in the frame domain and temporal Conv in the time domain \cite{le2023deep, rajasegaran2023benefits}. Some other recent work, like the ECO-lite method, have attempted to combine 2D and 3D schemes. This approach reduces the computational overhead of low-level feature extraction by only applying 3D Conv to higher-level feature maps \cite{zolfaghari2018eco}.

Enhancing computational efficiency is a critical goal in human action recognition, as it can significantly expedite model deployment and facilitate real-time action recognition in video streams. However, it is a delicate balance between maintaining high accuracy and ensuring model efficiency, one that continues to be the focus of ongoing research \cite{dasari2022human}.

\subsection{Generalization}
\label{sec:Generalization}
As highlighted in Section \ref{sec:datasets}, most recent studies have utilized large-scale video datasets for training. However, these datasets predominantly feature videos from limited domains, such as YouTube or various online platforms. This raises an important question: How can models trained on such specific datasets be effectively applied to videos originating from a broader range of sources? Consequently, one of the paramount challenges in Human Activity Recognition (HAR) centers on enhancing the models' ability to generalize across diverse video content. The ability to generalize across various sources has indeed become a significant hurdle, particularly with the rapid evolution of video modeling technology.

The solution to this issue is fundamentally rooted in improving the generalization capabilities of video models. Activities that involve musical instrument interaction, like playing the violin, piano, and flute, are inherently compositional and exhibit fine-grained differences. Musical instruments exhibit considerable physical diversity, from their design and interaction with the human body to the techniques employed to produce sound. Similarly, what may appear as minor posture adjustments to the untrained eye can signify vastly different activities to experts in fields such as gymnastics. Addressing this challenge, developing new, varied, and specialized datasets has emerged as a promising solution, an initiative that numerous researchers are actively exploring. Developing HAR models with robust generalization capabilities is essential to effectively recognize and interpret human actions across various contexts and video sources. Pursuing this goal remains a key focus of ongoing research in this dynamic area.

\subsection{Others Challenges and Issues}
Beyond the primary challenges previously discussed, researchers must navigate numerous other issues in HAR. Various survey papers have outlined these challenges, providing unique perspectives and insights into the intricacies of the field \cite{kong2022human}. Human action recognition in video is a complex task that involves understanding not just the motion of individual parts of a human figure but also their interrelations and the overall context and human intent. Although significant progress has been made in this field, several limitations remain as follows:

\subsubsection{Occlusion}
In occlusion, parts of the human body may be occluded or hidden from view during an action, making it challenging for a system to recognize the action correctly. As shown in pictures 2 and 4 of Figure~\ref{fig:MP}, a significant portion of the body is obscured by the door and stair railing, making it difficult to recognize the action. The presence of occlusion in human action presents several challenges in building a robust human action recognition system. Occlusion can result in the loss of important motion data, affecting the extraction of relevant features from the data, increasing complexity in data processing, and disrupting temporal analysis \cite{vernikos2023human, chang2020towards}. Hence, dealing with occlusion remains a significant and unresolved challenge in human action recognition \cite{shi2023occlusion}.

\subsubsection{Viewpoint Variation}
Viewpoint variation in HAR is a complex and challenging problem for several reasons. When a video is captured from different viewpoints, the appearance of human actions can change dramatically. As shown in picture 3 of Figure~\ref{fig:MP}, we can not see most of the body parts because of the viewing angle. Current systems may struggle to recognize actions from different viewpoints and pose challenges to HAR systems \cite{zhao2023multifeature}. Human actions can involve a variety of interactions and complex movements with the environment, which can appear differently when observed from various viewpoints \cite{singh2023recent}. The challenge of recognizing human actions from arbitrary viewpoints arises due to several factors, including changes in the visual representation of the action, variations in spatial relationships, and the presence of occlusions, all of which depend on the angle of observation \cite{ghosh2023deep}.

\subsubsection{Background Clutter and Lighting Conditions}
Background clutter and lighting conditions are other challenges for human action recognition systems. Changes in lighting and cluttered or changing backgrounds can affect the ability of systems to detect and recognize actions \cite{bousmina2023hybrid, gowada2023unethical}. Background clutter and lighting conditions significantly impact the performance of HAR systems in various ways. Cluttered backgrounds can easily confuse HAR systems by adding additional noises and causing occlusion of important features, leading to false positives. Changing lighting conditions can create visibility issues, such as shadows. In real-world scenarios, where lighting changes dynamically, it is challenging to build adaptable and robust HAR systems that can maintain consistent performance under these varying conditions \cite{hussain2023low}.

\subsubsection{The Lack of Large-Scale \& Well-Annotated Datasets}
The lack of large-scale, well-annotated datasets poses various challenges for HAR systems, particularly those designed using machine learning and deep learning approaches. Large-scale and diverse video datasets covering a wide range of actions from various environments, cultures, age groups, and settings are essential. The absence of such diversified, large-scale datasets may result in biased HAR systems \cite{song2024gtautoact}. Additionally, a lack of large-scale datasets and poor-quality annotations can lead to system failures in recognizing the correct actions \cite{li2023sar}.

\subsubsection{Human Movement Complexity}
Human movements and actions are highly complex and diverse, presenting significant challenges for human action recognition systems in detecting and recognizing these actions in videos \cite{serpush2021complex}. Human actions are not uniform, dynamically changing with each movement and varying from person to person. This variability complicates the design of a robust HAR model. Furthermore, the challenge is compounded by the simultaneous movement of multiple body parts, where capturing and recognizing these concurrent movements becomes a complex task for HAR systems. Adding to this complexity is the unpredictability of human actions, such as unexpectedly falling down while walking, which HAR systems may find difficult to anticipate and accurately identify the actions \cite{lin2023hieve}.

\subsubsection{Real-Time Human Action Applications}
Real-time human action recognition applications are essential for the fast and efficient processing of complex human actions from videos, providing immediate responses without delay as the actions occur. However, it is challenging due to several factors, including computational complexity, latency issues, dynamic and unpredictable human actions, and resource constraints on edge devices \cite{wensel2023vit, kumar2024human}. Many applications require real-time recognition, but processing video data and recognizing actions in real-time can be computationally expensive and may not be feasible with current technology, particularly for high-resolution videos \cite{sarraf2023optimal, diraco2023review}.

\subsubsection{Interactions}
Interactions with objects and other people are also not easy tasks in HAR. Many human actions involve interaction with objects or other people. Recognizing such interactions can be challenging, particularly when the objects or other people are also moving \cite{verma2023human, faure2023holistic}.

\section{Future Directions in HAR}

Human Action Recognition (HAR) is rapidly advancing and fueled by innovations in machine learning, deep learning, and video processing technologies. Despite these strides, the challenges outlined in Section~\ref{sec:research challenges} reveal several vital areas for further exploration. This survey highlights the potential of hybrid models and recent research utilizing multiple modalities for enhanced performance. While vision-based approaches are practical in recognizing human actions in videos, incorporating additional data streams—such as audio, sensor data, and textual information (e.g., scene descriptions)—can significantly improve accuracy and generalization \cite{sun2022human}. By leveraging multimodal learning, HAR systems can better handle complex and occluded environments.

Another contribution to our survey is the open-world recognition section~\ref{sec:Open World Recognition}. We mentioned in videos how vital it is for HAR models to deal with novel actions in different scenarios. Due to the scarcity of annotated datasets, particularly for rare or intricate actions, techniques like open-world, few-shot, and zero-shot learning are gaining traction \cite{ruan2024advances, estevam2021zero}. These methods allow HAR models to generalize from minimal examples or identify actions not previously encountered during training.

This survey highlights various model architectures and data modalities used in HAR. Understanding how models leverage data and make decisions has become critical to advancing deep learning-based real-world applications, Explainable AI (XAI) is a rapidly growing field that focuses on these insights. Typically, XAI techniques are tailored for specific vision architectures, including CNNs and Vision Transformers (ViTs). Commonly used methods for CNNs include saliency maps, Layer-wise Relevance Propagation (LRP), Integrated Gradients, and Class Activation Mapping (CAM) \cite{mersha2024explainable}. Some HAR works have started to explore this space \cite{jeyakumar2023x, roy2021explainable, pellano2024movements}, but with the explosion of XAI, we anticipate significant future efforts to align the state-of-the-art in HAR with XAI.

As HAR applications expand into real-time domains such as surveillance, autonomous driving, and assistive technologies, the demand for efficient, low-latency models becomes critical. A promising area for future development is action anticipation—predicting future actions based on partially observed sequences \cite{zhang2020egocentric}. Similarly, early action recognition, identifying an action at its onset \cite{wang2019progressive}, holds the potential for preventing accidents and issuing early warnings in critical scenarios.

Ethical considerations are increasingly crucial as HAR systems become integral to sensitive domains like security, healthcare, and workplace monitoring. Addressing biases in datasets that can lead to unfair or inaccurate outcomes for specific demographic groups is essential for ensuring fairness. Additionally, HAR models should be transparent and interpretable, allowing their decision-making processes to be understood and audited by users and stakeholders.

Future research should prioritize the development of efficient fusion techniques and models capable of processing diverse data modalities in real-time. Furthermore, building robust architectures that can learn effectively from sparse data and generalize to unseen actions is crucial for applications in real-world environments where collecting comprehensive datasets is often impractical.

\section{Conclusion}
\label{sec:summary}

In this survey, we have presented a comprehensive review of Human Action Recognition (HAR) vision-based approaches using deep learning networks, starting from the foundational ones and continuing with recent works. We discussed in detail most of the papers that have an impact on the field. Our novel SMART-Vision taxonomy (Figure \ref{fig:HARVennDiagrame}) presented in our survey aids researchers in the field by illustrating how innovations in Deep Learning are interconnected, and it guides them to relevant papers for each intersection through the tables provided in each subsection of the deep learning section. Through our taxonomy, we've found that foundational shifts in literature have evolved into advanced hybrid approaches that transcend the traditional methodologies in HAR. By analyzing the most recent literature, we found promising research directions gaining traction, including temporal reasoning, attention modules, several GCN methodologies, excitation modules, and utilizing vision-language models for HAR. 
We also identified Open-HAR as an emerging problem space ripe for innovation, with active research rapidly proposing new and diverse systems, evaluations, and metrics.
Our examination of HAR datasets (section \ref{sec:datasets}) provided an in-depth analysis of modern HAR datasets, noting the popularity of large-scale datasets such as Kinetics-700\cite{carreira2019short} and the use of additional data beyond RGB. 
We also discussed several nuanced challenges in HAR that have yet to be studied extensively.
Our survey has provided a broad snapshot of the state of Human Action Recognition, presenting new insights about hybrid approaches drawn from a breadth of literature and thoroughly examines current research and future directions.

\bibliographystyle{unsrtnat}
\bibliography{references}

\end{document}